\documentclass{article}

\usepackage{amsmath,amsfonts,amssymb,amsthm,bm}
\usepackage{booktabs} % for professional tables
\usepackage{enumitem}
\usepackage{nicefrac}       % compact symbols for 1/2, etc.
\usepackage{graphbox}
\usepackage{graphicx}
\usepackage{microtype}
\usepackage{caption}
\usepackage[framemethod=tikz]{mdframed}
\usepackage{subcaption}
\usepackage{hyperref}
\usepackage{afterpage}
\usepackage{tikz}
\usepackage{pgfplots}
\pgfplotsset{compat=newest}
\usepackage{setspace}

\usepackage[acronym, nomain, nonumberlist]{glossaries}
\makenoidxglossaries

\newacronym{mnist}{MNIST}{Modified National Institute of Standards and Technology}
\newacronym{ode}{ODE}{Ordinary Differential Equation}
\newacronym{svm}{SVM}{Support Vector Machine}
\newacronym{sgd}{SGD}{Stochastic Gradient Descent}
\newacronym{relu}{ReLU}{Rectified Linear Unit}
\newacronym{scm}{SCM}{Soft Committee Machine}
\newacronym{lwf}{LwF}{Learning without Forgetting}
\newacronym{ewc}{EWC}{Elastic Weight Consolidation}
\newacronym{iid}{i.i.d.}{independent and identically distributed}

\setlist[itemize]{noitemsep}

\graphicspath{{Figs/}{Figs/}{Figs/}}

\newcommand{\textcite}[1]{\citet{#1}}

\glsdisablehyper

\newcommand{\erf}{\mathrm{erf}}

\usepackage{xcolor}
\definecolor{C0}{HTML}{1f77b4}
\definecolor{C1}{HTML}{ff7f0e}
\definecolor{C2}{HTML}{2ca02c}
\definecolor{C3}{HTML}{d62728}
\definecolor{C4}{HTML}{9467bd}
\definecolor{C5}{HTML}{8c564b}
\definecolor{C6}{HTML}{e377c2}
\definecolor{C7}{HTML}{7f7f7f}
\definecolor{C8}{HTML}{bcbd22}
\definecolor{C9}{HTML}{17becf}
\usepackage{hyperref}

\usepackage[accepted]{icml2021}

\icmltitlerunning{Continual Learning in the Teacher-Student Setup}

\begin{document}

\twocolumn[
\icmltitle{Continual Learning in the Teacher-Student Setup: Impact of Task Similarity}




\begin{icmlauthorlist}
\icmlauthor{Sebastian Lee}{imperial}
\icmlauthor{Sebastian Goldt}{sissa}
\icmlauthor{Andrew Saxe}{oxford,cifar,fair}
\end{icmlauthorlist}

\icmlaffiliation{imperial}{Imperial College, London, UK}
\icmlaffiliation{sissa}{International School of Advanced Studies (SISSA), Trieste, Italy}
\icmlaffiliation{oxford}{Department of Experimental Psychology, University of
  Oxford, UK}
\icmlaffiliation{cifar}{CIFAR Azrieli Global Scholars program, CIFAR, Toronto, Canada}
\icmlaffiliation{fair}{Facebook AI Research}

\icmlcorrespondingauthor{Sebastian Lee}{sebastian.lee14@imperial.ac.uk}
\icmlcorrespondingauthor{Andrew Saxe}{andrew.saxe@psy.ox.ac.uk}

\icmlkeywords{Machine Learning, ICML}

\vskip 0.3in
]



\printAffiliationsAndNotice{}%\icmlEqualContribution} % otherwise use the standard text.

\begin{abstract}
  Continual learning—the ability to learn many tasks in sequence—is critical for
  artificial learning systems. Yet standard training methods for deep networks
  often suffer from catastrophic forgetting, where learning new tasks erases
  knowledge of earlier tasks. While catastrophic forgetting labels the
  problem, the theoretical reasons for interference between tasks remain
  unclear. Here, we attempt to narrow this gap between theory and practice by
  studying continual learning in the teacher-student setup. We extend previous
  analytical work on two-layer networks in the teacher-student setup to multiple
  teachers. Using each teacher to represent a different task, we investigate how
  the relationship between teachers affects the amount of forgetting and
  transfer exhibited by the student when the task switches. In line with recent
  work, we find that when tasks depend on similar features, intermediate task
  similarity leads to greatest forgetting. However, feature similarity is only
  one way in which tasks may be related. The teacher-student approach allows us
  to disentangle task similarity at the level of \emph{readouts}
  (hidden-to-output weights) and \emph{features} (input-to-hidden
  weights). We find a complex interplay between both types of similarity,
  initial transfer/forgetting rates, maximum transfer/forgetting, and
  long-term transfer/forgetting. Together, these results
  help illuminate the diverse factors contributing to catastrophic forgetting.
\end{abstract}

\section{Introduction}

One of the biggest open challenges in machine learning is the ability to
effectively perform continual learning: learning tasks sequentially. A
significant hurdle in getting systems to do this effectively is that models
trained on task~A followed by task~B will struggle to learn task~B without
un-learning task~A. This is known as catastrophic interference or
\emph{catastrophic forgetting}~\cite{mccloskey1989catastrophic,
  goodfellow2013empirical}, which occurs because weights that contain important
information for the first task are overwritten by information relevant to the
second. The harmful effects of catastrophic forgetting are not limited to
continual learning.  They also play a role in multi-task learning, reinforcement
learning and standard supervised learning, for example under distribution
shift~\cite{arivazhagan2019massively, toneva2018empirical}.

As a result, the
phenomenon has received increased interest in recent
years. % and numerous methods
%to mitigate its effects have been proposed (mostly centred around three themes: regularisation, replay, 
%and dynamic architectures)~\cite{parisi2019continual}. 
In neuroscience, much work has been done to understand the brain's ability to
consolidate learning from earlier tasks, thereby making it relatively robust to
forgetting~\cite{flesch2018comparing, cichon2015branch,
  yang2014sleep}. Similarly, a series of works has started a systematic
empirical analysis of this phenomenon in deep
networks~\cite{parisi2019continual, mirzadeh2020understanding,
  neyshabur2020being, nguyen2019toward, ruder2017learning}. These works consistently observed a counter-intuitive
role of the similarity between tasks A and B, with \emph{intermediate} task
similarity leading to worst forgetting~\cite{ramasesh2020anatomy,
  doan2020theoretical, nguyen2019toward}.

The purpose of this work is to tackle continual learning from the complementary
perspective of high-dimensional teacher-student models~\cite{gardner1989,
  seung1992statistical, biehl1993learning, zdeborova2016statistical}. These models are a popular framework for studying
machine learning problems in a controlled setting, and have recently seen a
surge of interest in attempts to understand generalisation in deep neural
networks.

% The following commands seem to lead to compilation errors
\newcommand{\xb}{\mathbf{x}}
\newcommand{\wb}{\mathbf{W}}
\newcommand{\wvecb}{\mathbf{w}}
\newcommand{\vb}{\mathbf{v}}
\newcommand{\hb}{\mathbf{h}}
\newcommand{\zb}{\mathbf{Z}}

\definecolor{teacher_color1}{HTML}{2A9D8F}
\definecolor{teacher_color2}{HTML}{E9C46A}

\definecolor{blue_color1}{HTML}{5465ff}
\definecolor{blue_color2}{HTML}{788bff}
\definecolor{blue_color3}{HTML}{9bb1ff}
\definecolor{blue_color4}{HTML}{bfd7ff}

\definecolor{orange_color1}{HTML}{E9C46A}
\definecolor{orange_color2}{HTML}{F4A261}
\definecolor{turquoise_color1}{HTML}{2A9D8F}
\definecolor{turquoise_color2}{HTML}{4E8098}

\definecolor{green_color1}{HTML}{245501}
\definecolor{green_color2}{HTML}{73a942}

% \definecolor{shadecolor}{HTML}{F2F2F2}
% \definecolor{shadecolor2}{HTML}{D9D9D9}
% % \definecolor{shadecolor2}{HTML}{D0CECE}
% \definecolor{shadecolor3}{HTML}{b4b4b4}

\colorlet{shadecolor}{gray!10}
\colorlet{shadecolor2}{gray!20}
\colorlet{shadecolor3}{black!80}

\newcommand{\forgetting}[1]{\mathcal{F}_{#1}}
\newcommand{\transfer}[1]{\mathcal{T}_{#1}}

% \begin{figure*}[t!]
%   \centering
%   \includegraphics[width=0.33\linewidth]{{fig1_ts_standard.png}}%
%   \includegraphics[width=0.33\linewidth]{{fig1_ts_continual.png}}%
%   \includegraphics[width=0.33\linewidth]{{fig1_schematic.png}}
%   \caption{\label{fig:figure1} \textbf{Continual learning in the teacher-student
%       setup} \textbf{(a)} In the vanilla teacher-student setup, a ``student''
%     network is trained on i.i.d.~inputs with labels provided by a teacher
%     network. \textbf{(b)} To model continula learning, we train a two-layer
%     neural network~\eqref{eq: forward}, the \emph{student}, on a succession of
%     two tasks, $A$ and $B$. We give her two heads bla bla. Two tasks are
%     modelled by two different teacher networks. \textbf{(c)}~We plot a
%     typical learning curve for the student, where we show her generalisation
%     error~\eqref{eq:eg} during training. The solid line is prediction from our
%     theory; the crosses are obtained through a numerical simulation of a network
%     with input dimension $D=\ldots$. We also show the key quantities of interest
%     in this study, which are the forgetting~\eqref{eq:forgetting} and
%     transfer~\eqref{eq:transfer}. \emph{Parameters:} input dimension, learning
%     rate, teacher-teacher overlap, K, M, etc.}
% \end{figure*}

\input{figures_code/figure_1}

\paragraph{Main contributions} 
\begin{itemize}
\item We analyse continual learning in two-layer neural networks by deriving a closed set of equations which predict the test error of the network trained on a succession of tasks using one-pass (or online) SGD, extending classical work on standard supervised learning by~\textcite{saad1995exact, riegler1995line}.
\item Using these equations, we show that intermediate task similarity leads to greatest forgetting in our model.
\item We disentangle task similarity on the level of features (input-to-hidden weights) and readouts (input-to-hidden weights) and describe the effect of both types of similarity on forgetting and transfer in infinitely wide networks. We find that feature and readout similarity contribute in complex and sometimes non-symmetric ways to a range of forgetting and transfer metrics.
\end{itemize}

We summarise our approach in~\autoref{fig:figure1}. In the classical
teacher-student setup (illustrated in ~\autoref{fig:figure1}a), a ``student'' neural network is trained on synthetic data where inputs $\xb\in\mathbb{R}^D$ are drawn
element-wise i.i.d.~from the normal distribution and labels are generated by a
``teacher'' network~\cite{gardner1989}. To model continual
learning, here we consider a setup with two teachers (denoted by $\dagger$ and
$\ddag$), which correspond to two tasks to be learned in succession. Let $\phi(\xb; \wb, \vb)$ denote the output of a two-layer network with $L$ hidden neurons, first and second layer weights $\wb\in\mathbb{R}^{L\times D}$ and $\vb\in\mathbb{R}^{L}$, and activation $g$ after the hidden layer,~\emph{i.e.}
\begin{equation}
    \phi(\xb; \wb, \vb) = \sum_{l=1}^L {\vb}_l g\left(\frac{{\wb}_l\xb}{\sqrt{D}}\right). \label{eq: forward}
\end{equation}
We generally use $K$ ($M$) for the number of hidden neurons of the student (teacher). In
the first phase of training (left side of~\autoref{fig:figure1}b), labels are
generated by the first teacher via
$y^\dagger=\phi(\xb; \wb^\dagger, \vb^\dagger)$, and student outputs are given
by $\hat{y}^\dagger=\phi(\xb; \wb, \hb^\dagger)$. Training proceeds
using~\gls{sgd} on the squared error of $y^\dagger$, $\hat{y}^\dagger$ in the online regime, where at each step of SGD we draw a new sample $(x, y)$ to evaluate the gradients, until the
task \emph{switch}. We follow a standard \textit{multi-headed} approach to
continual learning~\cite{pmlr-v70-zenke17a, farquhar2018towards}, in which the
student keeps its first-layer weights for the new task, but adds a set of head
weights. Thus in the second phase of training, the error is computed over
$y^\ddag$, $\hat{y}^\ddag$. Retaining both heads allows us to continually
monitor the performance of the student on both tasks after switch, and in theory
permits the student to represent both teachers perfectly if given sufficient
hidden units.

The generalisation error of the student on the
two tasks can be defined as
\begin{multline}
  \label{eq:eg}
  \epsilon^*(\wb, \hb^*, \wb^*, \vb^*)
  \equiv\\
  \frac{1}{2}\left\langle\right.[\phi(\xb; \wb^*, \vb^*) \left.- \phi(\xb; \wb, \hb^*)]^2\right\rangle,
\end{multline}
where $*$ denotes either task $\dagger$ or $\ddag$, and the average $\langle \cdot \rangle$ is taken over the input distribution
$\xb$ for a given set of teacher and student weights. Note in the online SGD setting, there is no distinction between train and test error. We emphasise that the student has the same set of first-layer weights ($\wb$) for
both tasks, but different head weights $\hb^\dagger$, $\hb^\ddag$.

Our main theoretical contribution is a set of dynamical equations that predict
the evolution of the test error~\autoref{eq:eg} during the course of training in the limit
of large input dimension $D\to\infty$ with $K, M\sim O(1)$,
see~\autoref{sec:odes}. We plot the theoretical prediction in~\autoref{fig:figure1}c
together with a single simulation (crosses); even at moderate input size
$D=10^4$, the agreement is good. We observe that
the student error on the first task (green) decreases in the first period of training. After switching
tasks at $\tilde s=5\cdot 10^5$, the error of the student on the second task (yellow) decreases, but the error on the
first task increases. We define \emph{forgetting} and \emph{transfer} at time
$\tilde{s} + t$ as
\begin{align}
  \def\arraystretch{1.5}
  \text{Forgetting:   } \mathcal{F}_t&\equiv  \epsilon^\dagger|_{\tilde{s} + t} -  \epsilon^\dagger|_{\tilde{s}} \label{eq:forgetting},\\
  \text{Transfer:   } \mathcal{T}_t&\equiv \epsilon^\ddag|_{\tilde{s}} - \epsilon^\ddag|_{\tilde{s} + t} \label{eq:transfer}, 
\end{align}
see~\autoref{fig:figure1}c.
An increase in error for the first task after the switch corresponds to positive
forgetting, while a reduction in error for the second task corresponds to
positive transfer. An alternative definition of transfer would compare the performance of the continual learner on task $B$ to the performance of a student that was trained directly on that task. However, this definition introduces additional hyper-parameters which need to be accounted for, such as the distribution of weights at initialisation and at the switching time. Since our focus in this manuscript is on catastrophic forgetting, we focus on the simpler definition of transfer in \eqref{eq:transfer}, and leave an exploration of other transfer measures to future work.

A fundamental question in continual learning is the relationship between forgetting/transfer and the task similarity. While one might expect forgetting to decrease with increasing
task similarity,~\textcite{ramasesh2020anatomy}---through a series of careful experiments on the
CIFAR10 and CIFAR100 datasets---observed  that \emph{intermediate} task similarity leads to greatest
forgetting. We were able to reproduce their results for the two-layer neural networks~\eqref{eq: forward}, see~\autoref{app: deeper}.The primary objective of this work is now to use our multi-teacher-student setup, which gives us full control over teacher similarity, to analyse dependence of forgetting and transfer on task similarity theoretically.

% The student has two sets of hidden to output weights, i.e. two
% heads, one for each teacher; let us denote these by
% $\hb^\dagger\in\mathbb{R}^{M}$ and $\hb^\ddag\in\mathbb{R}^{M}$ such that the
% forward pass of the teacher uses $\hb^\dagger$ when the student is being trained
% on outputs generated by teacher $\dagger$ and $\hb^\ddag$ when trained on
% outputs generated by teacher $\ddag$.  In the classical teacher student
% $\phi(\xb; \wb, \vb)$, with weights $\wb$ and $\vb$ and activation $g$ after the
% hidden layer,
% \begin{equation}
%   \label{eq: forward}
%     \phi(\xb; \wb, \vb) = \sum_{m=1}^M {\vb}_m g\left(\frac{{\wb}_m\xb}{\sqrt{D}}\right),
% \end{equation}
% where $M$ is the number of neurons in the network. The key idea of the
% teacher-student framework is to train the network on a synthetic data set, where
% inputs $\xb\in\mathbb{R}^D$ are drawn element-wise i.i.d.~ from the normal
% distribution, while labels $y$ are obtained from a random, but fixed neural
% network called the \emph{teacher}. % Training data is given by
% % $\mathcal{D}=\langle\xb^\mu, y^\mu\rangle_{\mu=1}^T$, where $T$ is the number of
% % training steps taken.

 % Let $D$ be the input dimension to the networks, $K$ be the
% hidden dimension of the teachers, and $M$ be the hidden dimension of the
% student.  

\subsection{Further Related Work} 

The \textbf{teacher-student framework} has a long
history in studying the dynamics of learning in neural network
models~\cite{gardner1989, seung1992statistical, watkin1993,
  engel2001statistical, zdeborova2016statistical} and has recently experienced a
surge of activity in the machine learning community~\cite{zimmer2014teacher,
  zhong2017recovery, tian2017analytical, du2018gradient,
  soltanolkotabi2018theoretical, aubin2018committee, saxe2018information,
  Baity-Jesi2018, goldt2019dynamics, ghorbani2019limitations,
  yoshida2019datadependence, ndirango2019generalization, gabrie2020meanfield,
  bahri2020statistical, zdeborova2020understanding, advani2020highdimensional}. While this article went to press, a preprint by \textcite{asanuma2021statistical} appeared which analyses continual learning in a teacher-student setup for linear regression.
  
The teacher-student approach has recently been used to study \textbf{transfer learning}, both in linear networks~\cite{lampinen2018analytic} and in non-linear perceptron models~\cite{dhifallah2021phase}, which correspond to the $K=M=1$ case of our setup. While the transfer of knowledge from one task to the next is an important aspect in continual learning, the latter is crucially also interested in the retention--or forgetting--of knowledge about the first task. This can be most clearly seen in the fact that in transfer learning, there is only one set of student head weights. Indeed, we will find an interesting interplay between transfer and forgetting in our models of continual learning.

\textbf{Continual learning in the NTK regime}\hspace*{1em} \citet{doan2020theoretical}
analysed the impact of task similarity, and also found increasing task
similarity leads to more forgetting. The key difference to our work is that
their study focuses on the neural tangent kernel (NTK)~\cite{jacot2018neural} or ``lazy''
regime~\cite{chizat2019lazy} of two-layer networks, where the first layer of
weights stays approximately constant throughout
training. \textcite{bennani2020generalisation} gave guarantees on the error achieved with orthogonal
gradient descent in the same regime. Here, we focus on the regime where the
weights of the network move significantly and are thus able to learn features,
which will be key to our analysis in~\autoref{sec:odes} and to our
disentangling of feature vs.~readout similarity in~\autoref{sec:read_out}.

The \textbf{dynamics of two-layer neural networks} trained using online SGD in
the classic teacher-student setup of~\autoref{fig:figure1}a was first studied
in a series of classic papers by \textcite{Biehl1995} and~\textcite{saad1995exact}
who derived a set of closed ODEs that track the test error of the student (see
also~\textcite{saad1995line, Biehl1996, saad2009line} for further results
and~\textcite{goldt2019dynamics} for a recent proof of these equations).  Here,
we extend this type of analysis to the continual learning model
of~\autoref{fig:figure1}b. The aforementioned works all consider the limit of
large input dimension $D\to\infty$, while the number of neurons is of
order~1. The complementary ``mean-field'' limit of finite input dimension and an
infinite number of hidden neurons was analysed~\cite{mei2018mean, chizat2018,
  sirignano2020mean, rotskoff2018parameters}. We will turn to this limit to
disentangle the impact of feature and readout similarity
in~\autoref{sec:read_out}.

Many methods for \textbf{combating catastrophic interference} have been
proposed, often taking the form of regularisation, architecture expansion,
and/or replay~\cite{parisi2019continual,
  farquhar2018towards}. Regularisation-based methods constrain weights to retain
information about earlier tasks~\cite{pmlr-v70-zenke17a, li2017learning,
  kirkpatrick2017overcoming}; architectural methods add capacity to the network
for each new task~\cite{rusu2016progressive}; and replay methods store data from
earlier tasks to interleave when learning new tasks~\cite{McClelland1995,
  shin2017continual}.

% Theoretically, the key advantage of the student-teacher framework, as compared
% to standard training on a database of examples, is that it offers direct access
% to the ``ground truth'' generative model of the data. This access in turn
% provides an understanding of various aspects of the problem including the
% representational capacity necessary for the function to be learned; a notion of
% `distance' between the model being trained and the target function; and the
% ability to exactly evaluate quantities like the generalisation error.
% Crucially, the student-teacher setting and the theoretical framework we
% construct to describe learning (see~\autoref{sec:theory}) makes the problem
% analytically tractable in a way that is not possible with real-world datasets
% such as ImageNet.

% \input{figures_code/figure_1_final}
% \input{figures_code/student_teacher_combined}
% \input{figures_code/vanilla_student_teacher}
% \input{figures_code/multi_student_teacher}

\vspace*{2em} 

\section{Continual learning in the large input limit}
\label{sec:odes}

We begin by studying the impact of task similarity on the dynamics and the performance  of learning in the limit of large input dimension $D~\to~\infty$, while the number of neurons $K, M \sim O(1)$. 

\paragraph{Training} We train the student using online stochastic gradient descent on the
$L_2$ loss. Each new input $\xb$ is fed to the teacher to compute the target
output via $y^* = \phi(\xb;\wb^*,\vb^*)$, while the student prediction is given by
$\hat{y}^\dagger = \phi(\xb;\wb,\hb^*)$. The student's weights in both layers are
updated through gradient descent on $\frac{1}{2}(\hat{y}^* - y^*)^2.$ The~\gls{sgd}
weight updates are given by:
\begin{subequations}
    \label{eq:sgd}
    \begin{align}
    	\wvecb_k^{\mu+1} &= \wvecb_k^{\mu} - \frac{\alpha_\wb}{\sqrt{D}}v_k^{*\mu}g'(\lambda_k^\mu)\Delta^{*\mu} \xb^\mu \label{eq: wupdate} \\
	    h_k^{*\mu+1} &= h_k^{*\mu} -
                       \frac{\alpha_\hb}{D}g(\lambda_k^\mu)\Delta^{*\mu}, \label{eq:
                       vupdate}
    \end{align}
\end{subequations}
where $\alpha_\wb$ is the learning rate for the feature weights, $\alpha_\hb$ is the learning rate for the head weights, and
\begin{align}
	\Delta^{\dagger\mu} &\equiv \sum_k h_k^{\dagger\mu}g(\lambda_k^\mu) - \sum_m v_m^\dagger g(\rho_m^\mu);\\
	\Delta^{\ddag\mu} &\equiv \sum_k h_k^{\ddag\mu}g(\lambda_k^\mu) - \sum_p v_p^\ddag g(\eta_p^\mu).
\end{align}
We have also introduced the \emph{local fields} 
\begin{equation}
    \rho_m \equiv \frac{\wvecb_m\xb}{\sqrt{D}}, \qquad \eta_p \equiv \frac{\wvecb_p\xb}{\sqrt{D}}, \qquad \lambda_k \equiv \frac{\wvecb_k\xb}{\sqrt{D}}
\end{equation}
of the $m^{\text{th}}$ teacher $\dagger$ unit, $n^{\text{th}}$ teacher $\ddag$ unit, and $k^{\text{th}}$ student unit, respectively. In general, 
indices $i, j, k, l$ are used for hidden units of the student; $m, n$ for hidden units of $\dagger$; and $p, q$ for hidden units of $\ddag$. 
Initial weights are taken i.i.d.~from the normal distribution with standard deviation $\sigma_0$. The
different scaling of the learning rates for first and second-layer
weights guarantees the existence of a well-defined limit of the SGD
dynamics as $D\to\infty$. We make the crucial assumption that at each step
of the algorithm, we use a previously unseen sample $(\xb, y^*)$. This limit of
infinite training data is variously known as online learning or
one-shot/single-pass SGD. We note that in general the head
weights could also be matrices if a teacher has multiple output nodes, but we focus on the case of a single output here to keep notation light.

\paragraph{The ``order parameters'' of the problem} The key quantity in our analysis is the test error~\autoref{eq:eg}, which (e.g. for $\dagger$) can be written more explicitly as
\begin{multline}
    \label{eq:eg_explicit}
	\epsilon^\dagger(\wb, \wb^\dagger, \hb^\dagger, \vb^\dagger) =\\
	\frac{1}{2}\left\langle\left[\sum_{k=1}^K h_k^\dagger g(\lambda_k) - \sum_{m=1}^Mv_m^\dagger g(\rho_m)\right]^2\right\rangle.
\end{multline}
To evaluate the average, the input~$\xb$ only appears via products with the student weights $(\lambda_k)$ and likewise for the teacher; we can
hence replace the high-dimensional averages over $\xb$ with an average over the
$K+M$ ``local fields''~$\lambda$ and $\rho$. Since we take the inputs element-wise i.i.d.~from the standard Gaussian distribution, we have $\langle x_i \rangle=0$ and $\langle x_i x_j \rangle = \delta_{ij}$. It also follows immediately that the local fields are jointly Gaussian, with mean $\langle\lambda_k\rangle=\langle\rho_m\rangle=0$. The test error can hence be written as a function of only the second moments of the joint distribution of $(\rho, \lambda)$, which we define as
\begin{gather}
   \label{eq:op}
    q_{kl} \equiv \langle\lambda_k\lambda_l\rangle, \quad r_{km}  \equiv \langle\lambda_k\rho_m\rangle, \quad t_{mn} \equiv \langle\rho_m\rho_n\rangle;
\end{gather}
and the second-layer weights of the students. In other words, asymptotically
\begin{equation}
  \label{eq:eg_order}
  \lim_{D\to\infty}	\epsilon^\dagger(\wb, \wb^\dagger, \hb^\dagger, \vb^\dagger) = \epsilon^\dagger(\mathbf{Q}, \mathbf{R}, \mathbf{T}, \hb^\dagger, \vb^\dagger).
\end{equation}
where $\mathbf{Q}=(q_{kl})$, etc. Note there is an equivalent formulation for
$\ddag$ with the $\eta$ local field and relevant second-layer weights.  These
overlap matrices, or ``order parameters'' in statistical physics jargon, have a
clear physical interpretation, which can be seen when evaluating the averages
explicitly. The so-called teacher-student overlap, $r_{km}$ for example:
\begin{equation}
  \label{eq:r_explicit}
    r_{km}  \equiv \langle \lambda_k\rho_m \rangle = \frac{\wvecb_k \wvecb_m^\dagger}{D},
\end{equation}
quantifies the overlap or similarity between the weights of the $k^\text{th}$ hidden unit of the student and the $m^\text{th}$ hidden unit of the
teacher. Similarly, $q_{kl}$ gives the self-overlap of the $k$th and $l$th student nodes, and $t_{mn}$ gives the (static) self-overlap of teacher nodes.

\paragraph{Task similarity} The teacher-student setup gives us precise control over the task similarity via the overlap between the first-layer weights of different teachers,
\begin{equation}
    \label{eq: v_overlap}
    v_{mp} \equiv \langle\rho_m\eta_p\rangle = \frac{1}{D}\wvecb^\dagger_m\wvecb^\ddag_p,
\end{equation}
which we can tune to observe its effects on the dynamics of continual learning.

\subsection{Results}
\label{sec:ode_results}

\begin{figure}[!t]
% \begin{mdframed}[backgroundcolor=shadecolor,linecolor=shadecolor]
% \begin{mdframed}[backgroundcolor=shadecolor2,linecolor=shadecolor2]
% \centering
% \hspace{-1em}
    \pgfplotsset{
		width=0.48\textwidth,
		height=0.42\textwidth,
		xtick={0, 200000, 400000, 600000, 800000},
		xlabel={$s$},
		}
    \begin{tikzpicture}
        \fill[shadecolor2, opacity=0] (0, 0) rectangle (0.5\textwidth, 0.4\textwidth);
        \node [anchor=north west] at (0, 0.4\textwidth) {\emph{(a)}};
		\begin{axis}
			[
			at={(1.3cm, 1.1cm)},
			scaled x ticks = false,
			xmin=0, xmax=1000000,
			ymin=-4.5, ymax=-0.5,
			ylabel={$\log{\epsilon}$},
			y label style={at={(axis description cs:-0.1, 0.5)}, rotate=0, anchor=south},
		]
		\addplot graphics [xmin=0, xmax=1000000,ymin=-4.5,ymax=-0.5] {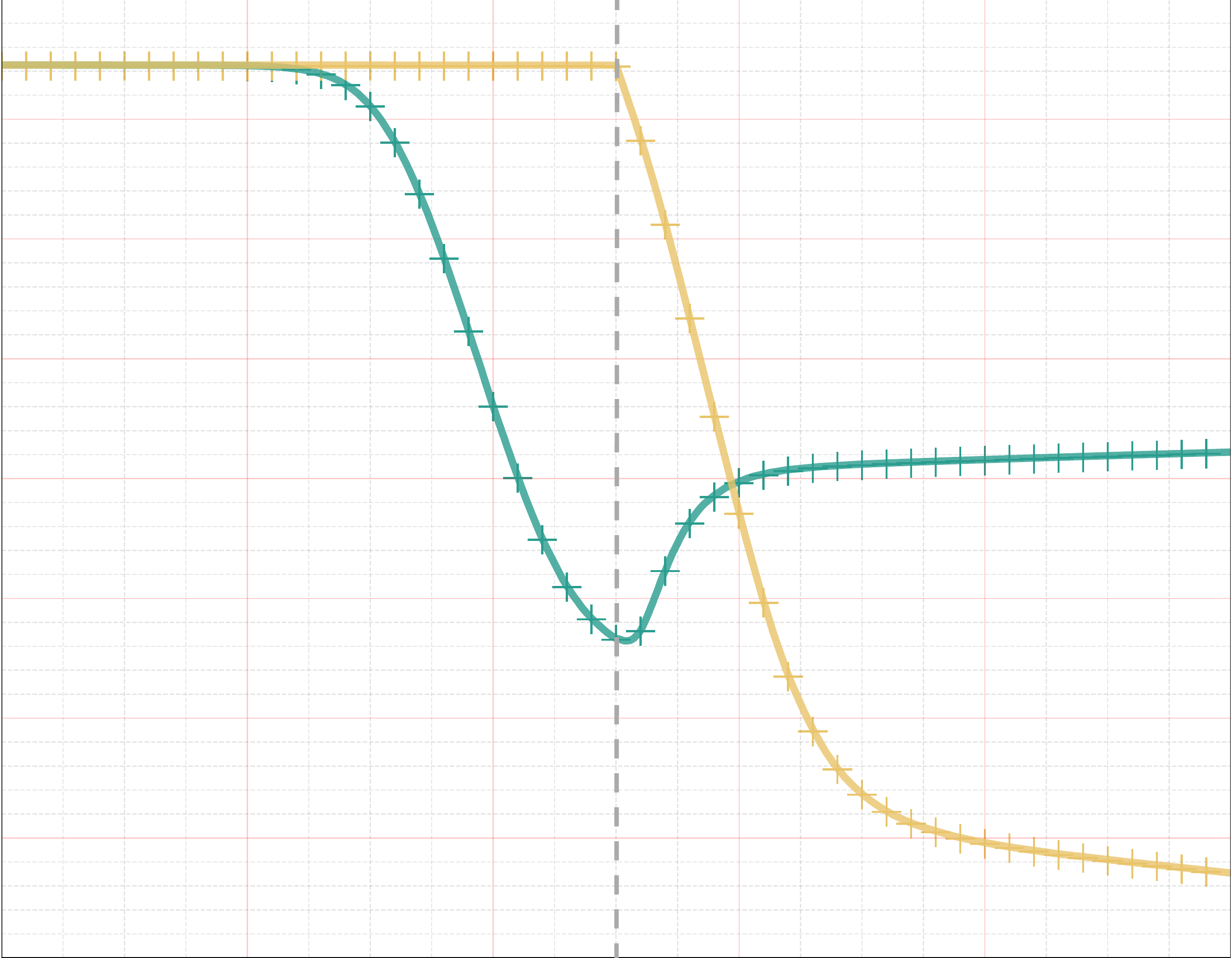};
		\end{axis}
% 		\node [text=shadecolor3] at (-1.2, 4.4) {\emph{(a)}};
% % 		\filldraw[draw=black,fill=white] (0.3, 0.3) rectangle (4, 2.8);
		\node at (3.5, 1.5) {\small $\epsilon^\dagger$ simulation};
		\node [text=teacher_color1] at (2, 1.5) {\tiny + + + + +};
	   % \draw [dashed, teacher_color1, line width=0.5mm] (0.4, 5.3) -- (1, 5.3);
	    \node at (3.2, 3) {\small $\epsilon^\dagger$ ODE};
	    \draw [teacher_color1, line width=0.5mm] (1.6, 3) -- (2.4, 3);
	    \node at (3.5, 2) {\small $\epsilon^\ddag$ simulation};
	   % \draw [dashed, teacher_color2, line width=0.5mm] (0.4, 4.8) -- (1, 4.8);
	   \node [text=teacher_color2, font=\bf] at (2, 2) {\tiny + + + + +};
	    \node at (3.2, 2.5) {\small $\epsilon^\ddag$ ODE};
	    \draw [teacher_color2, line width=0.5mm] (1.6, 2.5) -- (2.4, 2.5);
	    \node at (0.28\textwidth, 1.5) {\tiny{$\tilde{s}$}};
	\end{tikzpicture}%
% 	\end{mdframed}
% 	\vspace{0.1em}
	\pgfplotsset{
		width=0.27\textwidth,
		height=0.28\textwidth,
		scaled x ticks=false,
		xlabel={\small $s$},
		xtick={0,400000,800000},
		every tick label/.append style={font=\tiny},
		y label style={at={(axis description cs:-0.1, 0.5)}, rotate=0, anchor=south},
		x label style={at={(axis description cs:0.5, -0.23)}, rotate=0, anchor=south},
		}
	\begin{tikzpicture}
% 		\node at (2.5, 3.8) {\small ReLU};
        \fill[shadecolor2, opacity=0] (0, 0) rectangle (0.25\textwidth, 0.25\textwidth);
        \node [anchor=north west] at (0, 0.25\textwidth) {\emph{(b)}};
%         \node [text=shadecolor3] at (-1.1, 2.1) {\emph{(b)}};
		\begin{axis}
			[
			at={(0.8cm, 0.75cm)},
			anchor=south west,
			xmin=0, xmax=1000000,
			ymin=0, ymax=1.2,
			ylabel={\tiny $\mathbf{Q}$},
		]
		\addplot graphics [xmin=0, xmax=1000000,ymin=0,ymax=1.2] {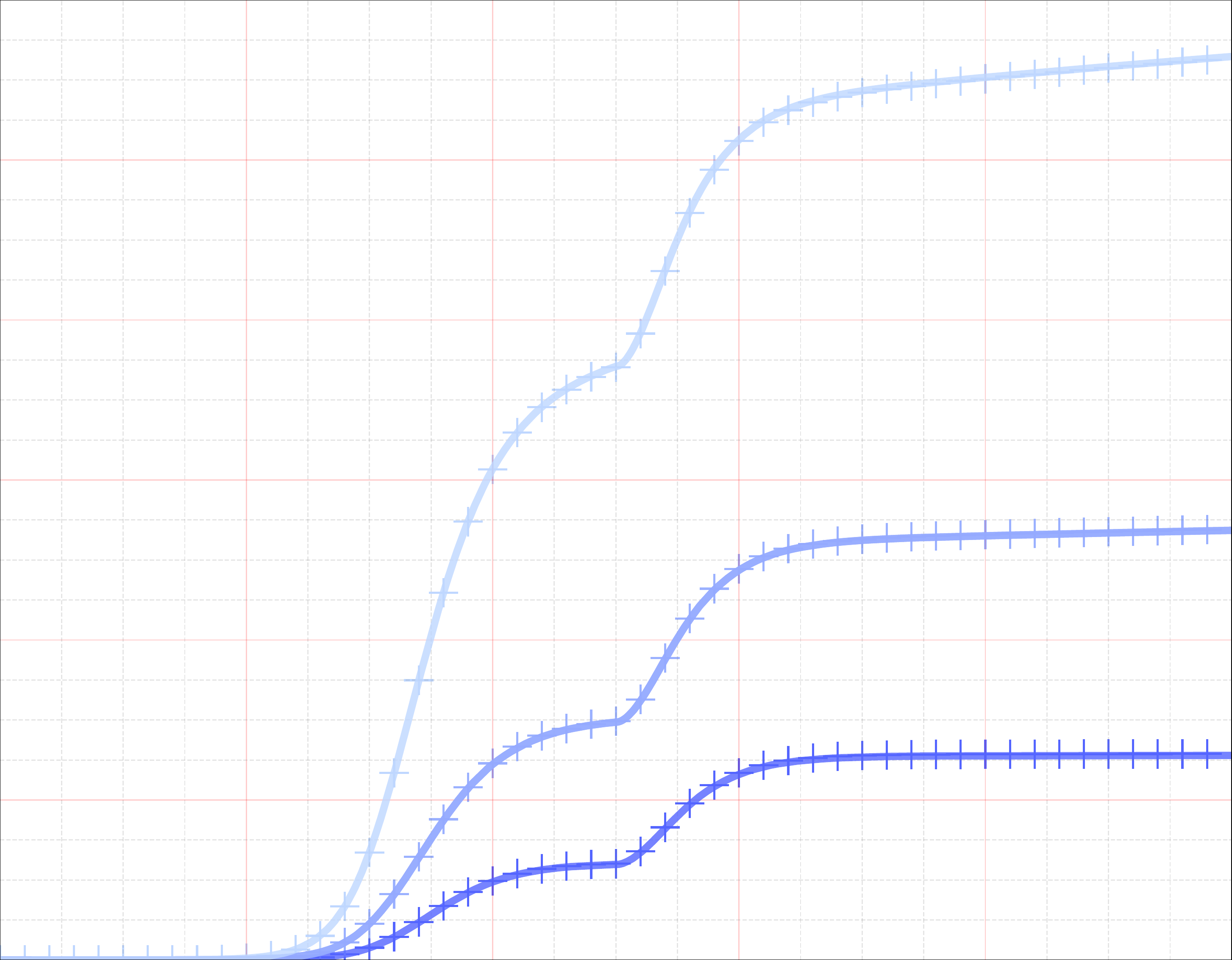};
		\end{axis}
% % 		\filldraw[draw=black,fill=white] (0.15, 2.1) rectangle (1.5, 3.6);
		\node at (1.7, 3.6) {\tiny $q_{00}$};
		\node [text=blue_color1] at (1.2, 3.55) {\tiny + + +};
	    \draw [blue_color1, line width=0.5mm] (1, 3.65) -- (1.45, 3.65);
	    \node at (1.7, 3.3) {\tiny $q_{01}$};
	    \node [text=blue_color2] at (1.2, 3.25) {\tiny + + +};
	    \draw [blue_color2, line width=0.5mm] (1, 3.35) -- (1.45, 3.35);
	    \node at (1.7, 3) {\tiny $q_{10}$};
	    \node [text=blue_color3] at (1.2, 2.95) {\tiny + + +};
	    \draw [blue_color3, line width=0.5mm] (1, 3.05) -- (1.45, 3.05);
	    \node at (1.7, 2.7) {\tiny $q_{11}$};
	    \node [text=blue_color4] at (1.2, 2.65) {\tiny + + +};
	    \draw [blue_color4, line width=0.5mm] (1, 2.75) -- (1.45, 2.75);
	\end{tikzpicture}%
% 		\hspace{0.1em}
	\begin{tikzpicture}
% 		\node at (2.5, 3.8) {\small Scaled Error Function};
        \fill[shadecolor2, opacity=0] (0, 0) rectangle (0.25\textwidth, 0.25\textwidth);
        \node [anchor=north west] at (0, 0.25\textwidth) {\emph{(c)}};
%         \node [text=shadecolor3] at (-1.1, 2.1) {\emph{(c)}};
		\begin{axis}
			[
			at={(0.8cm, 0.75cm)},
			anchor=south west,
			xmin=0, xmax=1000000,
			ytick={-1, -0.5, 0, 0.5},
			ymin=-1, ymax=1.0,
			ylabel={\tiny $\hb$}
		]
		\addplot graphics [xmin=0, xmax=1000000,ymin=-1,ymax=1] {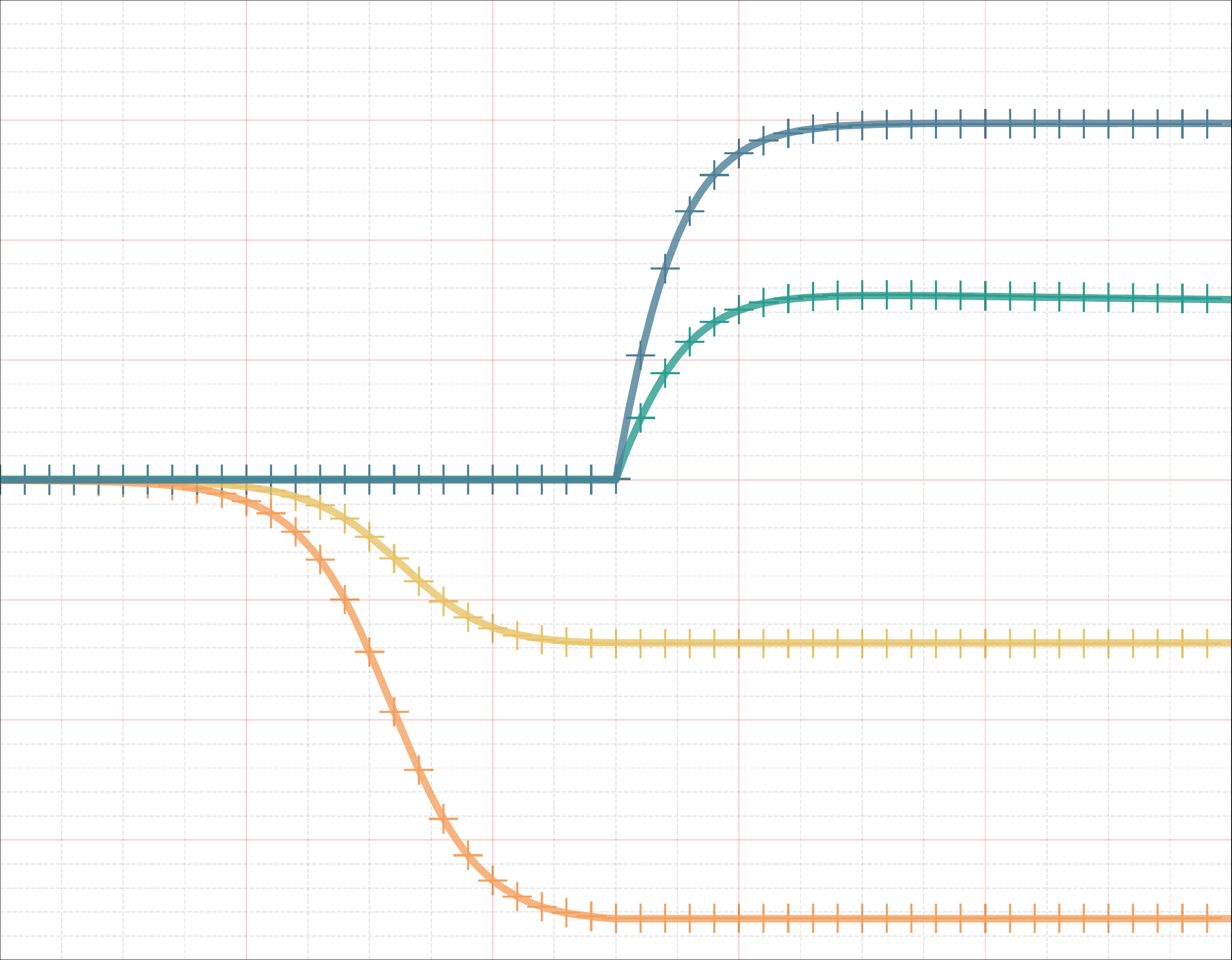};
		\end{axis}
% % 		\filldraw[draw=black,fill=white] (0.15, 1.6) rectangle (1.3, 3.6);
		\node at (1.7, 3.7) {\tiny $h^\dagger_0$};
		\node [text=orange_color1] at (1.2, 3.65) {\tiny + + +};
	    \draw [orange_color1, line width=0.5mm] (1, 3.75) -- (1.45, 3.75);
	    \node at (1.7, 3.35) {\tiny $h^\dagger_1$};
	    \node [text=orange_color2] at (1.2, 3.3) {\tiny + + +};
	    \draw [orange_color2, line width=0.5mm] (1, 3.4) -- (1.45, 3.4);
	    \node at (1.7, 3) {\tiny $h^\ddag_0$};
	    \node [text=turquoise_color1] at (1.2, 2.95)  {\tiny + + +};
	    \draw [turquoise_color1, line width=0.5mm] (1, 3.05) -- (1.45, 3.05);
	    \node at (1.7, 2.65) {\tiny $h^\ddag_1$};
	    \node [text=turquoise_color2] at (1.2, 2.6) {\tiny + + +};
	    \draw [turquoise_color2, line width=0.5mm] (1, 2.7) -- (1.45, 2.7);
	   % \node at (1.6, 3) {\tiny $\tanh(x)$};
	   % \draw [dashed, red] (0.3, 3) -- (1, 3);
	\end{tikzpicture}%
	\vspace{0.05em}
	\begin{tikzpicture}
% 		\node at (2.5, 3.8) {\small Sigmoid};
        \fill[shadecolor2, opacity=0] (0, 0) rectangle (0.25\textwidth, 0.25\textwidth);
        \node [anchor=north west] at (0, 0.25\textwidth) {\emph{(d)}};
%         \node [text=shadecolor3] at (-1.1, 1.9) {\emph{(d)}};
		\begin{axis}
			[
			at={(0.8cm, 0.75cm)},
			anchor=south west,
			xmin=0, xmax=1000000,
			ytick={-0.8, -0.4},
			ymin=-1.2, ymax=0.05,
			ylabel={\tiny $\mathbf{R}$}
		]
		\addplot graphics [xmin=0, xmax=1000000,ymin=-1.2,ymax=0.05] {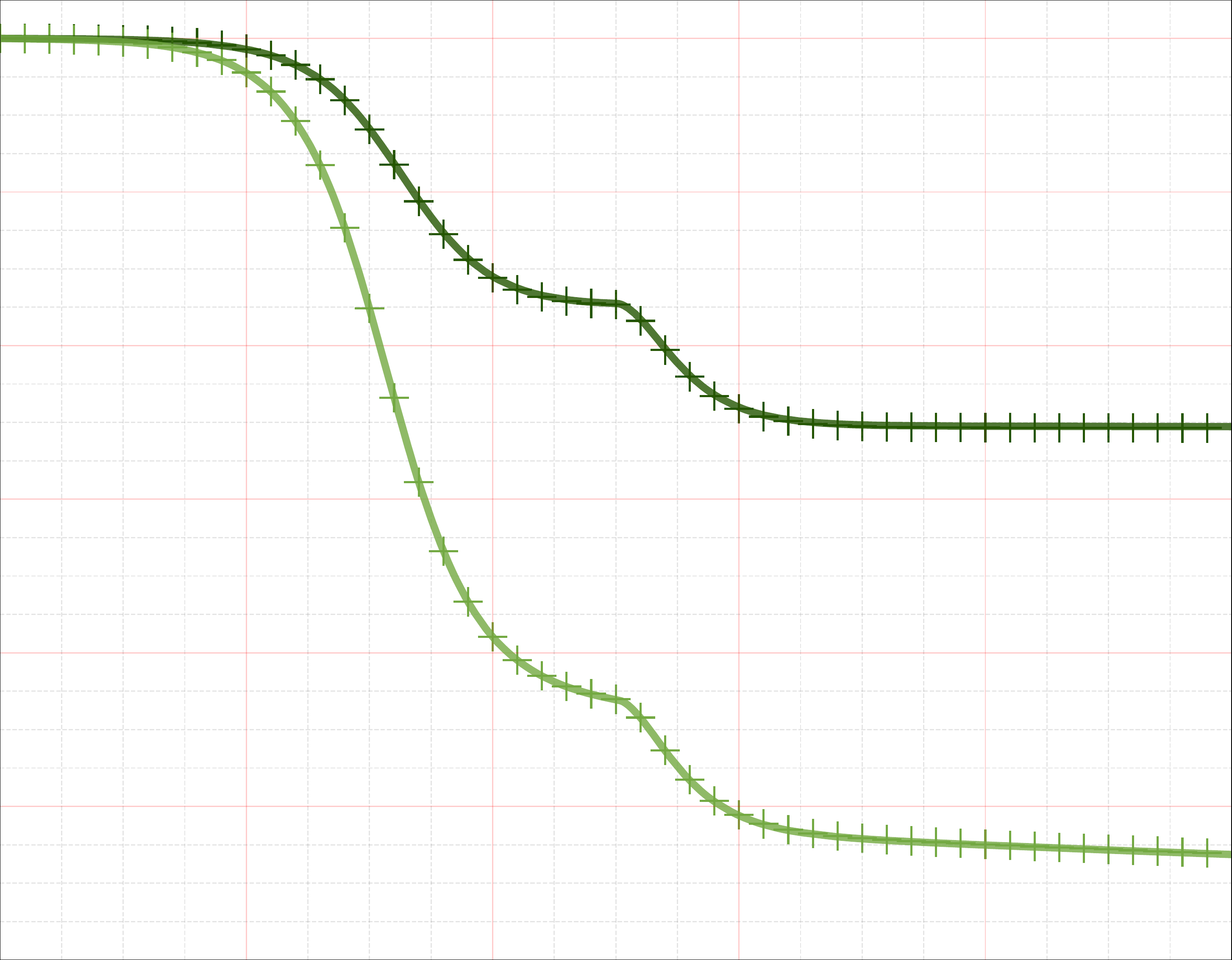};
		\end{axis}
% % 		\filldraw[draw=black,fill=white] (0.15, 0.15) rectangle (1.5, 1.05);
		\node at (1.7, 1.3) {\tiny $r_{00}$};
		\node [text=green_color1] at (1.2, 1.25) {\tiny + + +};
	    \draw [green_color1, line width=0.5mm] (1, 1.35) -- (1.45, 1.35);
	    \node at (1.7, 1) {\tiny $r_{10}$};
	    \node [text=green_color2] at (1.2, 0.95) {\tiny + + +};
	    \draw [green_color2, line width=0.5mm] (1, 1.05) -- (1.45, 1.05);
	\end{tikzpicture}%
    % \hspace{0.1em}
    \begin{tikzpicture}
% 		\node at (2.5, 3.8) {\small Linear};
        \fill[shadecolor2, opacity=0] (0, 0) rectangle (0.25\textwidth, 0.25\textwidth);
        \node [anchor=north west] at (0, 0.25\textwidth) {\emph{(e)}};
%         \node [text=shadecolor3] at (-1.1, 1.9) {\emph{(e)}};
		\begin{axis}
			[
			at={(0.8cm, 0.75cm)},
			anchor=south west,
			xmin=0, xmax=1000000,
			ytick={-0.8, -0.4},
			ymin=-1.2, ymax=0.05,
			ylabel={\tiny $\mathbf{U}$}
		]
		\addplot graphics [xmin=0, xmax=1000000,ymin=-1.2,ymax=0.05] {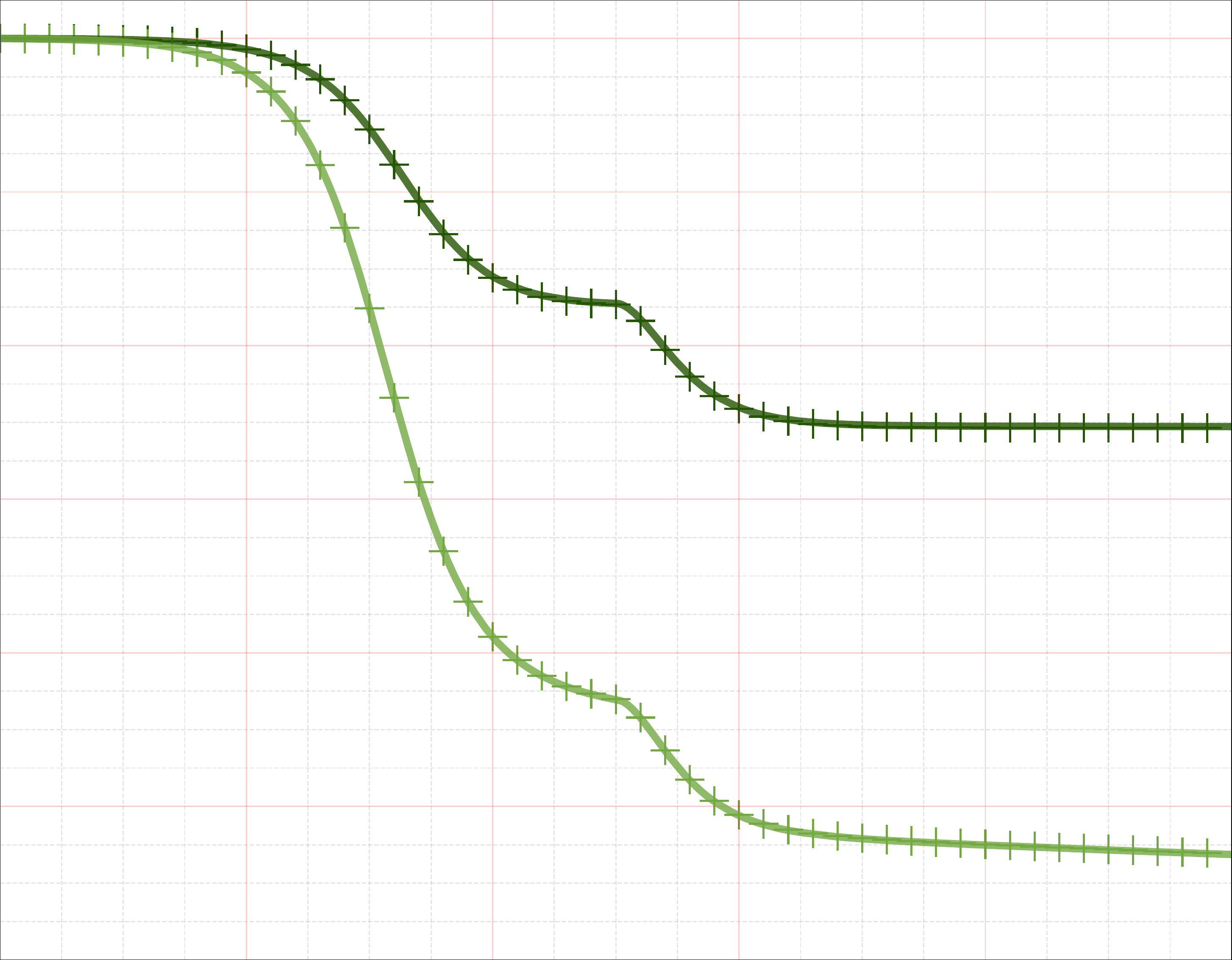};
		\end{axis}
% % 		\filldraw[draw=black,fill=white] (0.15, 0.15) rectangle (1.5, 1.05);
		\node at (1.7, 1.3) {\tiny $u_{00}$};
		\node [text=green_color1] at (1.2, 1.25) {\tiny + + +};
	    \draw [green_color1, line width=0.5mm] (1, 1.35) -- (1.45, 1.35);
	    \node at (1.7, 1) {\tiny $u_{10}$};
	    \node [text=green_color2] at (1.2, 0.95) {\tiny + + +};
	    \draw [green_color2, line width=0.5mm] (1, 1.05) -- (1.45, 1.05);
	\end{tikzpicture}%
% 	\end{mdframed}
	\caption[\gls{ode} verification for overlap matrices and generalisation
    errors.]{\textbf{Asymptotic theory matches finite-sized simulations.}\label{fig:overlay_plot_basic} Plots of progression of generalisation error (\emph{a}) and order parameters (\emph{b-e})
      for both neural network simulations (crosses) and \gls{ode}
      solutions (solid lines). In this example, the teachers are fully overlapping as evidenced by the identical trajectories of $\mathbf{U}$ and $\mathbf{R}$, the student-teacher$^\dagger$ and student-teacher$^\ddag$ overlaps respectively. There is
      a teacher switch at step 25,000. There is close agreement between the
      simulation and differential equations (to within the expected $1/\sqrt{N}$
      deviation).}
\end{figure}

\input{figures_code/forgetting_transfer_combined}

\subsubsection{An asymptotically exact theory for the dynamics of continual learning} 

The test
error~\autoref{eq:eg} can be written in terms of the order
parameters~\autoref{eq:op}, so to compute the test error at all times we need to
describe the evolution of $\mathbf{Q}$ etc.~during training of the student using
SGD~\autoref{eq:sgd}. Such equations were derived in the vanilla teacher-student
setup by~\textcite{saad1995exact, riegler1995line}, and here we extend this approach to
our continual learning model. We illustrate their derivation for the
teacher-student overlap $r_{km}$~\autoref{eq:r_explicit}. Taking the inner product
of~\autoref{eq: wupdate} (in the case of $*=\dagger$) with $\wvecb_n^\dagger$
and substituting the SGD update~\autoref{eq: wupdate} yields
\begin{align}
  \mathrm{d}r_{km}^{\mu} \equiv r_{km}^{\mu+1} - r_{km}^\mu & = \frac{\wvecb_k^{\mu+1}\wvecb_m^\dagger}{D} - \frac{\wvecb_k^\mu \wvecb_m^\dagger }{D}\\
                                                       &= -\frac{\alpha_\wb}{D} h_k^{\dagger\mu} g'(\lambda_k^\mu)\Delta^{\dagger\mu} \rho_m^\mu~\label{eq:dr}
\end{align}
In the thermodynamic limit $D\to\infty$,
the normalised number of steps $\tau \equiv \mu/D$ becomes a continuous,
time-like variable and we can write:
\begin{equation}
  \frac{\mathrm{d}r_{kn}}{\mathrm{d}\tau} = -\alpha_\wb h_k^\dagger \langle g'(\lambda_k)\Delta^\dagger \rho_n\rangle.
\end{equation}
The remaining averages like $\langle g'(\lambda_k)\lambda_\ell g(\rho_n)\rangle$
are simple three-dimensional integrals over the Gaussian random variables
$(\lambda_k, \lambda_\ell,\rho_m)$ and can be evaluated analytically for
$g(x)=\erf(x/\sqrt{2})$ and for linear networks with $g(x)=x$. Furthermore,
these averages can be expressed only in term of the order parameters, and so the
equations close.  The ODEs for $\mathbf{Q}$~(\autoref{eq: odeq}),
$\mathbf{U}$~(\autoref{eq: uipexp}), as well as the student head weights,
$\hb^\dagger$, and $\hb^\ddag$~(\autoref{eq: hstar_exp}), are given
in~\autoref{app: derivation}.

In~\autoref{fig:overlay_plot_basic}, we show test errors and order parameters
obtained from numerically solving the ODEs (lines) and from a single simulation
(crosses). We find that the agreement between theory and simulations is very
good both for the test error and on the level of individual order parameters
even at intermediate system size ($D=10^4$).

\subsubsection{Impact of task similarity}\label{sec: feature_sim_exp}

We integrated the ODEs in the
simplest possible case to analyse the impact of task similarity. A student with $K=2$ neurons is trained on two teachers
with $M=1$ neuron each, all having sigmoidal activation
$g(x)=\erf(x/\sqrt{2})$\footnote{Code for all experiments and ODE simulations
  can be found at
  \href{https://github.com/seblee97/student_teacher_catastrophic}{https://github.com/seblee97/student\_teacher\_catastrophic}}. A subset of the experiments was also carried out on \gls{relu} networks (purely with network simulations) with broadly similar result; details are discussed in~\autoref{app: relu_networks}. For $M=1$, the task similarity $\mathbf{V}$~(\autoref{eq: v_overlap}) becomes a scalar quantity that we denote $V$, which is the cosine angle between the teachers' input-to-hidden weights. We parametrically generate teachers with specified similarities using the procedure described in \autoref{app: overlap_generation}. The teacher head weights are $\pm{1}$ and the input-hidden weights are normalised. For odd activation functions like the scaled error function, the sign of the input-to-hidden weights can be compensated for by the readout weights so it is sufficient to show results for $V\in[0,1]$. Note that the student has enough capacity to learn both teachers. \autoref{fig: empirical_task_sim}a shows
the generalisation error of the student on the first task, which decays exponentially after an initial
period of stationary error. After the switch at $\tilde s = 1\times 10^6$, the learning curves separate depending on the task similarity. 

We plot the
\textbf{forgetting}~\autoref{eq:forgetting} at different times after the switch vs.~$V$ in panel c. For teachers with
orthogonal first-layer weights ($V=0$), performance on the first task degrades after an initial period of no forgetting. For identical first-layer
weights ($V=1$), the initial rate of forgetting is large, but the student 
recovers and the error on the first task plateaus at a relatively low value.  In
both cases, forgetting is small compared to teachers with intermediate
correlations. Our model thus reproduces the empirical findings on deep networks
for image classification from~\citet{ramasesh2020anatomy}. We hypothesise that
while it is intuitive that similar teachers lead to small forgetting, orthogonal
teachers can be more easily separated by the student by specialising to the
respective teacher units. This separation is made harder by correlations between
the weights of the teachers, akin to the problem of source separation in signal
processing.

% Further points to make: (1) In particular, it may seem natural to expect fast
% initial forgetting rates to result in large maximum forgetting and large
% long-time forgetting. But in fact, we show that these three variables can
% behave differently. We note that prior analysis work has focused mainly on
% this rate~\cite{ramasesh2020anatomy}. (2) We find that the initial rate of
% forgetting is highly correlated with the initial rate of transfer. Forget
% faster, transfer faster. (3) However the final amount of transfer montonically
% increases with greater task similarity whereas the final amount of forgetting
% exhibits similar non-monotocity to the results found
% by~\cite{ramasesh2020anatomy} with greatest forgetting for intermediate task
% similarity.

To analyse \textbf{transfer}, we look at the generalisation on the second task $\ddag$ after the switch~\autoref{fig:
  empirical_task_sim}b. Just after the switch, higher overlap allows faster transfer. All students then reach a second plateau. Only students trained on tasks that are close to orthogonal break away form this second plateau and achieve an exponentially decaying generalisation error, whereas the other  students remain stuck. This is a remarkable result, since the same student starting from random initial weights would converge to the second teacher without problem. Indeed, for orthogonal teachers, a student trained to convergence on
the first task will have the equivalent of random initial weights for the
second task (up to the scaling of the
networks), explaining its better performance. Students trained on correlated tasks also converge to the teachers of the first, but this correlated initialisation leads to a loss of performance on the second task. We thus find that task similarity aids
short-term transfer but harms long-term transfer in this setting.

\section{Disentangling Feature and Read-out Similarity}
\label{sec:read_out}

\begin{figure}[!ht]
\centering
\begin{tikzpicture}
\node [opacity=1] at (0, 0) {\includegraphics[width=0.45\textwidth]{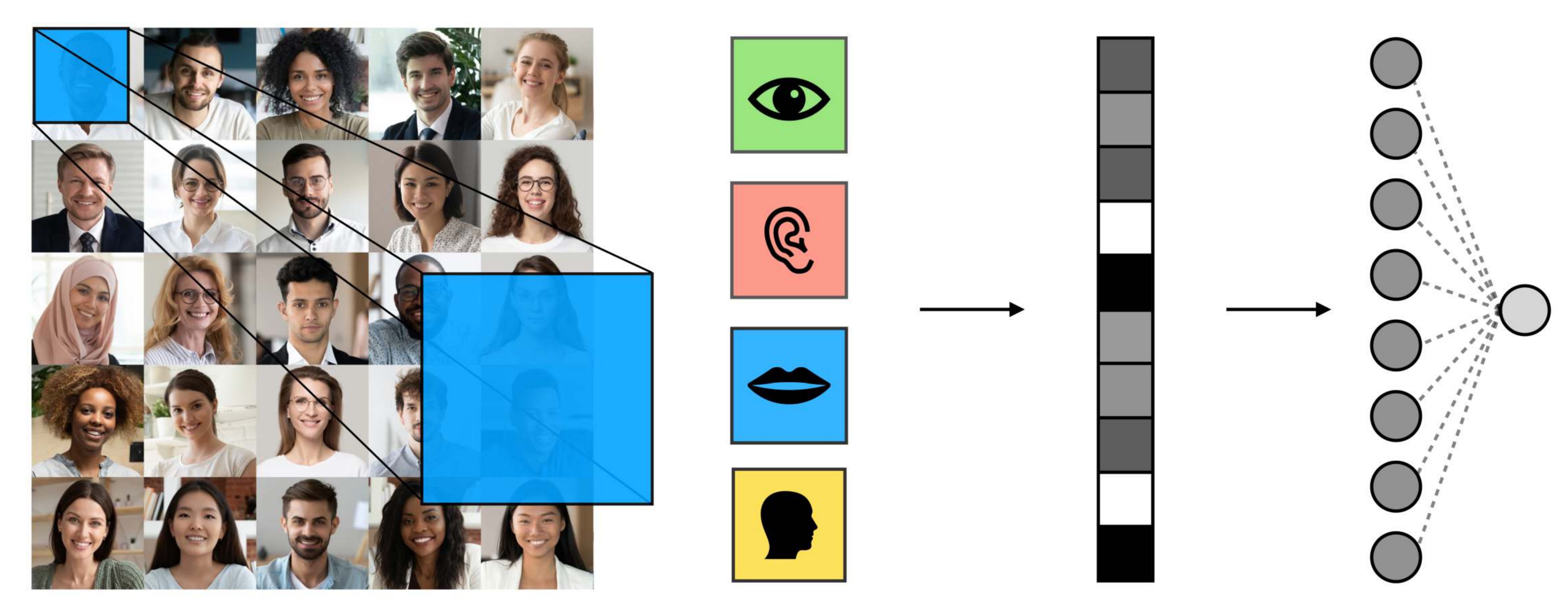}}; 
\node[align=center] at (-2.5, 1.6) {\scriptsize Input};
\node[align=center] at (0, 1.9) {\scriptsize Feature};
\node[align=center] at (0, 1.6) {\scriptsize Extractors};
\node[align=center] at (1.5, 1.9) {\scriptsize Feature};
\node[align=center] at (1.5, 1.6) {\scriptsize Representation};
% \node at (5.4, 3) {\scriptsize Read-out};
% \node at (2.9, -3) {\scriptsize $\tilde{\mathbf{x}}$};
% \node at (0, -3) {\scriptsize $h_\phi(\mathbf{x})$};  
% \node at (-4.5, -3) {\scriptsize $\mathbf{x}$};
% \node at (5.4, -3) {\scriptsize $f_\theta(\tilde{\mathbf{x}})$};
\vspace{0.5em}
\node at (-3, -2) {\scriptsize \emph{Task 1}};
\node at (-1.4, -2) {\scriptsize \emph{Task 2}};
\node at (1.1, -2) {\scriptsize \emph{Task 1}};
\node at (2.5, -2) {\scriptsize \emph{Task 2}};
\node [opacity=1] at (-2.9, -2.9) {\includegraphics[width=0.1\textwidth]{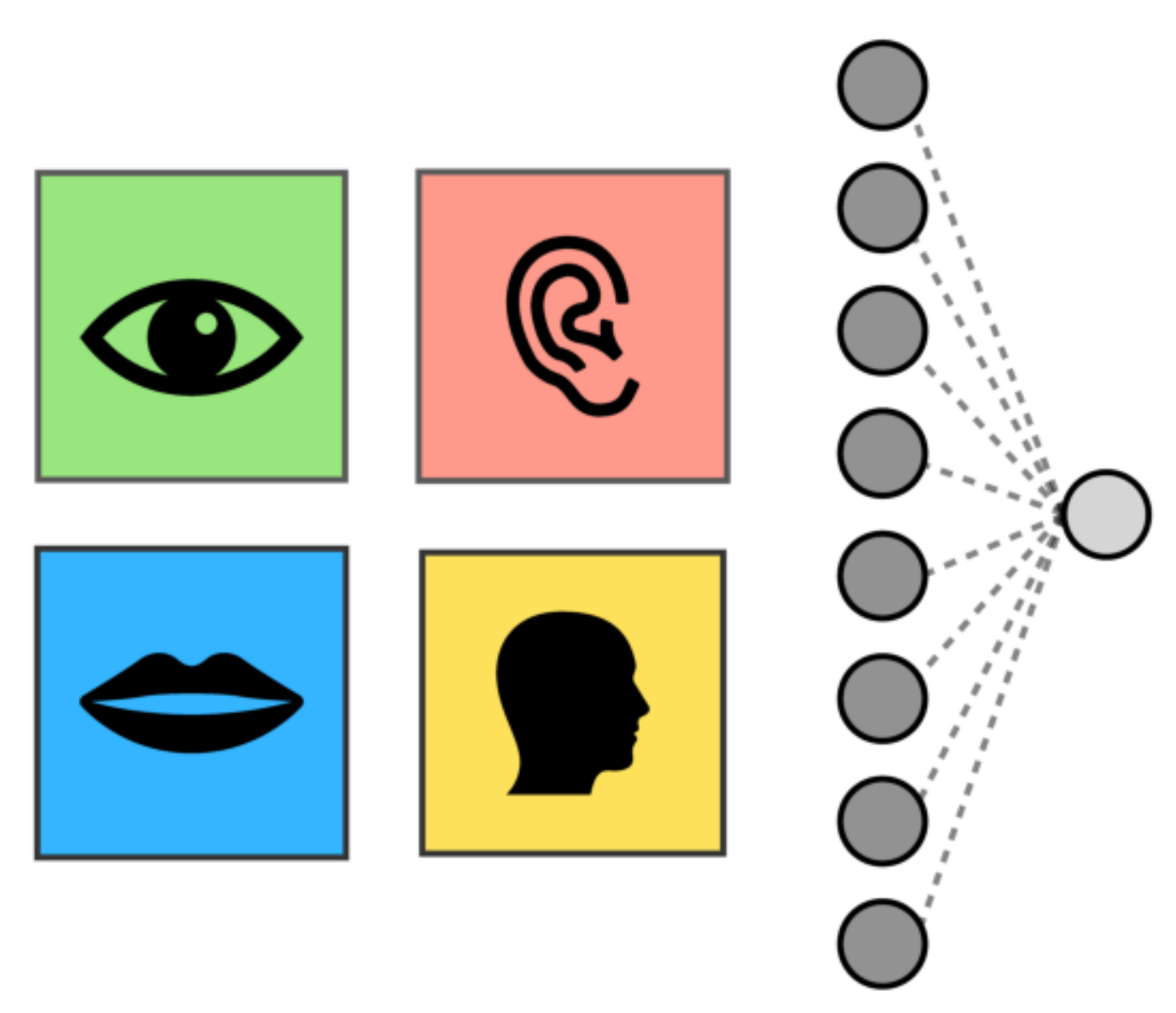}};
\node [opacity=1] at (-1.2, -2.9) {\includegraphics[width=0.1\textwidth]{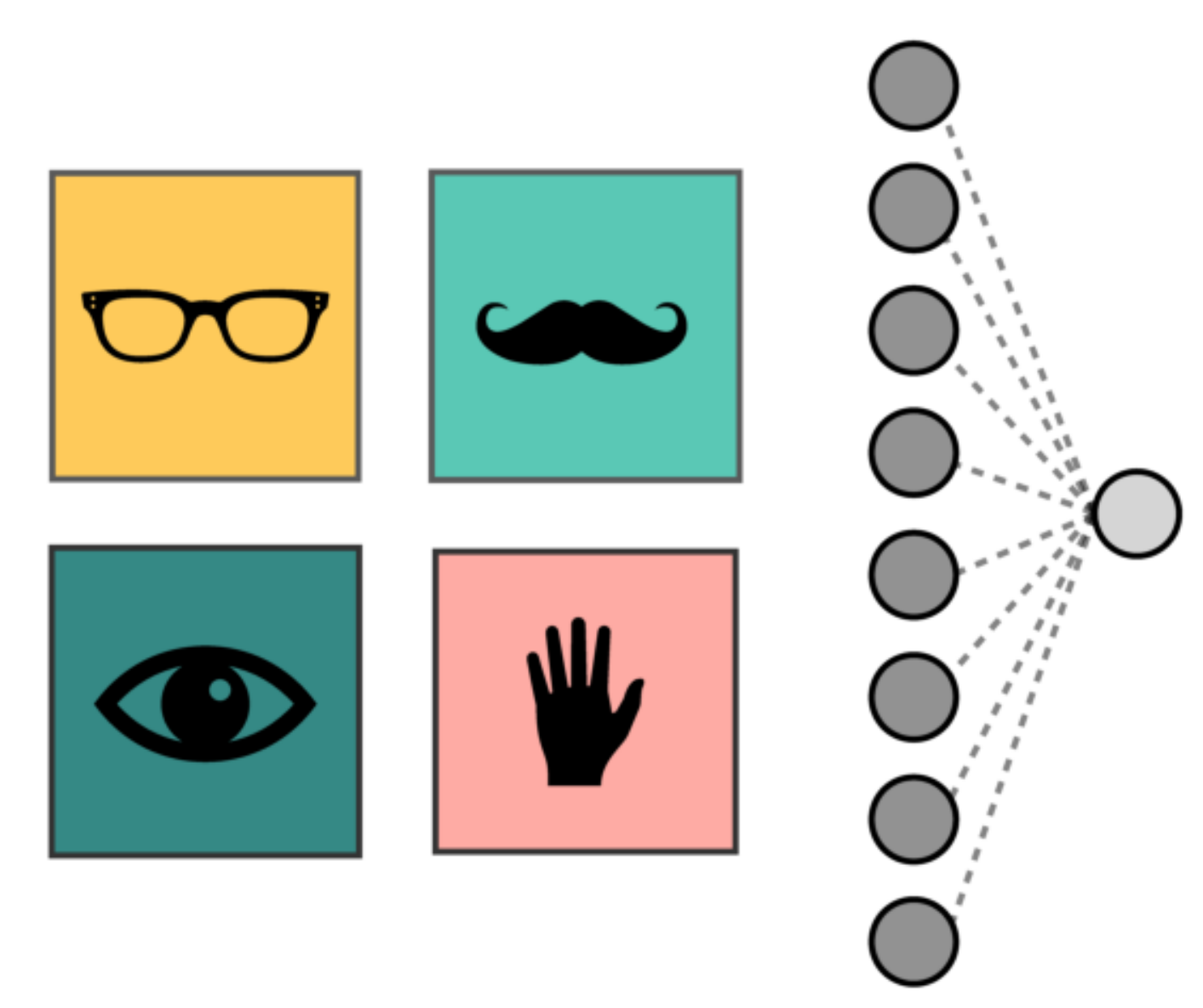}};
\node [opacity=1] at (1.2, -2.9) {\includegraphics[width=0.1\textwidth]{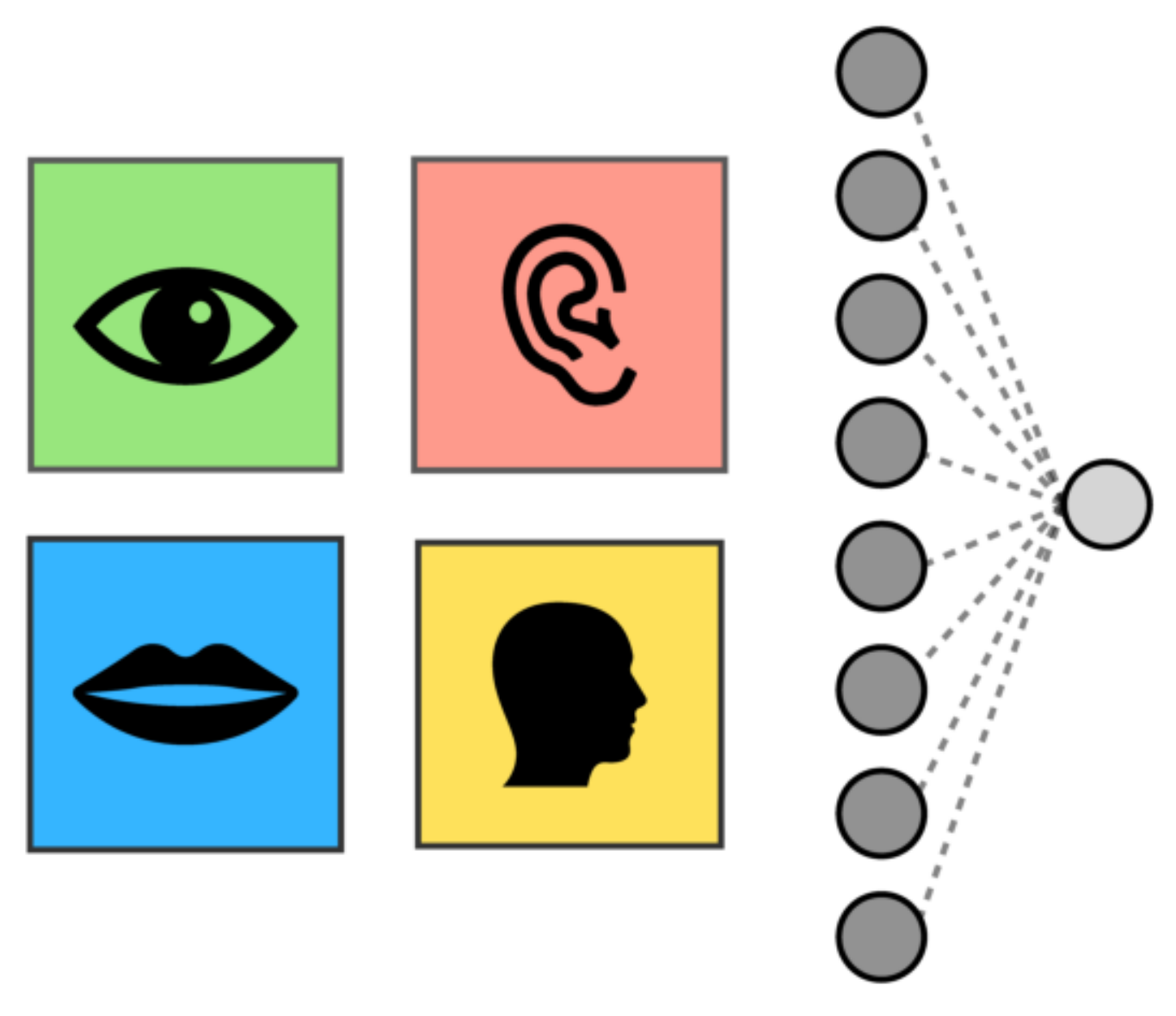}};
\node [opacity=1] at (2.9, -2.9) {\includegraphics[width=0.1\textwidth]{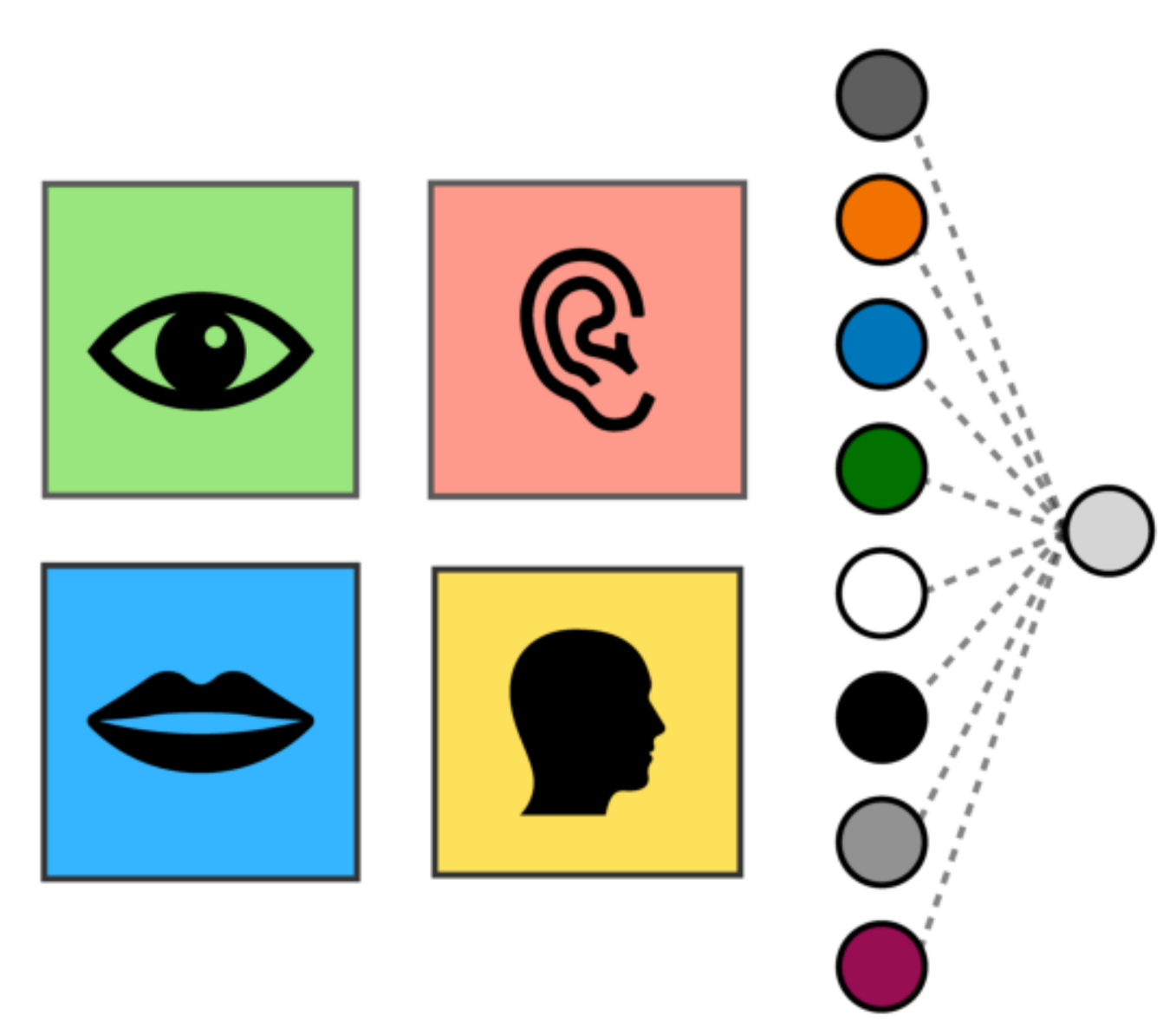}};
\node[align=center] at (-2, -4) {\scriptsize Different Features};
\node[align=center] at (-2, -4.3) {\scriptsize Identical Read-outs};
\node[align=center] at (2, -4) {\scriptsize Identical Features};
\node[align=center] at (2, -4.3) {\scriptsize Different Read-outs};
\end{tikzpicture}
\caption[Schematic demonstrating the distinction between feature similarity and read-out 
similarity.]{\label{fig: feature_readout}\textbf{Distinction between feature similarity and 
read-out similarity}. In a toy image task, a model consists of a set of feature extractors, which produce a feature representation vector 
from the input. This representation is fed into a subsequent \emph{readout} function (top). This distinction between feature extractors and readouts leads to two related notions of task similarity: feature similarity (bottom left), which is captured by~$\mathbf{V}$ in our framework~\autoref{eq: v_overlap}, 
and read-out similarity, which we describe using~$\tilde V$~\autoref{eq:tildeV}.
Some tasks may require the same features with different read-outs while others 
require different features but similar read-outs. For concrete examples of these similariity notiions, consider a continual learning task setting involving binary classification on MNIST images where the first task requires distinguishing between odd digits and even digits and the second requires distinguishing between digits less than or equal to 4 and digits greater than 4. Both tasks could be achieved with the same features but different readout functions. On the other hand two tasks where the first consists of counting blue squares and the second counting red circles will require different features but could use the same readout function.}
\end{figure}

In the previous sections, the task similarity is measured by the teacher-teacher
overlap $V$, which is a metric over the input to hidden weights of the teachers.
There is however another notion of task similarity that is relevant for
two-layer networks: the read-out similarity, which is a metric over the hidden
to output weights. A diagram showing the distinction between these similarities
for a toy image task is shown in~\autoref{fig: feature_readout}.  Most previous
studies~\cite{goodfellow2013empirical, ramasesh2020anatomy} have not directly
studied this distinction, although~\citet{ramasesh2020anatomy} provided evidence that the layers of a deep network that are closer to the input are more responsible to forgetting, pointing to the fact that different layers in a network might have different impact on forgetting. The  teacher-student framework allows us to
disentangle these different aspects of task similarity in detail.

\subsection{The Mean-Field Limit of Neural Networks}

In the large input-limit networks we were considering previously, the hidden
dimensions were small and there was no meaningful way of defining a similarity
over the hidden to output weights. For this reason, in this section we consider
networks in the \emph{mean-field limit} of large hidden dimension,
\begin{equation}
    \phi(\xb; \wb, \vb) = \frac{1}{M}\sum_m^M \vb_mg(\wb_m\xb),
\end{equation}
where we let the number of neurons $M\to\infty$ while the input dimension $D$
remains finite. Note the scaling is different here from the definition
in~\autoref{eq: forward}. In analogy to~\autoref{eq: v_overlap}, let us define
the teacher-teacher read-out similarity as:
\begin{equation}
    \label{eq:tildeV}
    \tilde{V} = \hb^\dagger\cdot \hb^\ddag.
\end{equation}

\newcommand{\btheta}{\boldsymbol{\theta}}
We can thus control both the feature and readout
similarities of the teachers, and measure forgetting and transfer of the student in the $(V, \tilde V)$ plane. However, the student
dynamics cannot be described by the simple set of ODEs from above; instead, the
dynamics of the student are captured by the time-evolution of the density
$\rho(\btheta)$ of the full set of parameters $\btheta$ of the
network~\cite{mei2018mean, rotskoff2018parameters, chizat2018,
  sirignano2018}. This density obeys a partial differential equation,
\begin{equation}
  \partial_{t} \rho_{t}(\boldsymbol{\theta})=\nabla_{\boldsymbol{\theta}} \cdot\left(\rho_{t}(\boldsymbol{\theta}) \nabla_{\boldsymbol{\theta}} \Psi\left(\boldsymbol{\theta} ; \rho_{t}\right)\right)
\end{equation}
where $\Psi(\btheta)$ can be thought of as a potential for the dynamics. This
PDE is hard to analyse, so for the remainder of the paper, we resort to
numerical experiments. 
Since the output dimension of the regression tasks is 1, we can construct
teacher readout weights for a given $\tilde{V}$ with the same procedure as was
used for the feature weights in previous sections. For the feature weights in
the mean-field regime, we first draw a weight matrix for the first teacher
element-wise i.i.d from the normal distribution and orthonormalise it:
$\wb^\dagger =(\wvecb^\dagger)\in\mathbb{R}^{D\times M}$. For the second teacher, the
feature weights are obtained from
\begin{equation}
    \wb^\ddag = \alpha\wb^\dagger + (1 - \alpha^2)\zb,
\end{equation}
where $\zb$ is another $D\times M$ matrix with i.i.d. Gaussian elements, and
$0 < \alpha < 1$ is an interpolation parameter. It can be easily verified that
$\alpha$ also can be interpreted as an overlap between the two feature weight
matrices, where 0 corresponds to orthogonal features and 1 corresponds to
identical features. To make this link clear, we abuse notation slightly and refer to $\alpha$ as $V$ in analogy with previous sections.

%In all these experiments, all weights of the student are learnable i.e.  there
%is no freezing of the first layer after the first task (as is the case for the
%frozen feature model in~\citet{ramasesh2020anatomy}). Simulation of dynamics in
%the frozen-feature regime is left for future work.

\subsection{Results}

Our results are presented in~\autoref{fig:
  2d_aggregate}, where we show (i) the initial rate of forgetting/transfer, (ii)
the maximum amount of forgetting/transfer and (iii) the long-time values of
forgetting/transfer (the value measured at the end of our training). Details of
the procedure used for computing these measures are given in~\autoref{app:
  2d_metric_procedure}.

\begin{figure}[t!]
    \centering
    \includegraphics[width=\columnwidth]{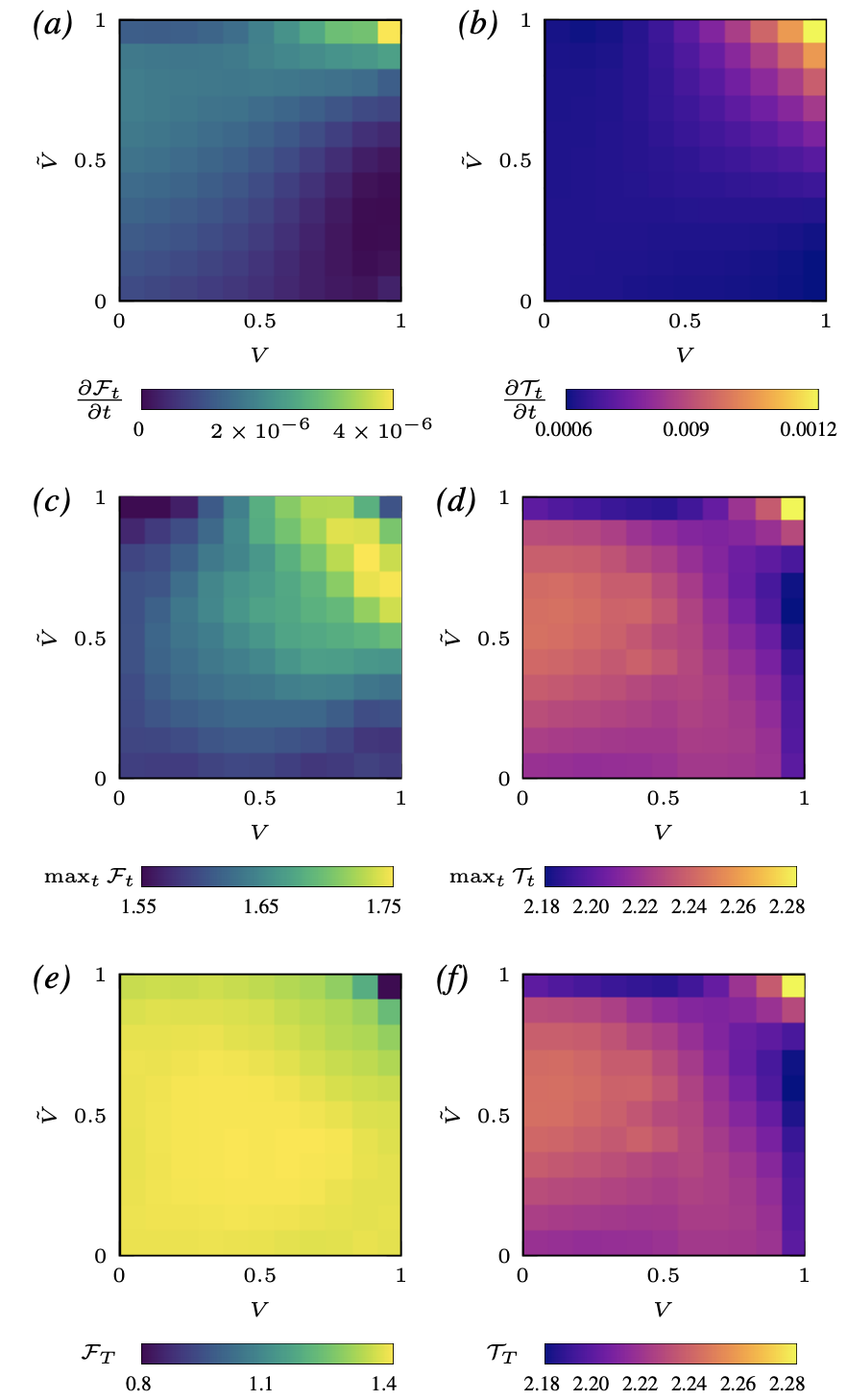}
	\caption[]{\textbf{The interplay of transfer and forgetting as a function of feature and readout similarity.} \emph{(a):} Initial rate of forgetting, $\partial\forgetting{t}/\partial t$; \emph{(b):} Initial rate of transfer, $\partial\transfer{t}/\partial t$; \emph{(c):} Aggregate (long-time) forgetting, $\forgetting{T}$; \emph{(d):} Aggregate (long-time) transfer, $\transfer{T}$; \emph{(e):} Max forgetting reached, $\max_t{\forgetting{t}}$; \emph{(f):} Max transfer reached, $\max_t{\transfer{t}}$. All quantities are plotted on a 2D grid of readout similarity $\tilde{V}$ vs. feature similarity $V$ in two-layer networks in the mean-field limit. \emph{Parameters:} $N=15, M=1000, K=250$.}\label{fig: 2d_aggregate}
\end{figure}

As a check on our setup, we observe that our results are consistent with the
preceding ODE limit simulations, as can be seen from the non-monotonic amount of
forgetting for the maximum forgetting metric  as a function of input
similarity $V$ for full readout similarity ($\tilde V=1$, \autoref{fig: 2d_aggregate}c top row); and the same positive relationship between feature similarity and transfer observed in the early phase of transfer in the ODE limit (\autoref{fig: 2d_aggregate}d top row). We now turn to describing behaviour in the full space of feature and readout similarities.

\subsubsection{Initial Forgetting \& Transfer Rate}\label{sec:2D_initial_rates}
First, we examine the rate of forgetting and transfer at the moment the tasks are switched (\autoref{fig: 2d_aggregate}a-b). The transfer rate is approximately constant along each diagonal, such that it is an approximate function of the summed feature and readout similarity $\tilde V + V$, indicating that both similarities are roughly interchangeable. By contrast, the forgetting rate shows a differential effect of each similarity type, with high readout similarity causing faster forgetting. 

\subsubsection{Max \& Long-Time Forgetting \& Transfer}\label{sec:2D_max_long_time}
The results in~\autoref{fig: 2d_aggregate}c-f demonstrate an intricate
relationship between each type of task similarity and forgetting/transfer dynamics. We make several observations. First, the maximum and long-time metrics differ substantially for forgetting. For instance, the best scenario for limiting maximum forgetting is orthogonal features and fully aligned readouts ($V=0, \tilde V=1$), whereas for long-time forgetting it is complete task overlap ($\tilde V=1,V=1$). Intuitively, learning the same task twice might cause transient forgetting, but eventually will converge to the correct features for both tasks. Maximum forgetting is worst in a narrow band of high summed similarity, whereas long-time forgetting is worst at more moderate levels of summed similarity. Intuitively, very similar but subtly different tasks can produce large transient errors which are ultimately corrected at long times. Finally, for a fixed summed similarity $\tilde V+V$, forgetting is worst when both similarities are approximately equal. This finding generalises the observation that intermediate task similarity is most harmful to forgetting. For transfer, by contrast, the maximum and long-time metrics are highly correlated and often coincide (the point of maximum transfer is the long-time cutoff in our experiment). Outside of tasks that overlap completely, there is a slight trend for better transfer at moderate readout similarity and low feature similarity.

For a constant feature or readout similarity (that is, isolating any column or row with fixed $V$ or $\tilde V$ respectively), we typically observe a non-monotonic relationship between similarity and forgetting that peaks at some intermediate level of similarity. Hence the finding that intermediate amounts of similarity cause greatest forgetting, observed in the ODE limit, holds true at most fixed similarities (see~\autoref{app: isolated_row} for details). 

Finally, we note that transfer depends on readout similarity even for teachers with identical features. Readout similarity has a non-monotonic effect, as can be seen in the column corresponding to full feature similarity ($V=1$). This finding occurs despite the fact that the student network uses a distinct readout head for each task. We surmise that the readout weights bias the solution found by the student for the feature weights during training on the first task. This bias can have the effect of favouring subsequent learning with respect to a second teacher with readout weights that are more similar to those of the first. We show further empirical evidence for this phenomenon in~\autoref{app: larger_feature_movement}.

\section{Conclusion}

Overall, our results depict a complex relationship between task similarity, forgetting, and transfer dynamics. The degree of readout and feature similarity, as well as the timing and form of measurements, all matter in determining the outcome even qualitatively. By characterising the continual learning behaviour of simple gradient descent, we hope our experiments and theoretical framework will serve as a useful foundation for future investigations into the effect of proposed solutions for continual learning in the teacher-student setup, ultimately leading to improved algorithms. 

\section*{Acknowledgements}

We acknowledge support from a Sir Henry Dale Fellowship to A.S. from the Wellcome Trust and Royal Society (grant number 216386/Z/19/Z). A.S. is a CIFAR Azrieli Global Scholar in the Learning in Machines \& Brains programme. 

\bibliography{refs}

\begin{thebibliography}{61}
\providecommand{\natexlab}[1]{#1}
\providecommand{\url}[1]{\texttt{#1}}
\expandafter\ifx\csname urlstyle\endcsname\relax
  \providecommand{\doi}[1]{doi: #1}\else
  \providecommand{\doi}{doi: \begingroup \urlstyle{rm}\Url}\fi

\bibitem[Advani et~al.(2020)Advani, Saxe, and
  Sompolinsky]{advani2020highdimensional}
Advani, M., Saxe, A., and Sompolinsky, H.
\newblock High-dimensional dynamics of generalization error in neural networks.
\newblock \emph{Neural Networks}, 132:\penalty0 428 -- 446, 2020.

\bibitem[Arivazhagan et~al.(2019)Arivazhagan, Bapna, Firat, Lepikhin, Johnson,
  Krikun, Chen, Cao, Foster, Cherry, et~al.]{arivazhagan2019massively}
Arivazhagan, N., Bapna, A., Firat, O., Lepikhin, D., Johnson, M., Krikun, M.,
  Chen, M.~X., Cao, Y., Foster, G., Cherry, C., et~al.
\newblock Massively multilingual neural machine translation in the wild:
  Findings and challenges.
\newblock \emph{arXiv preprint arXiv:1907.05019}, 2019.

\bibitem[Asanuma et~al.(2021)Asanuma, Takagi, Nagano, Yoshida, Igarashi, and
  Okada]{asanuma2021statistical}
Asanuma, H., Takagi, S., Nagano, Y., Yoshida, Y., Igarashi, Y., and Okada, M.
\newblock Statistical mechanical analysis of catastrophic forgetting in
  continual learning with teacher and student networks.
\newblock \emph{arXiv preprint arXiv:2105.07385}, 2021.

\bibitem[Aubin et~al.(2018)Aubin, Maillard, Barbier, Krzakala, Macris, and
  Zdeborov{\'{a}}]{aubin2018committee}
Aubin, B., Maillard, A., Barbier, J., Krzakala, F., Macris, N., and
  Zdeborov{\'{a}}, L.
\newblock {The committee machine: Computational to statistical gaps in learning
  a two-layers neural network}.
\newblock In \emph{Advances in Neural Information Processing Systems 31}, pp.\
  3227--3238, 2018.

\bibitem[Bahri et~al.(2020)Bahri, Kadmon, Pennington, Schoenholz,
  Sohl-Dickstein, and Ganguli]{bahri2020statistical}
Bahri, Y., Kadmon, J., Pennington, J., Schoenholz, S., Sohl-Dickstein, J., and
  Ganguli, S.
\newblock {Statistical Mechanics of Deep Learning}.
\newblock \emph{Annual Review of Condensed Matter Physics}, 11\penalty0
  (1):\penalty0 501--528, 2020.

\bibitem[Baity-Jesi et~al.(2018)Baity-Jesi, Sagun, Geiger, Spigler, Arous,
  Cammarota, LeCun, Wyart, and Biroli]{Baity-Jesi2018}
Baity-Jesi, M., Sagun, L., Geiger, M., Spigler, S., Arous, G., Cammarota, C.,
  LeCun, Y., Wyart, M., and Biroli, G.
\newblock {Comparing Dynamics: Deep Neural Networks versus Glassy Systems}.
\newblock In \emph{Proceedings of the 35th International Conference on Machine
  Learning}, 2018.

\bibitem[Bennani \& Sugiyama(2020)Bennani and
  Sugiyama]{bennani2020generalisation}
Bennani, M.~A. and Sugiyama, M.
\newblock Generalisation guarantees for continual learning with orthogonal
  gradient descent.
\newblock \emph{arXiv preprint arXiv:2006.11942}, 2020.

\bibitem[Biehl \& Schwarze(1993)Biehl and Schwarze]{biehl1993learning}
Biehl, M. and Schwarze, H.
\newblock Learning drifting concepts with neural networks.
\newblock \emph{Journal of Physics A: Mathematical and General}, 26\penalty0
  (11):\penalty0 2651, 1993.

\bibitem[Biehl \& Schwarze(1995)Biehl and Schwarze]{Biehl1995}
Biehl, M. and Schwarze, H.
\newblock {Learning by on-line gradient descent}.
\newblock \emph{J. Phys. A. Math. Gen.}, 28\penalty0 (3):\penalty0 643--656,
  1995.

\bibitem[Biehl et~al.(1996)Biehl, Riegler, and W{\"{o}}hler]{Biehl1996}
Biehl, M., Riegler, P., and W{\"{o}}hler, C.
\newblock {Transient dynamics of on-line learning in two-layered neural
  networks}.
\newblock \emph{Journal of Physics A: Mathematical and General}, 29\penalty0
  (16), 1996.

\bibitem[Chizat \& Bach(2018)Chizat and Bach]{chizat2018}
Chizat, L. and Bach, F.
\newblock On the global convergence of gradient descent for over-parameterized
  models using optimal transport.
\newblock In \emph{Advances in Neural Information Processing Systems 31}, pp.\
  3040--3050, 2018.

\bibitem[Chizat et~al.(2019)Chizat, Oyallon, and Bach]{chizat2019lazy}
Chizat, L., Oyallon, E., and Bach, F.
\newblock On lazy training in differentiable programming.
\newblock In \emph{Advances in Neural Information Processing Systems 33}, pp.\
  forthcoming, 2019.

\bibitem[Cichon \& Gan(2015)Cichon and Gan]{cichon2015branch}
Cichon, J. and Gan, W.-B.
\newblock Branch-specific dendritic ca 2+ spikes cause persistent synaptic
  plasticity.
\newblock \emph{Nature}, 520\penalty0 (7546):\penalty0 180--185, 2015.

\bibitem[Dhifallah \& Lu(2021)Dhifallah and Lu]{dhifallah2021phase}
Dhifallah, O. and Lu, Y.~M.
\newblock Phase transitions in transfer learning for high-dimensional
  perceptrons.
\newblock \emph{arXiv preprint arXiv:2101.01918}, 2021.

\bibitem[Doan et~al.(2020)Doan, Bennani, Mazoure, Rabusseau, and
  Alquier]{doan2020theoretical}
Doan, T., Bennani, M., Mazoure, B., Rabusseau, G., and Alquier, P.
\newblock A theoretical analysis of catastrophic forgetting through the ntk
  overlap matrix.
\newblock \emph{arXiv preprint arXiv:2010.04003}, 2020.

\bibitem[Du et~al.(2018)Du, Lee, Tian, Singh, and Poczos]{du2018gradient}
Du, S., Lee, J., Tian, Y., Singh, A., and Poczos, B.
\newblock Gradient descent learns one-hidden-layer {CNN}: Don’t be afraid of
  spurious local minima.
\newblock In \emph{Proceedings of the 35th International Conference on Machine
  Learning}, volume~80, pp.\  1339--1348, 2018.

\bibitem[Engel \& Van~den Broeck(2001)Engel and Van~den
  Broeck]{engel2001statistical}
Engel, A. and Van~den Broeck, C.
\newblock \emph{Statistical mechanics of learning}.
\newblock Cambridge University Press, 2001.

\bibitem[Farquhar \& Gal(2018)Farquhar and Gal]{farquhar2018towards}
Farquhar, S. and Gal, Y.
\newblock Towards robust evaluations of continual learning.
\newblock \emph{arXiv preprint arXiv:1805.09733}, 2018.

\bibitem[Flesch et~al.(2018)Flesch, Balaguer, Dekker, Nili, and
  Summerfield]{flesch2018comparing}
Flesch, T., Balaguer, J., Dekker, R., Nili, H., and Summerfield, C.
\newblock Comparing continual task learning in minds and machines.
\newblock \emph{Proceedings of the National Academy of Sciences}, 115\penalty0
  (44):\penalty0 E10313--E10322, 2018.

\bibitem[Gabri{\'{e}}(2020)]{gabrie2020meanfield}
Gabri{\'{e}}, M.
\newblock Mean-field inference methods for neural networks.
\newblock \emph{Journal of Physics A: Mathematical and Theoretical},
  53\penalty0 (22):\penalty0 223002, 2020.

\bibitem[Gardner \& Derrida(1989)Gardner and Derrida]{gardner1989}
Gardner, E. and Derrida, B.
\newblock {Three unfinished works on the optimal storage capacity of networks}.
\newblock \emph{Journal of Physics A: Mathematical and General}, 22\penalty0
  (12):\penalty0 1983--1994, 1989.

\bibitem[Ghorbani et~al.(2019)Ghorbani, Mei, Misiakiewicz, and
  Montanari]{ghorbani2019limitations}
Ghorbani, B., Mei, S., Misiakiewicz, T., and Montanari, A.
\newblock Limitations of lazy training of two-layers neural network.
\newblock In \emph{Advances in Neural Information Processing Systems 32}, pp.\
  9111--9121, 2019.

\bibitem[Goldt et~al.(2019)Goldt, Advani, Saxe, Krzakala, and
  Zdeborov{\'a}]{goldt2019dynamics}
Goldt, S., Advani, M., Saxe, A.~M., Krzakala, F., and Zdeborov{\'a}, L.
\newblock Dynamics of stochastic gradient descent for two-layer neural networks
  in the teacher-student setup.
\newblock In \emph{Advances in Neural Information Processing Systems}, pp.\
  6979--6989, 2019.

\bibitem[Goodfellow et~al.(2013)Goodfellow, Mirza, Xiao, Courville, and
  Bengio]{goodfellow2013empirical}
Goodfellow, I.~J., Mirza, M., Xiao, D., Courville, A., and Bengio, Y.
\newblock An empirical investigation of catastrophic forgetting in
  gradient-based neural networks.
\newblock \emph{arXiv preprint arXiv:1312.6211}, 2013.

\bibitem[Jacot et~al.(2018)Jacot, Gabriel, and Hongler]{jacot2018neural}
Jacot, A., Gabriel, F., and Hongler, C.
\newblock Neural tangent kernel: Convergence and generalization in neural
  networks.
\newblock In \emph{Advances in Neural Information Processing Systems 32}, pp.\
  8571--8580, 2018.

\bibitem[Kirkpatrick et~al.(2017)Kirkpatrick, Pascanu, Rabinowitz, Veness,
  Desjardins, Rusu, Milan, Quan, Ramalho, Grabska-Barwinska,
  et~al.]{kirkpatrick2017overcoming}
Kirkpatrick, J., Pascanu, R., Rabinowitz, N., Veness, J., Desjardins, G., Rusu,
  A.~A., Milan, K., Quan, J., Ramalho, T., Grabska-Barwinska, A., et~al.
\newblock Overcoming catastrophic forgetting in neural networks.
\newblock \emph{Proceedings of the national academy of sciences}, 114\penalty0
  (13):\penalty0 3521--3526, 2017.

\bibitem[Lampinen \& Ganguli(2018)Lampinen and Ganguli]{lampinen2018analytic}
Lampinen, A.~K. and Ganguli, S.
\newblock An analytic theory of generalization dynamics and transfer learning
  in deep linear networks.
\newblock \emph{arXiv preprint arXiv:1809.10374}, 2018.

\bibitem[Li \& Hoiem(2017)Li and Hoiem]{li2017learning}
Li, Z. and Hoiem, D.
\newblock Learning without forgetting.
\newblock \emph{IEEE transactions on pattern analysis and machine
  intelligence}, 40\penalty0 (12):\penalty0 2935--2947, 2017.

\bibitem[McClelland et~al.(1995)McClelland, McNaughton, and
  O'Reilly]{McClelland1995}
McClelland, J., McNaughton, B., and O'Reilly, R.
\newblock Why there are complementary learning systems in the hippocampus and
  neocortex: insights from the successes and failures of connectionist models
  of learning and memory.
\newblock \emph{Psychological review}, 102\penalty0 (3):\penalty0 419--57, July
  1995.

\bibitem[McCloskey \& Cohen(1989)McCloskey and
  Cohen]{mccloskey1989catastrophic}
McCloskey, M. and Cohen, N.~J.
\newblock Catastrophic interference in connectionist networks: The sequential
  learning problem.
\newblock In \emph{Psychology of learning and motivation}, volume~24, pp.\
  109--165. Elsevier, 1989.

\bibitem[Mei et~al.(2018)Mei, Montanari, and Nguyen]{mei2018mean}
Mei, S., Montanari, A., and Nguyen, P.-M.
\newblock A mean field view of the landscape of two-layer neural networks.
\newblock \emph{Proceedings of the National Academy of Sciences}, 115\penalty0
  (33):\penalty0 E7665--E7671, 2018.

\bibitem[Mirzadeh et~al.(2020)Mirzadeh, Farajtabar, Pascanu, and
  Ghasemzadeh]{mirzadeh2020understanding}
Mirzadeh, S.~I., Farajtabar, M., Pascanu, R., and Ghasemzadeh, H.
\newblock Understanding the role of training regimes in continual learning.
\newblock \emph{arXiv preprint arXiv:2006.06958}, 2020.

\bibitem[Ndirango \& Lee(2019)Ndirango and Lee]{ndirango2019generalization}
Ndirango, A. and Lee, T.
\newblock Generalization in multitask deep neural classifiers: a statistical
  physics approach.
\newblock In \emph{Advances in Neural Information Processing Systems}, pp.\
  15862--15871, 2019.

\bibitem[Neyshabur et~al.(2020)Neyshabur, Sedghi, and
  Zhang]{neyshabur2020being}
Neyshabur, B., Sedghi, H., and Zhang, C.
\newblock What is being transferred in transfer learning?
\newblock In \emph{Advances in neural information processing systems},
  volume~33, 2020.

\bibitem[Nguyen et~al.(2019)Nguyen, Achille, Lam, Hassner, Mahadevan, and
  Soatto]{nguyen2019toward}
Nguyen, C.~V., Achille, A., Lam, M., Hassner, T., Mahadevan, V., and Soatto, S.
\newblock Toward understanding catastrophic forgetting in continual learning.
\newblock \emph{arXiv preprint arXiv:1908.01091}, 2019.

\bibitem[Parisi et~al.(2019)Parisi, Kemker, Part, Kanan, and
  Wermter]{parisi2019continual}
Parisi, G.~I., Kemker, R., Part, J.~L., Kanan, C., and Wermter, S.
\newblock Continual lifelong learning with neural networks: A review.
\newblock \emph{Neural Networks}, 2019.

\bibitem[Ramasesh et~al.(2020)Ramasesh, Dyer, and Raghu]{ramasesh2020anatomy}
Ramasesh, V.~V., Dyer, E., and Raghu, M.
\newblock Anatomy of catastrophic forgetting: Hidden representations and task
  semantics.
\newblock \emph{arXiv preprint arXiv:2007.07400}, 2020.

\bibitem[Riegler \& Biehl(1995)Riegler and Biehl]{riegler1995line}
Riegler, P. and Biehl, M.
\newblock On-line backpropagation in two-layered neural networks.
\newblock \emph{Journal of Physics A: Mathematical and General}, 28\penalty0
  (20):\penalty0 L507, 1995.

\bibitem[Rotskoff \& Vanden-Eijnden(2018)Rotskoff and
  Vanden-Eijnden]{rotskoff2018parameters}
Rotskoff, G. and Vanden-Eijnden, E.
\newblock Parameters as interacting particles: long time convergence and
  asymptotic error scaling of neural networks.
\newblock In \emph{Advances in neural information processing systems}, pp.\
  7146--7155, 2018.

\bibitem[Ruder \& Plank(2017)Ruder and Plank]{ruder2017learning}
Ruder, S. and Plank, B.
\newblock Learning to select data for transfer learning with bayesian
  optimization.
\newblock \emph{arXiv preprint arXiv:1707.05246}, 2017.

\bibitem[Rusu et~al.(2016)Rusu, Rabinowitz, Desjardins, Soyer, Kirkpatrick,
  Kavukcuoglu, Pascanu, and Hadsell]{rusu2016progressive}
Rusu, A.~A., Rabinowitz, N.~C., Desjardins, G., Soyer, H., Kirkpatrick, J.,
  Kavukcuoglu, K., Pascanu, R., and Hadsell, R.
\newblock Progressive neural networks.
\newblock \emph{arXiv preprint arXiv:1606.04671}, 2016.

\bibitem[Saad(2009)]{saad2009line}
Saad, D.
\newblock \emph{On-line learning in neural networks}, volume~17.
\newblock Cambridge University Press, 2009.

\bibitem[Saad \& Solla(1995{\natexlab{a}})Saad and Solla]{saad1995exact}
Saad, D. and Solla, S.~A.
\newblock Exact solution for on-line learning in multilayer neural networks.
\newblock \emph{Physical Review Letters}, 74\penalty0 (21):\penalty0 4337,
  1995{\natexlab{a}}.

\bibitem[Saad \& Solla(1995{\natexlab{b}})Saad and Solla]{saad1995line}
Saad, D. and Solla, S.~A.
\newblock On-line learning in soft committee machines.
\newblock \emph{Physical Review E}, 52\penalty0 (4):\penalty0 4225,
  1995{\natexlab{b}}.

\bibitem[Saxe et~al.(2018)Saxe, Bansal, Dapello, Advani, Kolchinsky, Tracey,
  and Cox]{saxe2018information}
Saxe, A., Bansal, Y., Dapello, J., Advani, M., Kolchinsky, A., Tracey, B., and
  Cox, D.
\newblock {On the information bottleneck theory of deep learning}.
\newblock In \emph{ICLR}, 2018.

\bibitem[Seung et~al.(1992)Seung, Sompolinsky, and
  Tishby]{seung1992statistical}
Seung, H.~S., Sompolinsky, H., and Tishby, N.
\newblock Statistical mechanics of learning from examples.
\newblock \emph{Physical review A}, 45\penalty0 (8):\penalty0 6056, 1992.

\bibitem[Shin et~al.(2017)Shin, Lee, Kim, and Kim]{shin2017continual}
Shin, H., Lee, J.~K., Kim, J., and Kim, J.
\newblock Continual learning with deep generative replay.
\newblock In \emph{Advances in Neural Information Processing Systems}, pp.\
  2990--2999, 2017.

\bibitem[Sirignano \& Spiliopoulos(2019)Sirignano and
  Spiliopoulos]{sirignano2018}
Sirignano, J. and Spiliopoulos, K.
\newblock {Mean field analysis of neural networks: A central limit theorem}.
\newblock \emph{Stochastic Processes and their Applications}, 2019.

\bibitem[Sirignano \& Spiliopoulos(2020)Sirignano and
  Spiliopoulos]{sirignano2020mean}
Sirignano, J. and Spiliopoulos, K.
\newblock Mean field analysis of neural networks: A central limit theorem.
\newblock \emph{Stochastic Processes and their Applications}, 130\penalty0
  (3):\penalty0 1820--1852, 2020.

\bibitem[Soltanolkotabi et~al.(2018)Soltanolkotabi, Javanmard, and
  Lee]{soltanolkotabi2018theoretical}
Soltanolkotabi, M., Javanmard, A., and Lee, J.
\newblock Theoretical insights into the optimization landscape of
  over-parameterized shallow neural networks.
\newblock \emph{IEEE Transactions on Information Theory}, 65\penalty0
  (2):\penalty0 742--769, 2018.

\bibitem[Tian(2017)]{tian2017analytical}
Tian, Y.
\newblock An analytical formula of population gradient for two-layered relu
  network and its applications in convergence and critical point analysis.
\newblock In \emph{Proceedings of the 34th International Conference on Machine
  Learning - Volume 70}, pp.\  3404–3413, 2017.

\bibitem[Toneva et~al.(2018)Toneva, Sordoni, Combes, Trischler, Bengio, and
  Gordon]{toneva2018empirical}
Toneva, M., Sordoni, A., Combes, R. T.~d., Trischler, A., Bengio, Y., and
  Gordon, G.~J.
\newblock An empirical study of example forgetting during deep neural network
  learning.
\newblock \emph{arXiv preprint arXiv:1812.05159}, 2018.

\bibitem[Watkin et~al.(1993)Watkin, Rau, and Biehl]{watkin1993}
Watkin, T., Rau, A., and Biehl, M.
\newblock {The statistical mechanics of learning a rule}.
\newblock \emph{Reviews of Modern Physics}, 65\penalty0 (2):\penalty0 499--556,
  1993.

\bibitem[Xiao et~al.(2017)Xiao, Rasul, and Vollgraf]{xiao2017online}
Xiao, H., Rasul, K., and Vollgraf, R.
\newblock Fashion-mnist: a novel image dataset for benchmarking machine
  learning algorithms.
\newblock 2017.

\bibitem[Yang et~al.(2014)Yang, Lai, Cichon, Ma, Li, and Gan]{yang2014sleep}
Yang, G., Lai, C. S.~W., Cichon, J., Ma, L., Li, W., and Gan, W.-B.
\newblock Sleep promotes branch-specific formation of dendritic spines after
  learning.
\newblock \emph{Science}, 344\penalty0 (6188):\penalty0 1173--1178, 2014.

\bibitem[Yoshida \& Okada(2019)Yoshida and Okada]{yoshida2019datadependence}
Yoshida, Y. and Okada, M.
\newblock Data-dependence of plateau phenomenon in learning with neural network
  --- statistical mechanical analysis.
\newblock In \emph{Advances in Neural Information Processing Systems 32}, pp.\
  1720--1728, 2019.

\bibitem[Zdeborov{\'{a}}(2020)]{zdeborova2020understanding}
Zdeborov{\'{a}}, L.
\newblock {Understanding deep learning is also a job for physicists}.
\newblock \emph{Nature Physics}, 2020.

\bibitem[Zdeborov{\'a} \& Krzakala(2016)Zdeborov{\'a} and
  Krzakala]{zdeborova2016statistical}
Zdeborov{\'a}, L. and Krzakala, F.
\newblock Statistical physics of inference: Thresholds and algorithms.
\newblock \emph{Advances in Physics}, 65\penalty0 (5):\penalty0 453--552, 2016.

\bibitem[Zenke et~al.(2017)Zenke, Poole, and Ganguli]{pmlr-v70-zenke17a}
Zenke, F., Poole, B., and Ganguli, S.
\newblock Continual learning through synaptic intelligence.
\newblock In Precup, D. and Teh, Y.~W. (eds.), \emph{Proceedings of the 34th
  International Conference on Machine Learning}, volume~70 of \emph{Proceedings
  of Machine Learning Research}, pp.\  3987--3995, International Convention
  Centre, Sydney, Australia, 06--11 Aug 2017. PMLR.

\bibitem[Zhong et~al.(2017)Zhong, Song, Jain, Bartlett, and
  Dhillon]{zhong2017recovery}
Zhong, K., Song, Z., Jain, P., Bartlett, P., and Dhillon, I.
\newblock Recovery guarantees for one-hidden-layer neural networks.
\newblock In \emph{Proceedings of the 34th International Conference on Machine
  Learning-Volume 70}, pp.\  4140--4149, 2017.

\bibitem[Zimmer et~al.(2014)Zimmer, Viappiani, and Weng]{zimmer2014teacher}
Zimmer, M., Viappiani, P., and Weng, P.
\newblock Teacher-student framework: a reinforcement learning approach.
\newblock In \emph{AAMAS Workshop Autonomous Robots and Multirobot Systems},
  2014.

\end{thebibliography}
\bibliographystyle{icml2021}

\clearpage
\appendix
%%%%%%%%%%%%%%%%%%%%%%%%%%%%%%%%%%%%%%%%%%%%%%%%%%%%%%%%%%%%%%%%%%%%%%%%%%%%%%%
%%%%%%%%%%%%%%%%%%%%%%%%%%%%%%%%%%%%%%%%%%%%%%%%%%%%%%%%%%%%%%%%%%%%%%%%%%%%%%%
\onecolumn

%%%%%%%%%%%%%%
\numberwithin{equation}{section}% \renewcommand{\theequation}{S.\arabic{equation}}

\section{Reproducing the results of Ramasesh et al.~with two-layer neural networks}\label{app: deeper}

\begin{figure}[t!]
  \centering
  \vspace*{-.5em}
  \includegraphics[width=.6\linewidth]{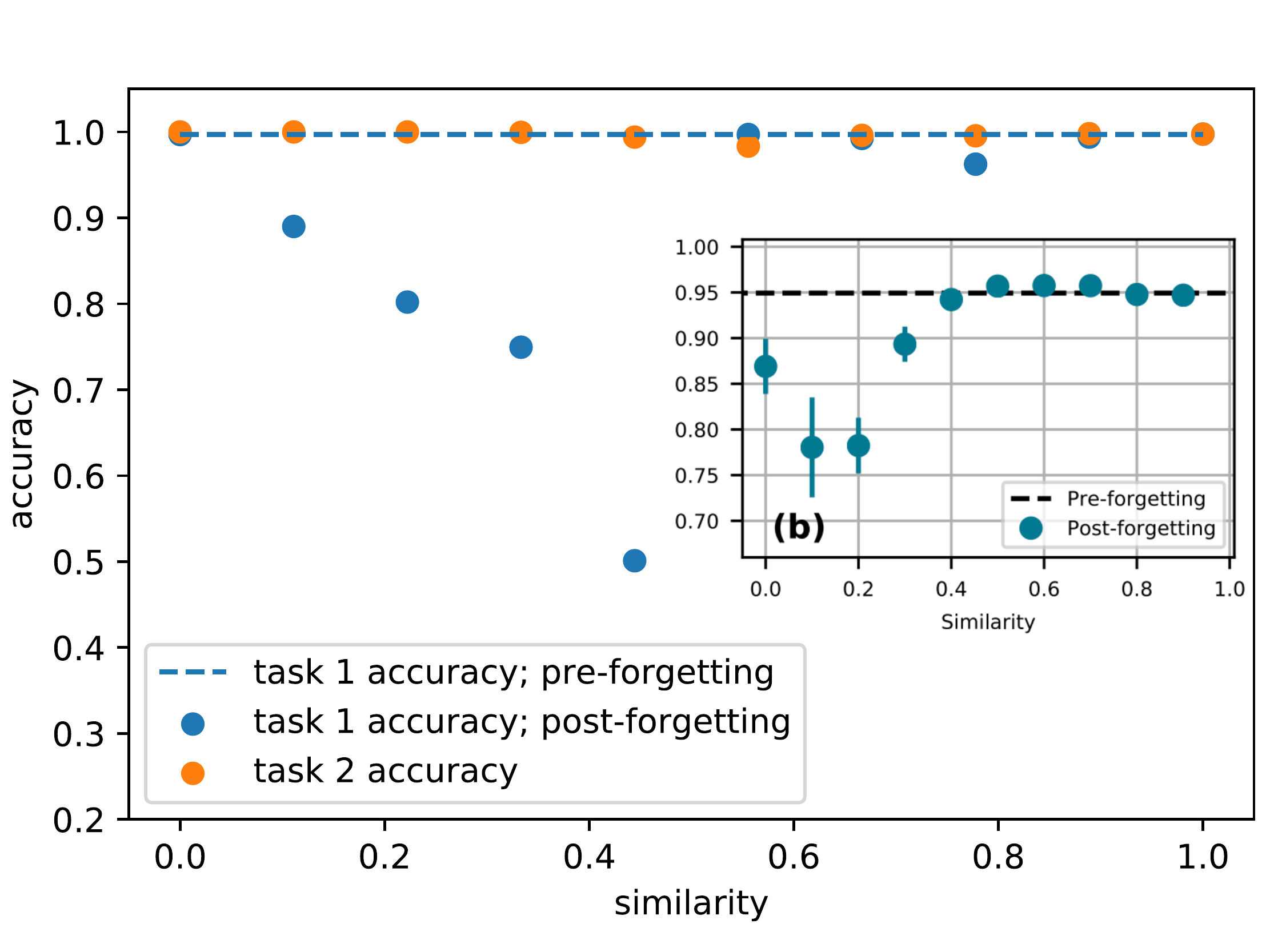}
  \vspace*{-1.5em}
  \caption{\label{fig:reproduction} \textbf{Task similarity in deep vs.~shallow networks}  We plot the accuracy  of a two-layer ReLU network with 8 neurons trained on two tasks. The first task is discriminating T-shirts from
    high-heels on Fashion MNIST (task 1). The second task is a linear
    interpolation in both inputs and labels between task 1 and long-sleeve shirts vs trainers. In the inset, we reproduce Fig.~5b of~\textcite{ramasesh2020anatomy} when training various deep networks on two tasks obtained by linearly interpolation of
    CIFAR10 images.  \emph{Parameters of our experiment}:
    learning rate 0.01, $D=784$.}
\end{figure}

We report in \autoref{fig:reproduction} a reproduction of an experiment showing that the two-layer networks trained on FashionMNIST~\cite{xiao2017online} reproduce a key observation of~\cite{ramasesh2020anatomy} made for VGG, ResNet and DenseNet on CIFAR10:  intermediate  task  similarity leads to worst forgetting.  To that end, we trained a two-layer ReLU network with 8 neurons to discriminate T-shirts from
    high-heels on Fashion MNIST (task 1). The second task was a linear
    interpolation in both inputs and labels between task 1 and long-sleeve shirts vs trainers. We see that at
    intermediate task similarity, or halfway along the linear interpolation between the two datasets, forgetting of the first task is the
    worst. This is the same behaviour~\textcite{ramasesh2020anatomy} found
    consistently for VGG, ResNet and DenseNet when linearly interpolating
    CIFAR10 images (we reproduce their Fig.~5b in the inset). Hence the toy model studied here reproduces this behaviour of more realistic setups. 
\section{Order Parameters}\label{app: order_parameters}

The full set of order parameters for the two-teacher student-teacher networks in the large input limit is given by:

\begin{align}
	\text{\footnotesize Student-Student Overlap, }\mathbf{Q}:  q_{kl} & \equiv \langle\lambda_k\lambda_l\rangle %= \frac{1}{N}\sum_{a, b}^N w_{ka}w_{lb}\langle x_ax_b\rangle 
	= \frac{1}{N}\wvecb_k\wvecb_l;\\
	\text{\footnotesize Teacher}^\dagger\text{\footnotesize -Teacher}^\dagger\text{\footnotesize Overlap, }\mathbf{T}: t_{nm} & \equiv \langle\rho_m\rho_n\rangle %= \frac{1}{N}\sum_{a, b}^N w^\dagger_{ma}w^\dagger_{nb}\langle x_ax_b\rangle 
	= \frac{1}{N}\wvecb^\dagger_m\wvecb^\dagger_n  ;\\
	\text{\footnotesize Student-Teacher}^\dagger\text{\footnotesize Overlap, }\mathbf{R}:  r_{km} & \equiv \langle\lambda_k\rho_m\rangle %= \frac{1}{N}\sum_{a, b}^N w_{ka}w^\dagger_{mb}\langle x_ax_b\rangle 
	= \frac{1}{N}\wvecb_k\wvecb^\dagger_m;\\
	\text{\footnotesize Teacher}^\ddag\text{\footnotesize -Teacher}^\ddag\text{\footnotesize Overlap, }\mathbf{S}: s_{pq} & \equiv \langle\eta_p\eta_q\rangle %= \frac{1}{N}\sum_{a, b}^N w^\ddag_{pa}w^\ddag_{qb}\langle x_ax_b\rangle
	= \frac{1}{N}\wvecb^\ddag_p\wvecb^\ddag_q ;\\
	\text{\footnotesize Student-Teacher}^\ddag\text{\footnotesize Overlap, }\mathbf{U}:  u_{kp}&\equiv \langle\lambda_k\eta_p\rangle %= \frac{1}{N}\sum_{a, b}^N w_{ka}w^\ddag_{pb}\langle x_ax_b\rangle
	= \frac{1}{N}\wvecb_k\wvecb^\ddag_p ; \\
	\text{\footnotesize Teacher}^\dagger\text{\footnotesize -Teacher}^\ddag\text{\footnotesize Overlap, }\mathbf{V}: v_{mp} & \equiv\langle\rho_m\eta_p\rangle% = \frac{1}{N}\sum_{a, b}^N w^\dagger_{ma}w^\ddag_{pb}\langle x_ax_b\rangle 
	= \frac{1}{N}\wvecb^\dagger_m\wvecb^\ddag_p .
\end{align}
\section{\gls{ode} Derivation}\label{app: derivation}

This section presents the derivation of the \gls{ode} formulation of the generalisation error for the student-multi-teacher continual learning framework.

\subsection{Generalisation Error in terms of Order Parameters}

Our aim is to formulate the generalisation error in terms of the macroscopic order parameters. Let us begin 
by multiplying out~\autoref{eq:eg},
\begin{align}
	%&= \frac{1}{2}\left\langle\left[\sum_{k=1}^Kh_k^\dagger g(\lambda_k)\sum_{i=1}^Kh_i^\dagger g(\lambda_i) + \sum_{m=1}^Mv_m^\dagger g(\rho_m)\sum_{n=1}^Mv_n^\dagger g(\rho_n) - 2\sum_{n=1}^Kh_k^\dagger g(\lambda_k)\sum_{m=1}^Mv_m^\dagger g(\rho_m)\right]\right\rangle\\
	\epsilon_g^\dagger = \frac{1}{2}\left\langle\left[\sum_{i, k}h_i^\dagger h_k^\dagger g(\lambda_i)g(\lambda_k)
	+\sum_{m, n}v_m^\dagger v_n^\dagger g(\rho_m)g(\rho_n) - 2\sum_{i,n}h_i^\dagger v_n^\dagger g(\lambda_i)g(\rho_n)\right]\right\rangle.
\end{align}
and similarly for the second student.
%\begin{align}
%	\epsilon_g^\ddag &= \frac{1}{2}\left\langle\left[\sum_{k=1}^Kh_k^\ddag g(\lambda_k)\sum_{i=1}^Kh_i^\ddag g(\lambda_i) + \sum_{p=1}^Mv_p^\ddag g(\eta_p)\sum_{q=1}^Mv_q^\ddag g(\eta_q) - 2\sum_{n=1}^Kh_k^\ddag g(\lambda_k)\sum_{p=1}^Mv_p^\ddag g(\eta_p)\right]\right\rangle\\
%	&= \frac{1}{2}\left\langle\left[\sum_{i, k}h_i^\ddag h_k^\ddag g(\lambda_i)g(\lambda_k) + \sum_{p, q}v_p^\ddag v_q^\ddag g(\eta_p)g(\eta_q) - 2\sum_{i,p}h_i^\ddag v_p^\ddag g(\lambda_i)g(\eta_p)\right]\right\rangle.
%\end{align}
These generalisation errors involve averages of local fields, which can be computed as integrals over a joint 
multivariate Gaussian probability distribution, all of the form
\begin{equation}
    \mathcal{P}(\beta, \gamma) = \frac{1}{\sqrt{(2\pi)^{F + H}|\tilde{\mathbf{C}}|}}\exp{\left\{-\frac{1}{2}(\beta, \gamma)^T\tilde{\mathbf{C}}^{-1}(\beta,\gamma)\right\}},
\end{equation}
where $\beta$ and $\gamma$ are local fields with number of units $F$ and $H$ respectively, and $\tilde{\mathbf{C}}$ is 
a covariance matrix suitably projected down from
\[
\mathbf{C} = 
\begin{pmatrix} 
	\mathbf{Q} & \mathbf{R} & \mathbf{U}\\
	\mathbf{R}^T & \mathbf{T} & \mathbf{V}\\
	\mathbf{U}^T & \mathbf{V}^T & \mathbf{S}
\end{pmatrix}.
\]
We define
\begin{equation}
	I_2(f, h) \equiv \langle g(\beta)g(\gamma)\rangle, \label{eq: i2}
\end{equation}
where $f, h$ are the indices corresponding to the units of the local fields $\beta$ and $\gamma$. 
This allows us to write the generalisation errors as 
\begin{align}
	\epsilon_g^\dagger &= \frac{1}{2}\sum_{i,k}h^\dagger_ih^\dagger_kI_2(i,k) + \frac{1}{2}\sum_{n,m}v^\dagger_nv^\dagger_mI_2(n,m) - \sum_{i,n}h^\dagger_iv^\dagger_nI_2(i,n)\label{eq: generror1I}\\
	\epsilon_g^\ddag &= \frac{1}{2}\sum_{i,k}h^\ddag_ih^\ddag_kI_2(i,k) + \frac{1}{2}\sum_{p,q}v^\ddag_pv^\ddag_qI_2(p,q) - \sum_{i,p}h^\ddag_iv^\ddag_pI_2(i,p).\label{eq: generror2I}
\end{align}

\subsubsection{Sigmoidal Activation}
For the scaled error activation function, $g(x) = \text{erf}(x/\sqrt{2})$, there is an analytic 
expression for the $I_2$ integral purely in terms of the order parameters~\cite{saad1995exact}:
\begin{equation}
	I_2(i,k) = \frac{1}{\pi}\arcsin{\frac{q_{ik}}{\sqrt{(1 + q_{ii})(1 + q_{kk})}}}.
\end{equation}
In turn, we can similarly write the generalisation errors in terms of the order parameters only:
\begin{multline}
	\epsilon_g^\dagger = \frac{1}{\pi}\sum_{i,k} h^\dagger_ih^\dagger_k\arcsin{\frac{q_{ik}}{\sqrt{(1 + q_{ii})(1 + q_{kk})}}} + \frac{1}{\pi}\sum_{n,m} v^\dagger_nv^\dagger_m\arcsin{\frac{t_{nm}}{\sqrt{(1 + t_{nn})(1 + t_{mm})}}}\\
	+ \frac{2}{\pi}\sum_{i,n} h^\dagger_iv^\dagger_n\arcsin{\frac{r_{in}}{\sqrt{(1 + q_{ii})(1 + t_{nn})}}}
\end{multline}
\begin{multline}
	\epsilon_g^\ddag = \frac{1}{\pi}\sum_{i,k} h^\ddag_ih^\ddag_k\arcsin{\frac{q_{ik}}{\sqrt{(1 + q_{ii})(1 + q_{kk})}}} + \frac{1}{\pi}\sum_{p,q} v^\ddag_pv^\ddag_q\arcsin{\frac{s_{pq}}{\sqrt{(1 + s_{pp})(1 + s_{qq})}}}\\
	+ \frac{2}{\pi}\sum_{i,p} h^\ddag_iv^\ddag_p\arcsin{\frac{u_{ip}}{\sqrt{(1 + q_{ii})(1 + s_{pp})}}}.
\end{multline}

\subsection{Order Parameter Evolution (Training on $\dagger$)}

Having arrived at expressions for the generalisation error of both teachers in terms of the order parameters, we want to determine equations 
of motion for these order parameters from the weight update equations (\autoref{eq: wupdate} \& \autoref{eq: vupdate}). Trivially, the order 
parameters associated with the two teachers, $\mathbf{T}$ and $\mathbf{S}$ are constant over time, as are the head weights of the teachers, $\vb^\dagger, \vb^\ddag$. 
When training on $\dagger$, the student head weights corresponding to $\ddag$ are also stationary; it remains for us to find equations 
of motion for $\mathbf{R}, \mathbf{Q}, \mathbf{U}$ and $\mathbf{h}^\dagger$, which we derive below. The equivalent derivations
when training on teacher $\ddag$ can be made by using the update in~\autoref{eq: vupdate} instead.

\subsubsection{\gls{ode} for $\mathbf{R}$}

Consider the inner product of \autoref{eq: wupdate} (in the case of * = $\dagger$) with $\wvecb_n^\dagger$:
\begin{align}
	\wvecb_k^{\mu+1}\wvecb_n^\dagger - \wvecb_k^\mu \wvecb_n^\dagger &= -\frac{\alpha_\wb}{\sqrt{D}}h_k^{\dagger\mu} g'(\lambda_k^\mu)\Delta^{\dagger\mu} \xb^\mu \wvecb_n^\dagger\\
	&= -\alpha_\wb h_k^{\dagger\mu} g'(\lambda_k^\mu)\Delta^{\dagger\mu} \rho_n^\mu\\ 
	r_{kn}^{\mu+1} - r_{kn}^\mu&= -\frac{\alpha_\wb}{D} h_k^{\dagger\mu} g'(\lambda_k^\mu)\Delta^{\dagger\mu} \rho_n^\mu
\end{align}
If we let $\tau \equiv \mu/D$ and take the thermodynamic limit of $D\to\infty$, the time parameter becomes continuous and we can write:
\begin{equation}
	\frac{dr_{in}}{d\tau} = -\alpha_\wb h_i^\dagger \langle g'(\lambda_i)\Delta^\dagger \rho_n\rangle,
\end{equation}
where we have re-indexed $k\to i$.

\subsubsection{\gls{ode} for $\mathbf{Q}$}

Consider squaring \autoref{eq: wupdate} (here we can simply use * to denote training on either teacher).
\begin{align}
	\wvecb_k^{\mu+1}\wvecb_i^{\mu+1} - \wvecb_k^\mu \wvecb_i^\mu &=-\frac{\alpha_\wb}{\sqrt{D}}h_i^{*\mu} g'(\lambda_i^\mu)\Delta^{*\mu} \xb^\mu \wvecb_k^\mu -\frac{\alpha_\wb}{\sqrt{D}}h_k^{*\mu} g'(\lambda_k^\mu)\Delta^{*\mu} \xb^\mu \wvecb_i^\mu \nonumber\\ 
	&\quad \quad + \frac{\alpha_\wb^2}{D} h_i^{*\mu} g'(\lambda_i^\mu)h_k^{*\mu} g'(\lambda_k^\mu)(\Delta^{*\mu} \xb^\mu)^2 \\
	&= - \alpha_\wb h_i^{*\mu} g'(\lambda_i^\mu)\Delta^{*\mu}\lambda_k^\mu -\alpha_\wb h_k^{*\mu} g'(\lambda_k^\mu)\Delta^{*\mu} \lambda_i^\mu \nonumber\\
	&\quad \quad+\frac{\alpha_\wb^2}{D} h_i^{*\mu} g'(\lambda_i^\mu)h_k^{*\mu} g'(\lambda_k^\mu)(\Delta^{*\mu}\xb^\mu)^2 \\
	q^{\mu+1}_{ki} - q^{\mu}_{ki}&= -\frac{\alpha_\wb}{D} h_i^{*\mu} g'(\lambda_i^\mu)\Delta^{*\mu}\lambda_k^\mu - \frac{\alpha_\wb}{D} h_k^{*\mu} g'(\lambda_k^\mu)\Delta^{*\mu} \lambda_i^\mu\nonumber\\
	&\quad \quad + \frac{\alpha_\wb^2}{D^2} h_i^{*\mu} g'(\lambda_i^\mu)h_k^{*\mu} g'(\lambda_k^\mu)(\Delta^{*\mu}\xb^\mu)^2. 
\end{align}
Performing the same reparameterisation of $\mu$ and the same thermodynamic limit, we get:
\begin{align}
	\frac{dq_{ik}}{d\tau} = -\alpha_\wb h_i^* \langle g'(\lambda_i)\Delta^* \lambda_k\rangle - \alpha_\wb h_k^* \langle g'(\lambda_k)\Delta^* \lambda_i\rangle +\alpha_\wb^2 h_i^*h_k^*\langle g'(\lambda_i)g'(\lambda_k)\Delta^{*2}\rangle.
\end{align}
Note: in the limit, $(\xb^\mu)^2 \to D$ since individual samples are taken from a unit normal. Hence the $1/D$ limit 
remains the same decay rate for each term.

\subsubsection{\gls{ode} for $\mathbf{U}$}

Consider the inner product of \autoref{eq: wupdate} (in the case of * = $\dagger$) with $\wvecb_p^\ddag$:
\begin{align}
	\wvecb_k^{\mu+1}\wvecb_p^\ddag - \wvecb_k^\mu \wvecb_p^\ddag &= -\frac{\alpha_\wb}{\sqrt{D}}h_k^{\dagger\mu} g'(\lambda_k^\mu)\Delta^{\dagger\mu} \xb^\mu \wvecb_p^\ddag\\
	&= -\alpha_\wb h_k^{\dagger\mu} g'(\lambda_k^\mu)\Delta^{\dagger\mu} \eta_p^\mu\\
	u_{kp}^{\mu+1} - u_{kp}^\mu&= -\frac{\alpha_\wb}{D} h_k^{\dagger\mu} g'(\lambda_k^\mu)\Delta^{\dagger\mu} \eta_p^\mu.
\end{align}
If we let $\tau \equiv \mu/D$ and take the thermodynamic limit of $D\to\infty$:
\begin{equation}
	\frac{du_{ip}}{d\tau} = -\alpha_\wb h_i^* \langle g'(\lambda_i)\Delta^* \eta_p\rangle. \label{eq: uip}
\end{equation}

\subsubsection{\gls{ode} for $\hb^*$}

Here, we simply take the thermodynamic limit of \autoref{eq: vupdate} (for * = $\dagger$):
\begin{equation}
	\frac{dh_i^\dagger}{d\tau} = -\alpha_h\langle\Delta^\dagger g(\lambda_i)\rangle
\end{equation}

% \bibliographystyle{unsrt}
% \bibliography{week_2}

\section{Explicit Formulation}

We can go one step further and write the right hand sides of the \glspl{ode} in terms of more concise integrals. 
Recall that for no noise
\begin{equation}
	\Delta^{\dagger\mu} \equiv \sum_k h_k^{\dagger\mu}g(\lambda_k^\mu) - \sum_m v_m^\dagger g(\rho_m^\mu).
\end{equation}
Substituting this term into the \glspl{ode} above gives us the expanded versions below:
\begin{align}
	\frac{dr_{in}}{d\tau} &= -\alpha_\wb h_i^\dagger \left\langle g'(\lambda_i)\left[\sum_k h_k^\dagger g(\lambda_k) - \sum_m v_m^\dagger g(\rho_m)\right] \rho_n\right\rangle;\\
	\frac{dq_{ik}}{d\tau} &= -\alpha_\wb h_i^\dagger \left\langle g'(\lambda_i)\left[\sum_j h_j^\dagger g(\lambda_j) - \sum_m v_m^\dagger g(\rho_m)\right]\lambda_k\right\rangle \nonumber \\
	&\quad \quad -\alpha_\wb h_k^\dagger \left\langle g'(\lambda_k)\left[\sum_j h_j^\dagger g(\lambda_j) - \sum_m v_m^\dagger g(\rho_m)\right]\lambda_i\right\rangle \nonumber \\
	&\quad \quad \quad+ \alpha_\wb^2 h_i^\dagger h_k^\dagger \left\langle g'(\lambda_i)g'(\lambda_k)\left[\sum_j h_j^\dagger g(\lambda_j) - \sum_m v_m^\dagger g(\rho_m)\right]^2\right\rangle;\\
	\frac{du_{ip}}{d\tau} &= -\alpha_\wb h_i^\dagger \left\langle g'(\lambda_i)\left[\sum_k h_k^\dagger g(\lambda_k) - \sum_m v_m^\dagger g(\rho_m)\right]\eta_p\right\rangle;\\
	\frac{dh_i^\dagger}{d\tau} &= -\alpha_\hb\left\langle\left[\sum_k h_k^\dagger g(\lambda_k) - \sum_m v_m^\dagger g(\rho_m)\right]g(\lambda_i)\right\rangle.         
\end{align}
Similarly to the $I_2$ integral defined in~\autoref{eq: i2}, we further define:
\begin{align}
	&I_3(d, f, h) = \langle g'(\zeta)\beta g(\gamma)\rangle,\\
	&I_4(d, e, f, h) = \langle g'(\zeta)g'(\iota)g(\beta)g(\gamma)\rangle;
\end{align}
where $\zeta, \iota$ are local fields of the student with indices $d, e$; and $\beta, \gamma$ can be local 
fields of either student or teacher with indices $f, h$.
Substituting these definitions into the expanded \gls{ode} formulations gives:
% &J_2(e, f) = \langle g'(e)g'(f)\rangle
% and $e, f$ are local fields of the teacher. 
\begin{align}
	\frac{dr_{in}}{d\tau} &= \alpha_\wb h_i^\dagger\left[\sum_m^Mv_m^*I_3(i,n,m) - \sum_k^Kh_k^\dagger I_3(i,n,k)\right]; \label{eq: oder}\\
	\frac{dq_{ik}}{d\tau} &= \alpha_\wb h_i^\dagger \left[\sum_m^Mv^\dagger_mI_3(i,k,m) - \sum_j^Kh^\dagger_jI_3(i,k,j)\right] \nonumber\\
	& \quad + \alpha_\wb h_k^\dagger \left[\sum_m^Mv^\dagger_mI_3(k,i,m) - \sum_j^Kh^\dagger_jI_3(k,i,j)\right] \nonumber\\
	& \quad \quad +\alpha_\wb^2 h_i^\dagger h_k^\dagger\left[\sum_{j,l}^Kh^\dagger_jh^\dagger_lI_4(i,k,j,l) + \sum_{m,n}^Mv^\dagger_mv^\dagger_nI_4(i,k,m,n)\right. \nonumber \\
	& \quad \quad \quad \quad \quad \quad \left.- 2\sum_j^K\sum_m^Mv^\dagger_mh^\dagger_jI_4(i,k,j,m)\right]; \label{eq: odeq}\\
	\frac{du_{ip}}{d\tau} &= \alpha_\wb h_i^\dagger \left[\sum_m^Mv^\dagger_mI_3(i,p,m) - \sum_k^Kh^\dagger_kI_3(i,p,k)\right]; \label{eq: uipexp}\\
	\frac{dh_i^\dagger}{d\tau} &= \alpha_\hb\left[\sum_m^Mv^\dagger_mI_2(m,i) - \sum_k^Kh^\dagger_kI_2(k,i)\right]. \label{eq: hstar_exp}
\end{align}
This completes the picture for the dynamics of the generalisation error. It can be expressed purely 
in terms of the head weights and the $I$ integrals. For the case of the scaled error function we can 
evaluate the $I_2, I_3$, and $I_4$ analytically meaning we have an exact formulation of the generalisation 
error dynamics of the student with respect to both teachers in the thermodynamic limit. Further details 
on the integrals can be found in~\autoref{app: gaussian_integrals}. The next chapter introduces the 
experimental framework that compliments the theoretical formalism presented above.

\section{Gaussian Integrals under Scaled Error Function}\label{app: gaussian_integrals}

In the derivations of~\autoref{app: derivation}, we introduce a set of integrals over multivariate Gaussian distributions,
labelled $I_2$, $I_3$ and $I_4$. They are defined as:

\begin{align}
    I_2(f, h) &\equiv \langle g(\beta)g(\gamma)\rangle,\\
	I_3(d, f, h) &\equiv \langle g'(\zeta)\beta g(\gamma)\rangle,\\
	I_4(d, e, f, h) &\equiv \langle g'(\zeta)g'(\iota)g(\beta)g(\gamma)\rangle;
\end{align}
where $\zeta, \iota$ are local fields of the student with indices $d, e$; and $\beta, \gamma$ can be local 
fields of either student or teacher with indices $f, h$; and $g$ is the activation function. 

These integrals do not have closed form solutions for the ReLU activation. For the scaled error function however,
they can all be solved analytically. They are given by:

\begin{align}
    I_2 &= \frac{1}{\pi}\arcsin{\frac{c_{12}}{\sqrt{(1 + c_{11})(1 + c_{22})}}};\\
    I_3 &= \frac{2c_{23}(1 + c_{11}) - 2c_{12}c_{13}}{\sqrt{\Lambda_3}(1 + c_{11})};\\
    I_4 &= \frac{4}{\pi^2\sqrt{\Lambda_4}}\arcsin{\frac{\Lambda_0}{\sqrt{\Lambda_1\Lambda_2}}};
\end{align}
where 
\begin{align}
    \Lambda_0 &= \Lambda_4c_{34} - c_{23}c_{24}(1 + c_{11}) - c_{13}c_{14}(1 + c_{22}) + c_{12}c_{13}c_{24} + c_{12}c_{14}c_{23};\\
    \Lambda_1 &= \Lambda_4(1 + c_{33}) - c_{23}^2(1 + c_{11}) - c_{13}^2(1 + c_{22}) + 2c_{12}c_{13}c_{23};\\
    \Lambda_2 &= \Lambda_4(1 + c_{44}) - c_{24}^2(1 + c_{11}) - c_{14}^2(1 + c_{22}) + 2c_{12}c_{14}c_{24};\\
    \Lambda_3 &= (1 + c_{11})(1 + c_{33}) - c_{13}^2;\\
\end{align}
and where $c$ is the relevant projected down covariance matrix.
\section{Overlap Generation}\label{app: overlap_generation}

In~\autoref{sec: feature_sim_exp}, we investigate the effect of task similarity on forgetting. 
In our framework, the teachers act as tasks. From~\autoref{app: derivation}, we know that the learning dynamics
in the student can be fully described by the overlap parameters, which includes the teacher-teacher overlap matrix, $V$.
For our investigation we need a method to generate teachers with specific overlaps; specifically--- 
in the normalised teachers Ansatz, and for teachers with a single hidden unit---we perform simulations over 
the full range of $V$ from 0 to 1. In this configuration we simply need a procedure to generate two $N$-dimensional vectors, 
$\mathbf{v}_1$, $\mathbf{v}_2$, with an angle $\theta$ between them such that:
\begin{equation}
    \mathbf{v}_1\cdot \mathbf{v}_2 = \theta.
\end{equation}
Fortunately there is a standard algorithm for this. First we define two vectors
\[
\tilde{\mathbf{v}}_1 = 
\begin{pmatrix} 
	0\\
	1
\end{pmatrix};
\quad
\tilde{\mathbf{v}}_2 = 
\begin{pmatrix} 
	\sin{\theta}\\
	\cos{\theta}
\end{pmatrix}.
\]
Second, we generate an $N\times N$ orthogonal matrix, $R$. There is a standard sicpy implementation for this
based on QR decomposition of a random Gaussian 
matrix\footnote{\href{https://docs.scipy.org/doc/scipy/reference/generated/scipy.stats.ortho\_group.html}{SciPy Stats Module Docs}}.

Finally, multiply the first two columns of $R$ with either vector to generate the rotated vectors:
\begin{align}
    \mathbf{v}_1 = R[:, 1:2]\cdot \tilde{\mathbf{v}}_1;\\
    \mathbf{v}_2 = R[:, 1:2]\cdot \tilde{\mathbf{v}}_2.
\end{align}
\section{Experiment Details}\label{app: experiment_details}

In this section we provide details of experimental procedures used to obtain 
the graphs and figures presented in this work.

In the ODE limit investigation, the following parameters were used:
\begin{itemize}
    \item Input dimension = 10,000;
    \item Test set size = 50,000;
    \item \gls{sgd} optimiser;
    \item Mean squared error loss;
    \item Teacher weight initialisation: normal distribution with variance 1;
    \item Student weight initialisation: normal distribution with variance 0.001;
    \item Student hidden dimension: 2; 
    \item Teacher hidden dimension: 1;
    \item Learning rate: 1
\end{itemize}

In the mean-field limit investigation the following parameters were used:
\begin{itemize}
    \item Input dimension = 15;
    \item Test set size = 25,000;
    \item \gls{sgd} optimiser;
    \item Mean squared error loss;
    \item Teacher weight initialisation: normal distribution with variance 1;
    \item Student weight initialisation: normal distribution with variance 0.001;
    \item Student hidden dimension: 1000; 
    \item Teacher hidden dimension: 250;
    \item Learning rate: 5
\end{itemize}
\newpage
\section{Forgetting vs. $V$ at Multiple Intervals}\label{app: cross_secs}

\begin{figure}[htp]
\centering
	\pgfplotsset{
		width=0.21\textwidth,
		height=0.16\textwidth,
		every tick label/.append style={font=\tiny},
		xlabel={\small $V$},
		yticklabel style={
        /pgf/number format/fixed,
        /pgf/number format/precision=5
        },
        scaled y ticks=false
		}
	\begin{tikzpicture}
% 		\node at (2.5, 3.8) {\small ReLU};
		\begin{axis}
			[
			xmin=0, xmax=1,
			ymin=-0.004, ymax=0.004,
			ylabel={\small $\forgetting{0}$}
		]
		\addplot graphics [xmin=0, xmax=1,ymin=-0.004,ymax=0.004] {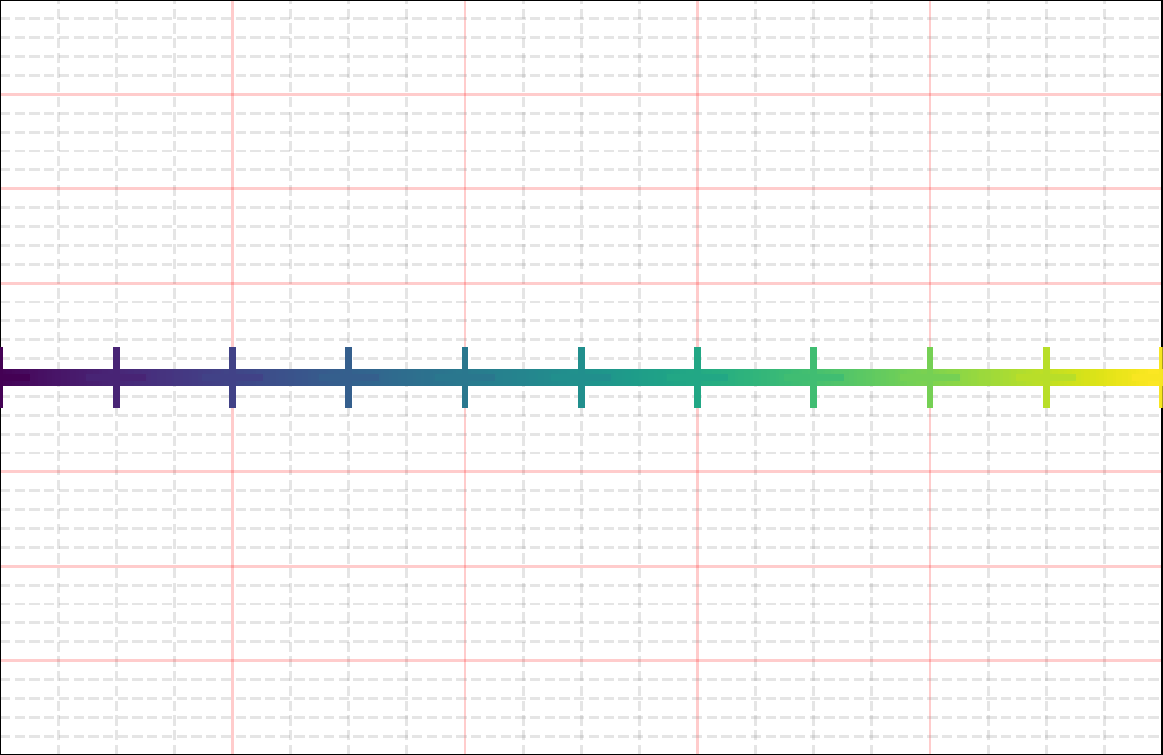};
		\end{axis}
	\end{tikzpicture}%
		\hspace{1em}
		\begin{tikzpicture}
% 		\node at (2.5, 3.8) {\small ReLU};
		\begin{axis}
			[
			xmin=0, xmax=1,
			ymin=0, ymax=0.005,
			ylabel={\small $\forgetting{10}$}
		]
		\addplot graphics [xmin=0, xmax=1,ymin=0,ymax=0.005] {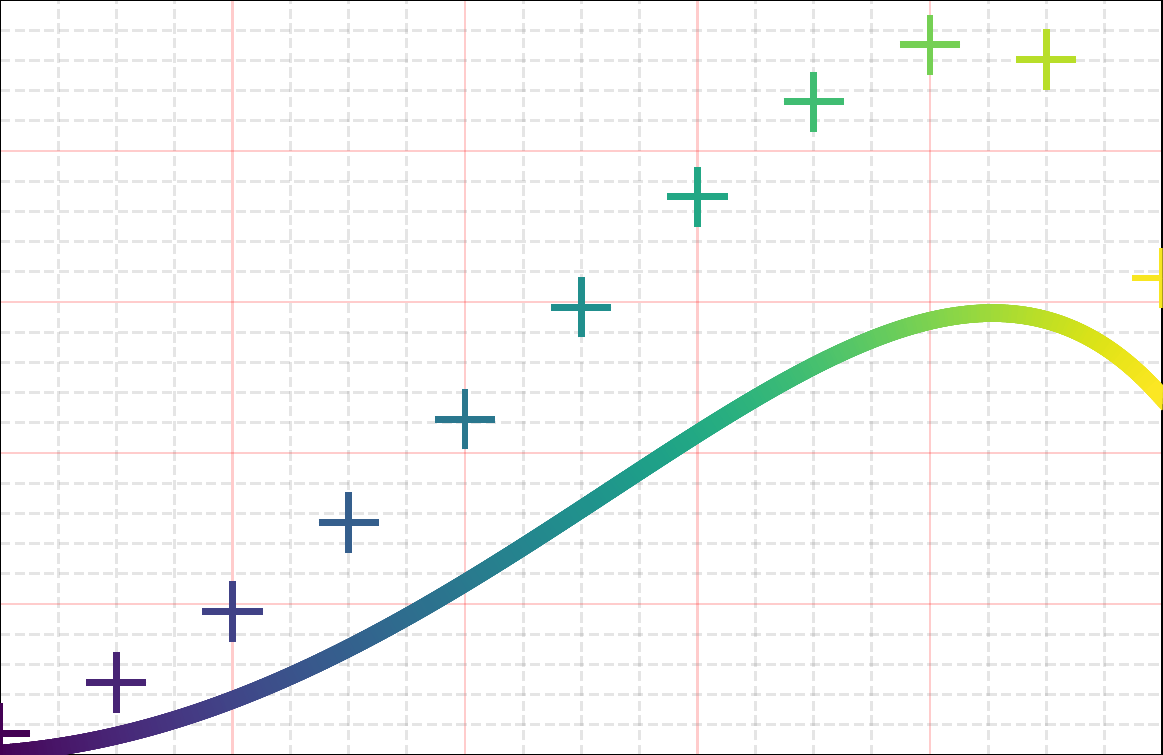};
		\end{axis}
	\end{tikzpicture}%
	\vspace{0.5em}
		\begin{tikzpicture}
% 		\node at (2.5, 3.8) {\small ReLU};
		\begin{axis}
			[
			xmin=0, xmax=1,
			ymin=0, ymax=0.035,
			ylabel={\small $\forgetting{20}$}
		]
		\addplot graphics [xmin=0, xmax=1,ymin=0,ymax=0.035] {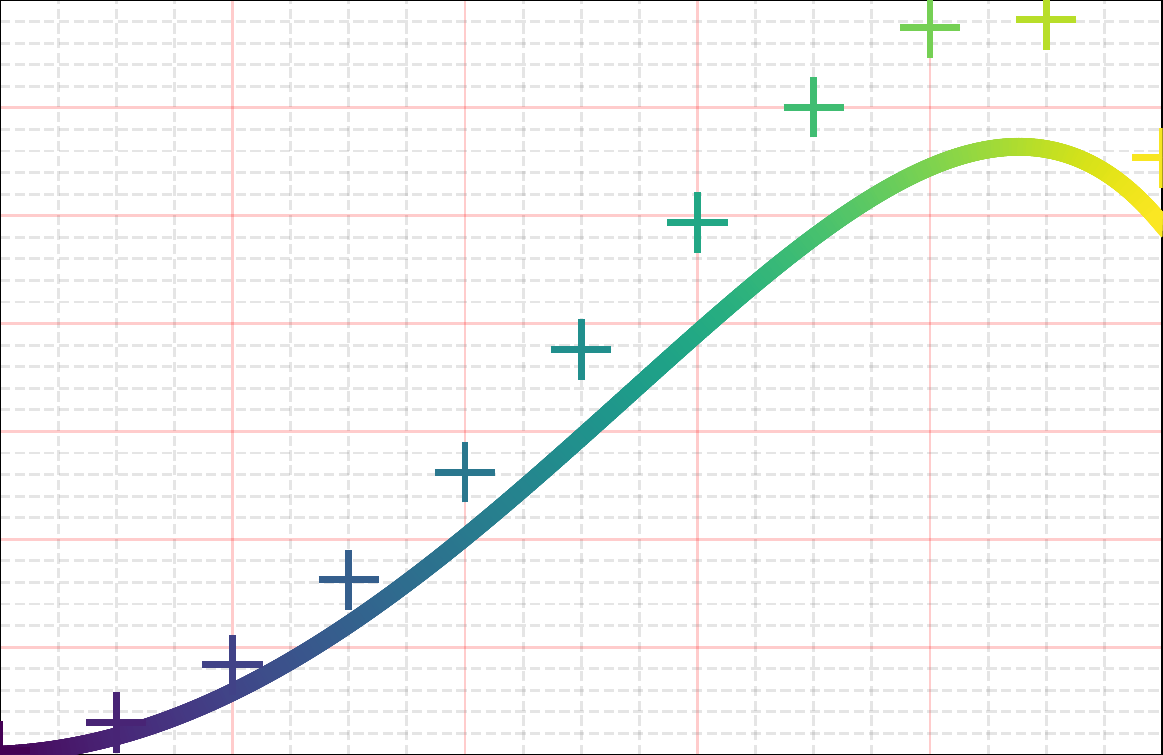};
		\end{axis}
	\end{tikzpicture}%
    \hspace{1em}
    	\begin{tikzpicture}
% 		\node at (2.5, 3.8) {\small ReLU};
		\begin{axis}
			[
			xmin=0, xmax=1,
			ymin=0, ymax=0.35,
			ylabel={\small $\forgetting{50}$}
		]
		\addplot graphics [xmin=0, xmax=1,ymin=0,ymax=0.35] {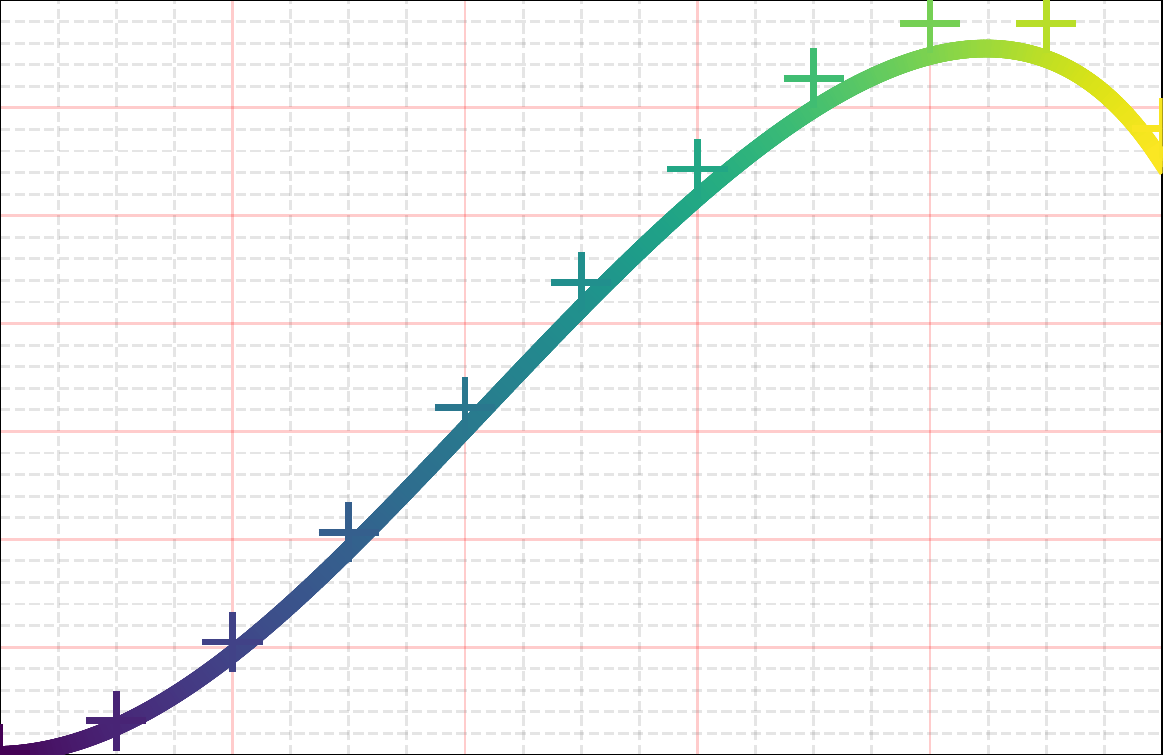};
		\end{axis}
	\end{tikzpicture}%
	\vspace{0.5em}
	\begin{tikzpicture}
% 		\node at (2.5, 3.8) {\small ReLU};
		\begin{axis}
			[
			xmin=0, xmax=1,
			ymin=0, ymax=1.2,
			ylabel={\small $\forgetting{100}$}
		]
		\addplot graphics [xmin=0, xmax=1,ymin=0,ymax=1.2] {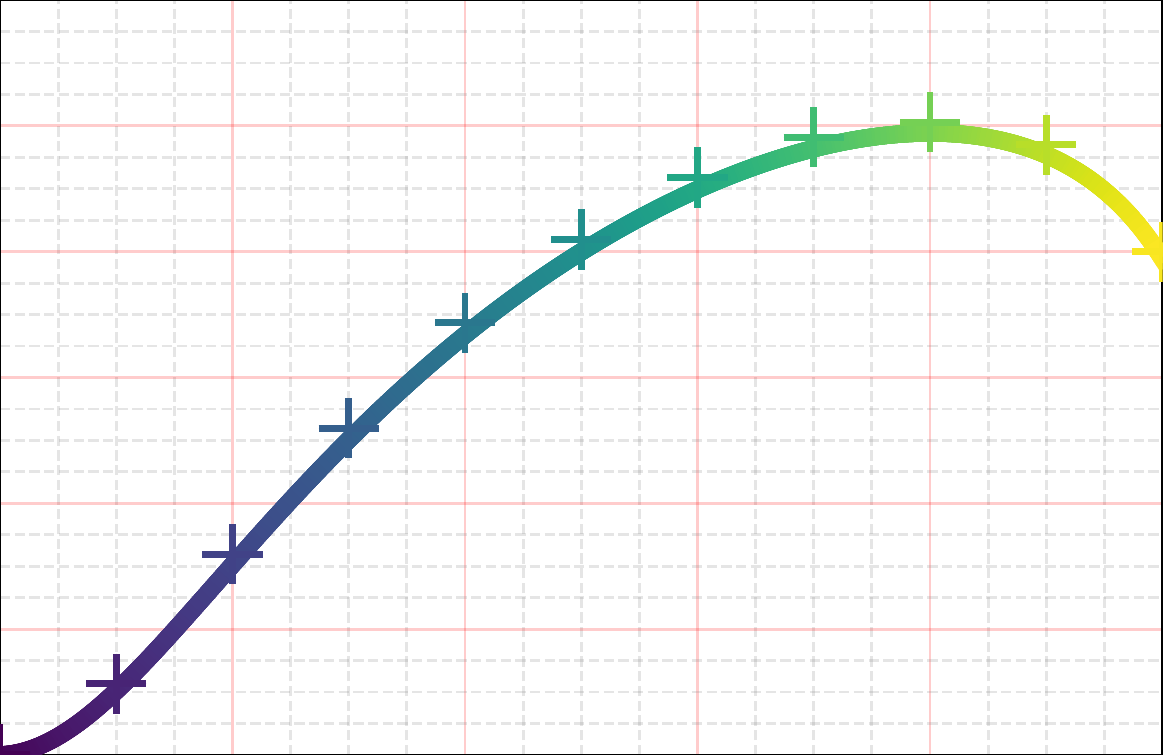};
		\end{axis}
	\end{tikzpicture}%
    \hspace{1em}
    	\begin{tikzpicture}
% 		\node at (2.5, 3.8) {\small ReLU};
		\begin{axis}
			[
			xmin=0, xmax=1,
			ymin=0, ymax=2,
			ylabel={\small $\forgetting{200}$}
		]
		\addplot graphics [xmin=0, xmax=1,ymin=0,ymax=2] {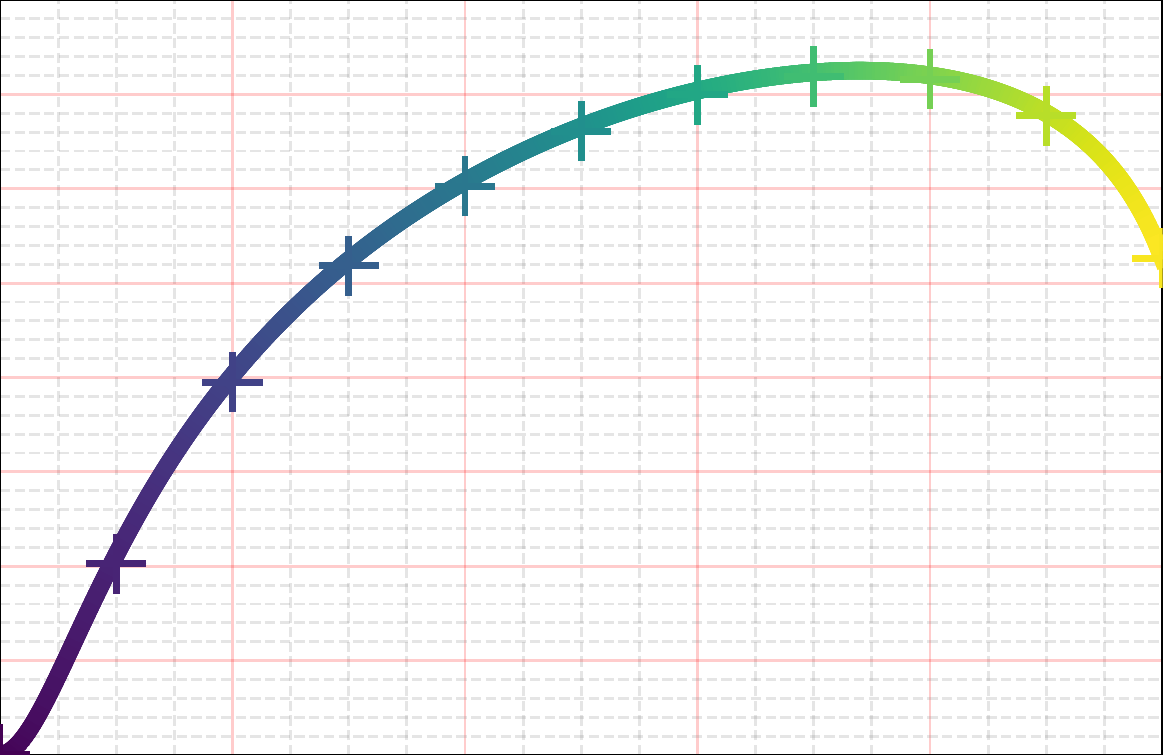};
		\end{axis}
	\end{tikzpicture}%
	\vspace{0.5em}
	\begin{tikzpicture}
% 		\node at (2.5, 3.8) {\small ReLU};
		\begin{axis}
			[
			xmin=0, xmax=1,
			ymin=0, ymax=3,
			ylabel={\small $\forgetting{500}$}
		]
		\addplot graphics [xmin=0, xmax=1,ymin=0,ymax=3] {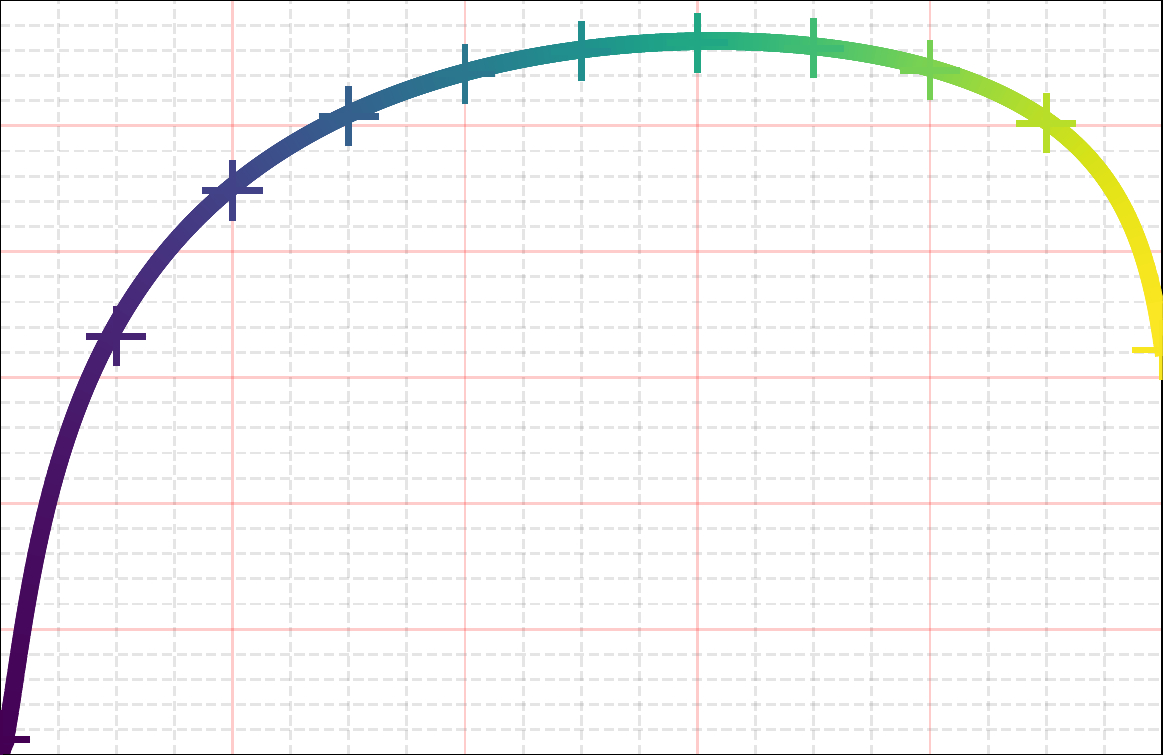};
		\end{axis}
	\end{tikzpicture}%
    \hspace{1em}
    	\begin{tikzpicture}
% 		\node at (2.5, 3.8) {\small ReLU};
		\begin{axis}
			[
			xmin=0, xmax=1,
			ymin=0, ymax=3.5,
			ylabel={\small $\forgetting{1000}$}
		]
		\addplot graphics [xmin=0, xmax=1,ymin=0,ymax=3.5] {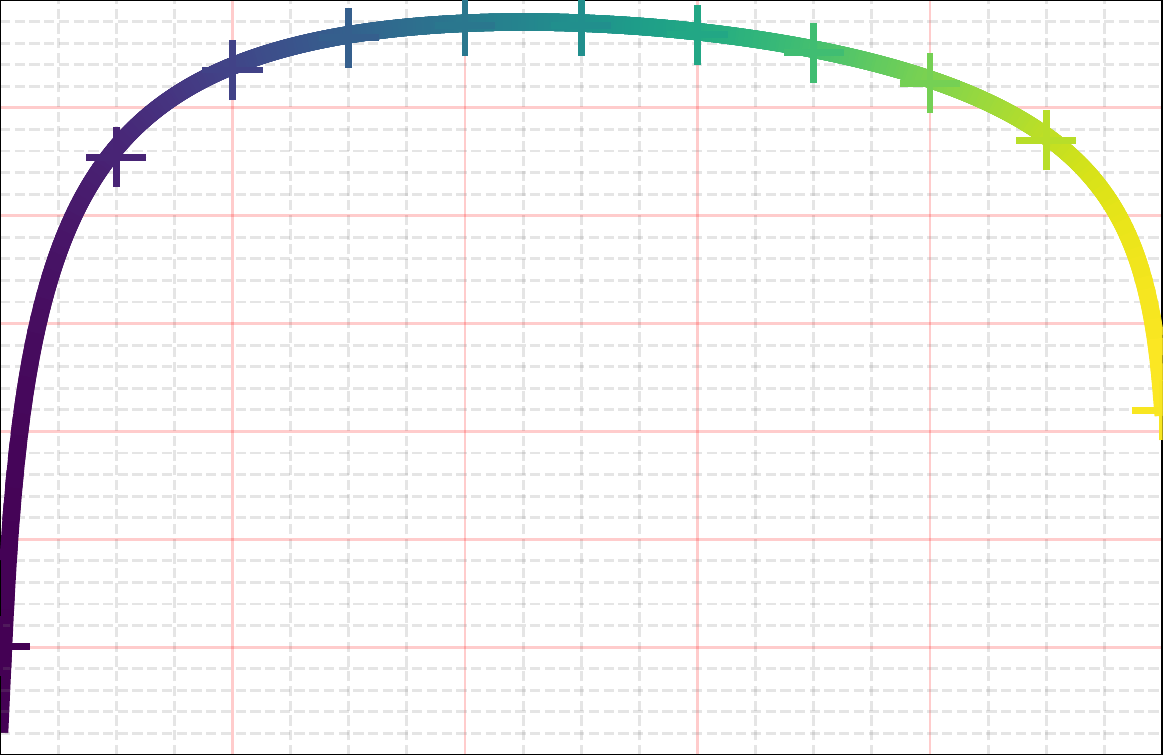};
		\end{axis}
	\end{tikzpicture}%
	\hspace{2em}
	\begin{tikzpicture}
% 		\node at (2.5, 3.8) {\small ReLU};
		\begin{axis}
			[
			xmin=0, xmax=1,
			ymin=1.4, ymax=3.7,
			ylabel={\small $\forgetting{2000}$}
		]
		\addplot graphics [xmin=0, xmax=1,ymin=1.4,ymax=3.7] {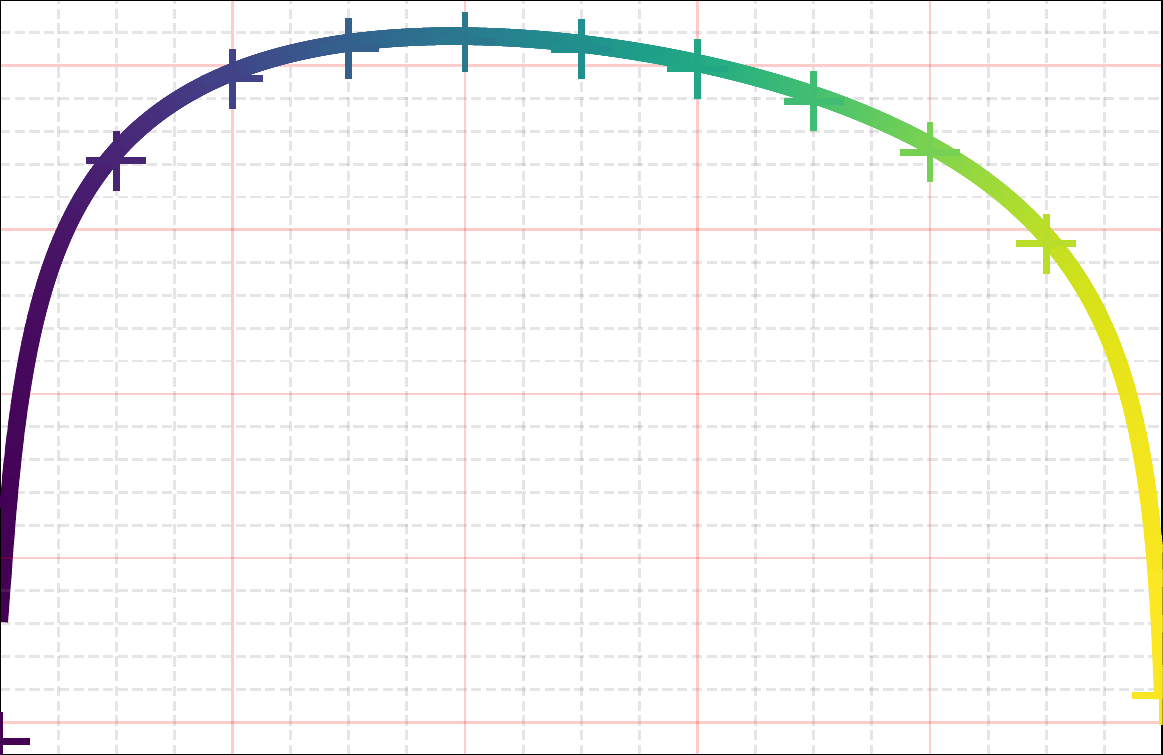};
		\end{axis}
	\end{tikzpicture}%
    \hspace{1em}
    	\begin{tikzpicture}
% 		\node at (2.5, 3.8) {\small ReLU};
		\begin{axis}
			[
			xmin=0, xmax=1,
			ymin=1.4, ymax=3.7,
			ylabel={\small $\forgetting{5000}$}
		]
		\addplot graphics [xmin=0, xmax=1,ymin=1.4,ymax=3.7] {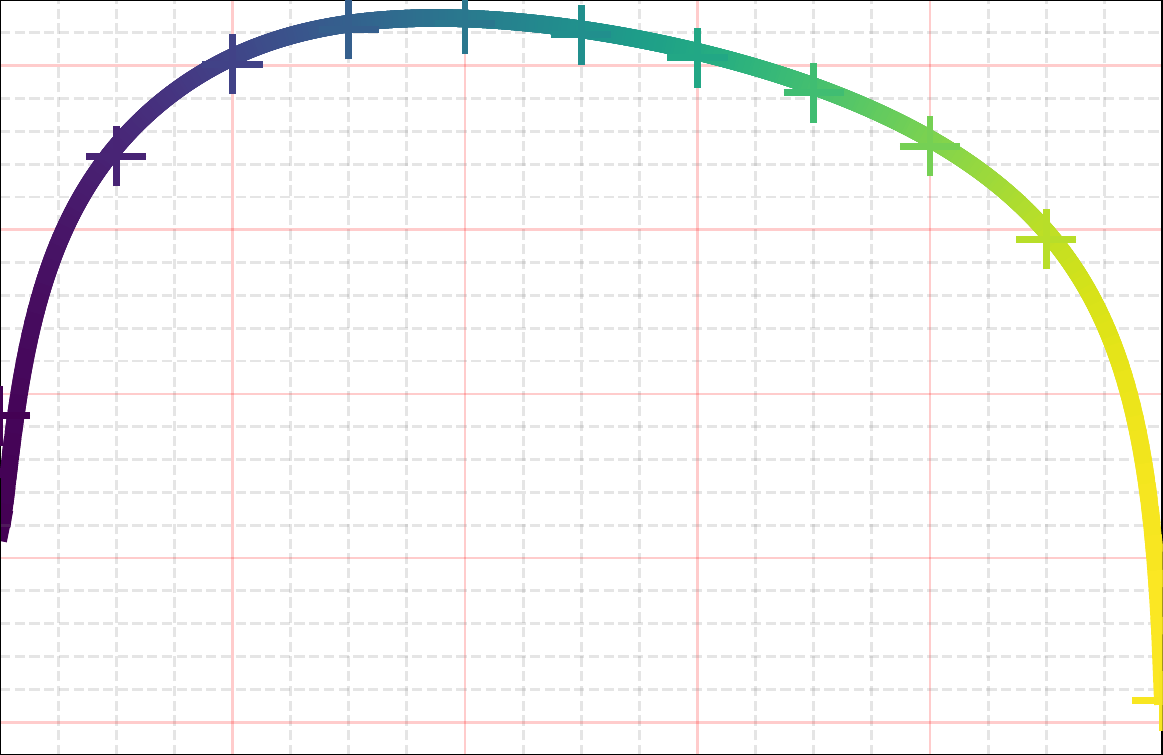};
		\end{axis}
	\end{tikzpicture}%
	\vspace{0.5em}
	\begin{tikzpicture}
% 		\node at (2.5, 3.8) {\small ReLU};
		\begin{axis}
			[
			xmin=0, xmax=1,
			ymin=1.5, ymax=3.8,
			ylabel={\small $\forgetting{10000}$}
		]
		\addplot graphics [xmin=0, xmax=1,ymin=1.5,ymax=3.8] {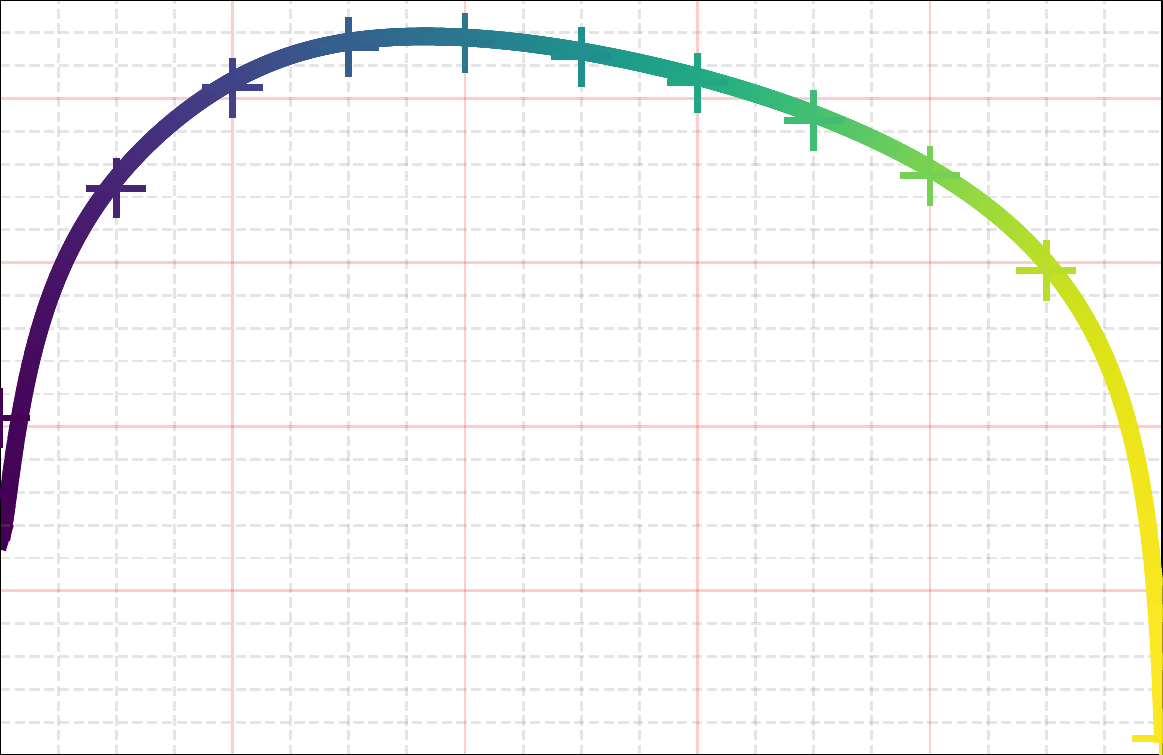};
		\end{axis}
	\end{tikzpicture}%
    \hspace{1em}
    	\begin{tikzpicture}
% 		\node at (2.5, 3.8) {\small ReLU};
		\begin{axis}
			[
			xmin=0, xmax=1,
			ymin=1.5, ymax=3.8,
			ylabel={\small $\forgetting{20000}$}
		]
		\addplot graphics [xmin=0, xmax=1,ymin=1.5,ymax=3.8] {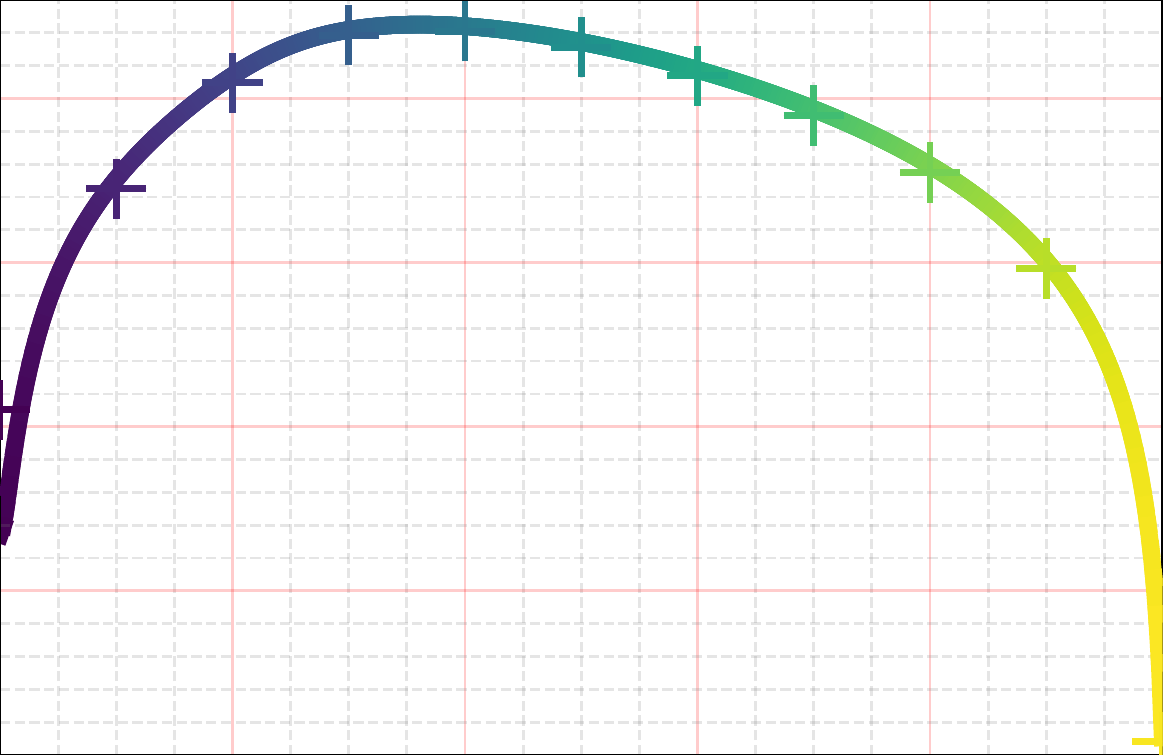};
		\end{axis}
	\end{tikzpicture}%
	\caption[Aggregate forgetting, $\forgetting{t}$, vs. teacher-teacher overlap, $V$, at 
	different time intervals post task-switch.]{Aggregate forgetting, $\forgetting{t}$, vs. 
	teacher-teacher overlap, $V$, at different time intervals post task-switch. A teacher-teacher 
	overlap of 0 corresponds to orthogonal teacher weight vectors, whereas a teacher-teacher overlap 
	of 1 corresponds to aligned teacher weight vectors. Forgetting is strongest for teachers that are 
	intermediately correlated, while the student is relatively robust to forgetting for aligned or 
	orthogonal teachers. The distribution of error changes moves significantly as time spent training 
	on the new task increases.}\label{fig: cross_sections}
\end{figure}

In~\autoref{fig: empirical_task_sim}, we show the cross section of forgetting vs. $V$ at a set of intervals after the task boundary. In~\autoref{fig: cross_sections}, we show this cross-section at a greater range of time delays after the switch.

\section{Forgetting vs. Feature Similarity, \gls{relu} Networks}\label{app: relu_networks}

\begin{figure}[!h]
    \centering
    % \hspace{1em}
    \pgfplotsset{
		width=0.4\textwidth,
		height=0.32\textwidth,
		xlabel={$s$},
		}
    \begin{tikzpicture}
        \node at (-1.8, 3.9) {\small $V$};
        \node at (-2.15, 3.7) {\small 1};
        \node at (-2.15, -0.2) {\small 0};
	    \node [opacity=1] at (-1.8, 1.7) {\includegraphics[width=0.23\textwidth, angle=90]{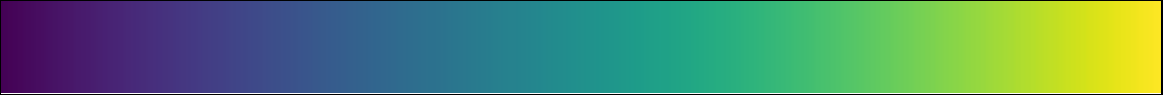}}; 
        \hspace{1em}
        \begin{axis}
            [
            scaled x ticks = true,
            xmin=0, xmax=100000,
            ymin=-8, ymax=0,
            ylabel={$\log{\epsilon^\dagger}$}
        ]
        \addplot graphics [xmin=0, xmax=150000,ymin=-12,ymax=0] {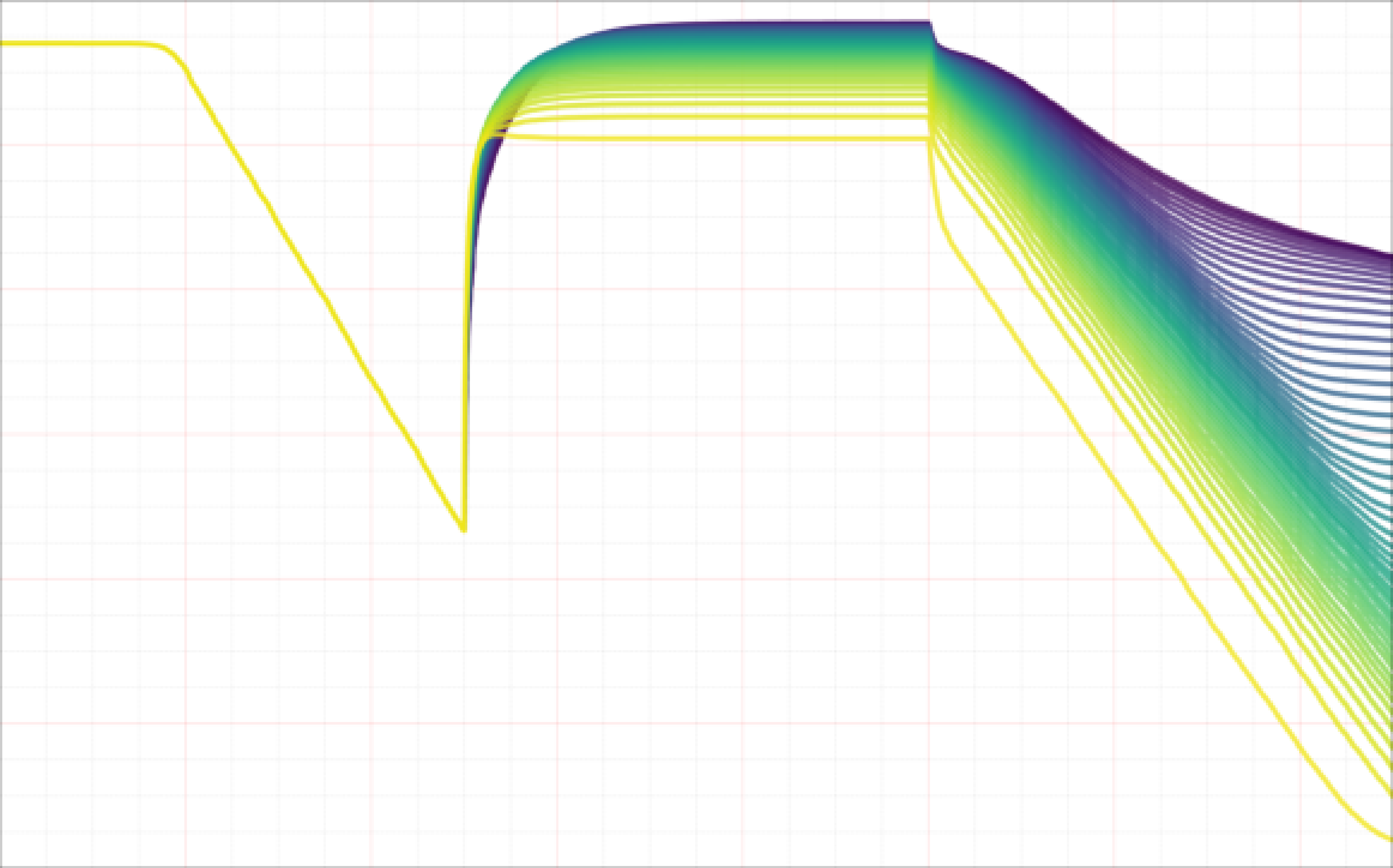};
        \end{axis}
    \end{tikzpicture}%
    \caption[$\epsilon^\dagger$ generalisation error evolutions for a range of teacher-teacher overlaps, $V$ for \gls{relu} networks.]{Generalisation 
    error with respect to first teacher, $\log{\epsilon^\dagger}$, vs. timestep, $s$, for a range of teacher-teacher 
    overlaps for \gls{relu} networks. Task switches occur at steps 50,000 and 100,000. This plot is the \gls{relu} equivalent 
    of~\autoref{fig: empirical_task_sim} in~\autoref{sec: feature_sim_exp}}\label{fig: relu_forgetting}
\end{figure}

\begin{figure}[h!]
\centering
	\pgfplotsset{
		width=0.21\textwidth,
		height=0.16\textwidth,
		xlabel={\small $V$},
		yticklabel style={
        /pgf/number format/fixed,
        /pgf/number format/precision=5
        },
        every tick label/.append style={font=\tiny},
        scaled y ticks=false
		}
	\begin{tikzpicture}
% 		\node at (2.5, 3.8) {\small ReLU};
		\begin{axis}
			[
			xmin=0, xmax=1,
			ymin=-0.0005, ymax=0.0035,
			ylabel={\small $\forgetting{10}$}
		]
		\addplot graphics [xmin=0, xmax=1,ymin=-0.0005,ymax=0.0035] {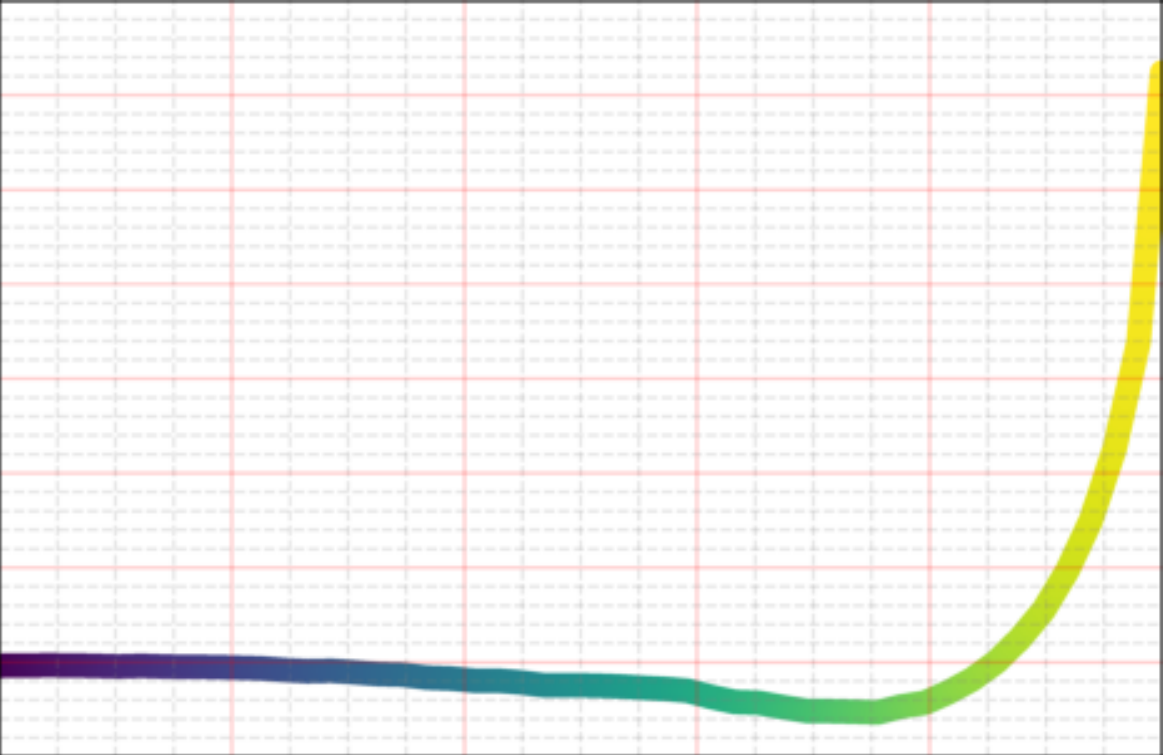};
		\end{axis}
	\end{tikzpicture}%
		\hspace{1em}
		\begin{tikzpicture}
% 		\node at (2.5, 3.8) {\small ReLU};
		\begin{axis}
			[
			xmin=0, xmax=1,
			ymin=0.8, ymax=2.2,
			ylabel={\small $\forgetting{100}$}
		]
		\addplot graphics [xmin=0, xmax=1,ymin=0.8,ymax=2.2] {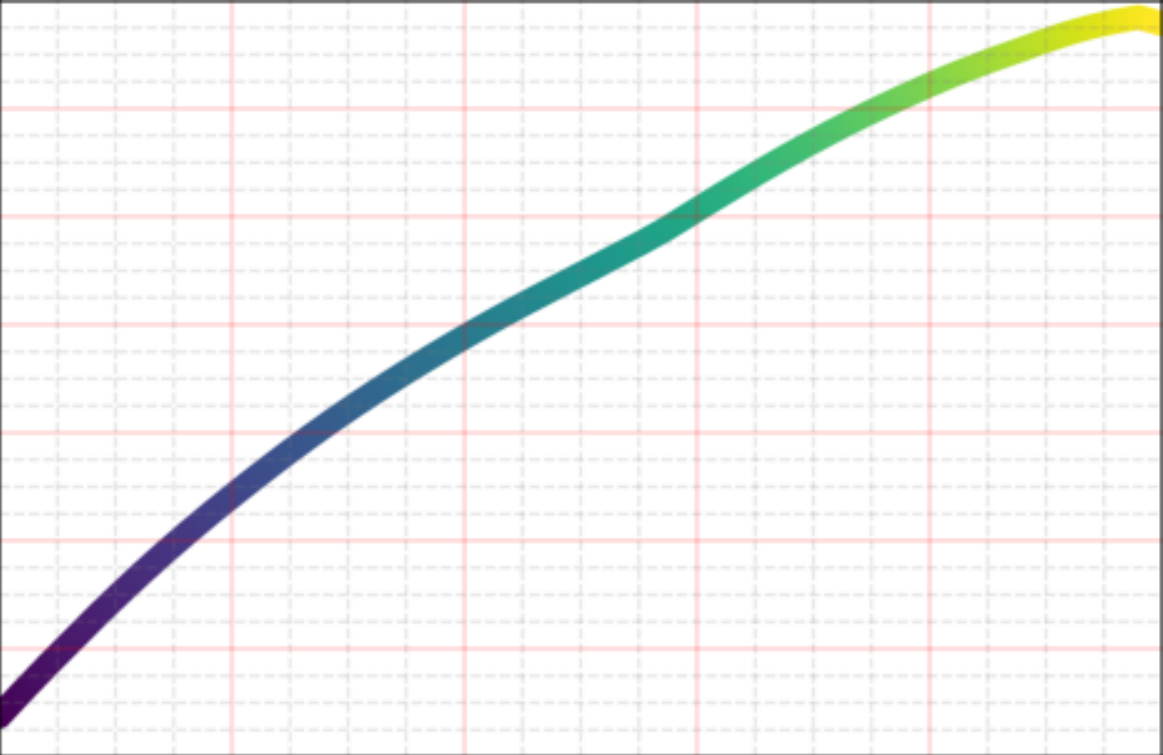};
		\end{axis}
	\end{tikzpicture}%
	\vspace{0.5em}
		\begin{tikzpicture}
% 		\node at (2.5, 3.8) {\small ReLU};
		\begin{axis}
			[
			xmin=0, xmax=1,
			ymin=3, ymax=4.2,
			ylabel={\small $\forgetting{500}$}
		]
		\addplot graphics [xmin=0, xmax=1,ymin=3,ymax=4.2] {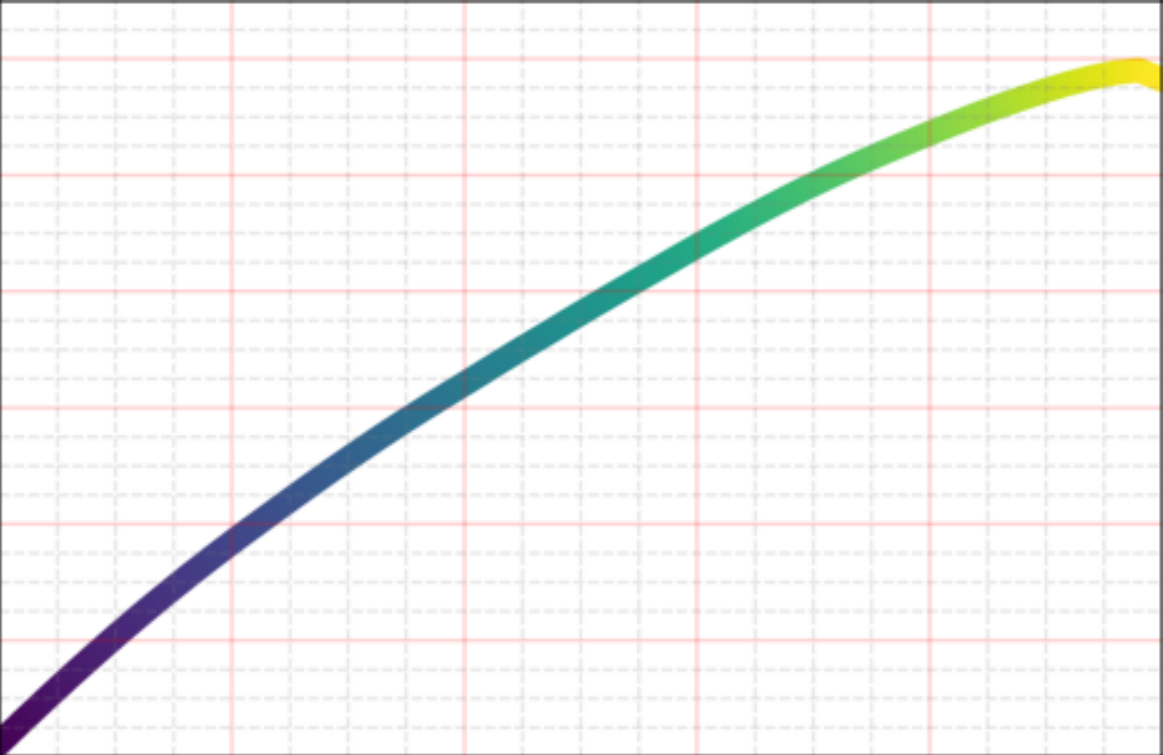};
		\end{axis}
	\end{tikzpicture}%
    \hspace{1em}
    \begin{tikzpicture}
        % 		\node at (2.5, 3.8) {\small ReLU};
                \begin{axis}
                    [
                    xmin=0, xmax=1,
                    ymin=3.8, ymax=5,
                    ylabel={\small $\forgetting{1000}$}
                ]
                \addplot graphics [xmin=0, xmax=1,ymin=3.8,ymax=5] {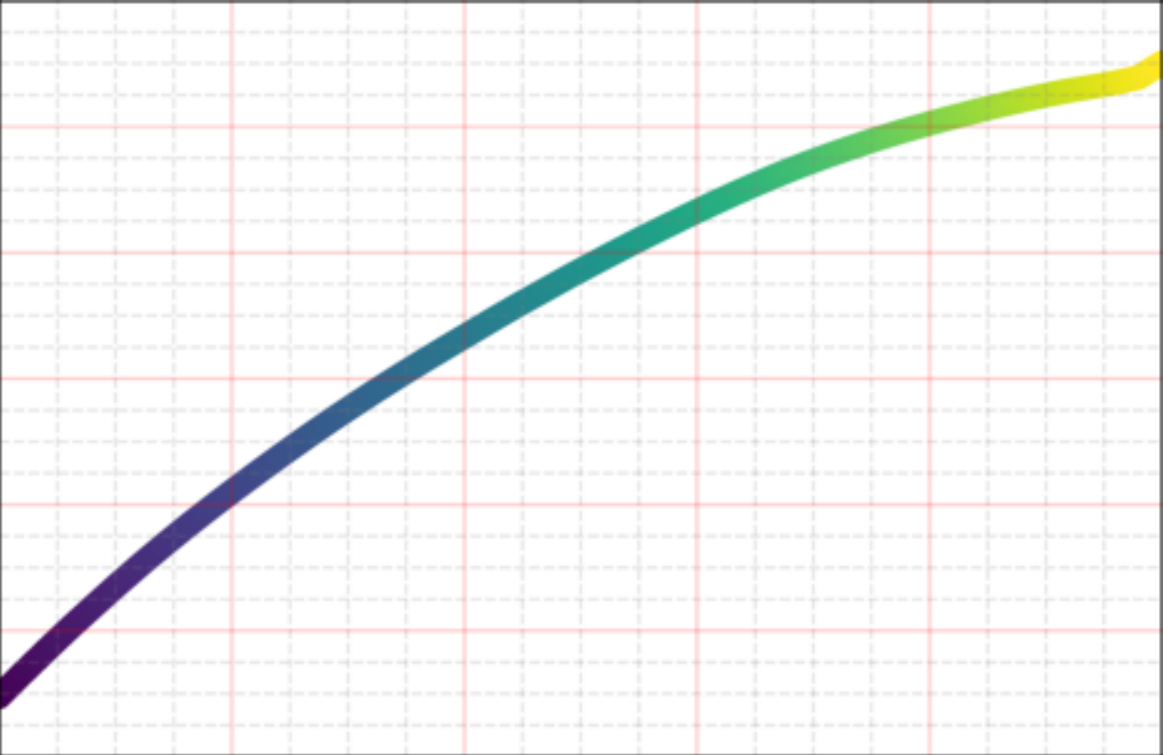};
                \end{axis}
            \end{tikzpicture}%
	\vspace{0.5em}
	\begin{tikzpicture}
% 		\node at (2.5, 3.8) {\small ReLU};
		\begin{axis}
			[
			xmin=0, xmax=1,
			ymin=4.6, ymax=5.5,
			ylabel={\small $\forgetting{2000}$}
		]
		\addplot graphics [xmin=0, xmax=1,ymin=4.6,ymax=5.5] {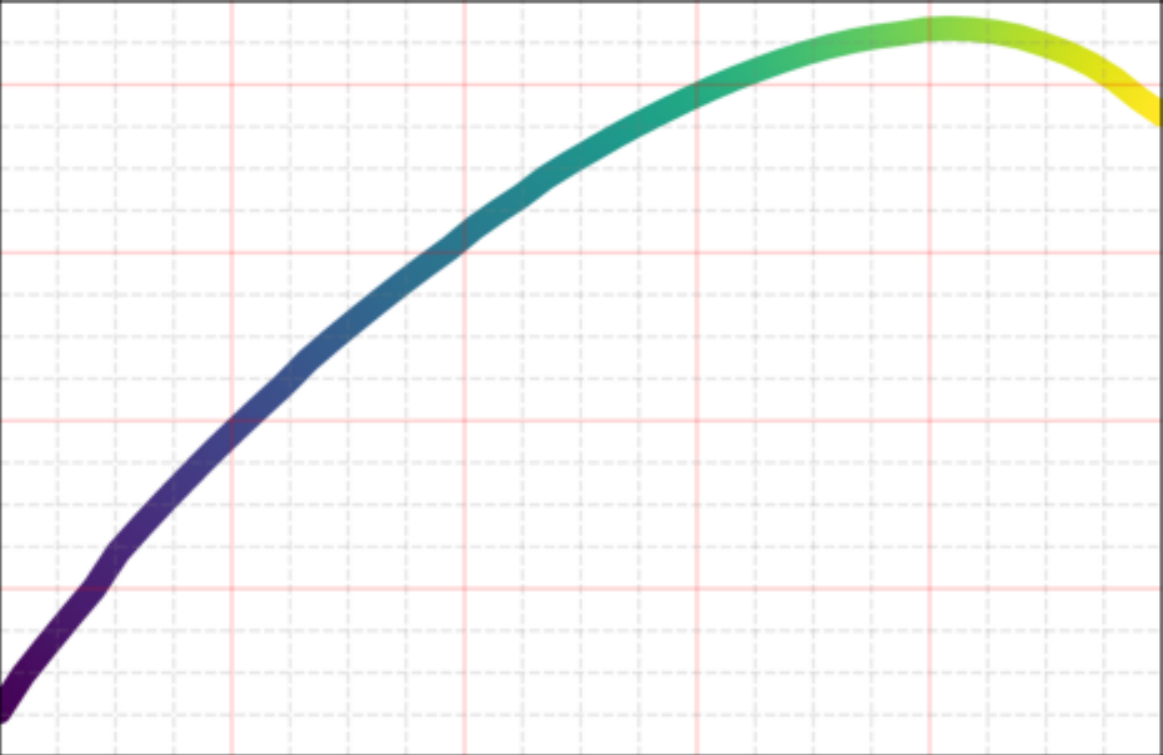};
		\end{axis}
	\end{tikzpicture}%
    \hspace{1em}
    	\begin{tikzpicture}
% 		\node at (2.5, 3.8) {\small ReLU};
		\begin{axis}
			[
			xmin=0, xmax=1,
			ymin=5.4, ymax=6.2,
			ylabel={\small $\forgetting{5000}$}
		]
		\addplot graphics [xmin=0, xmax=1,ymin=5.4,ymax=6.2] {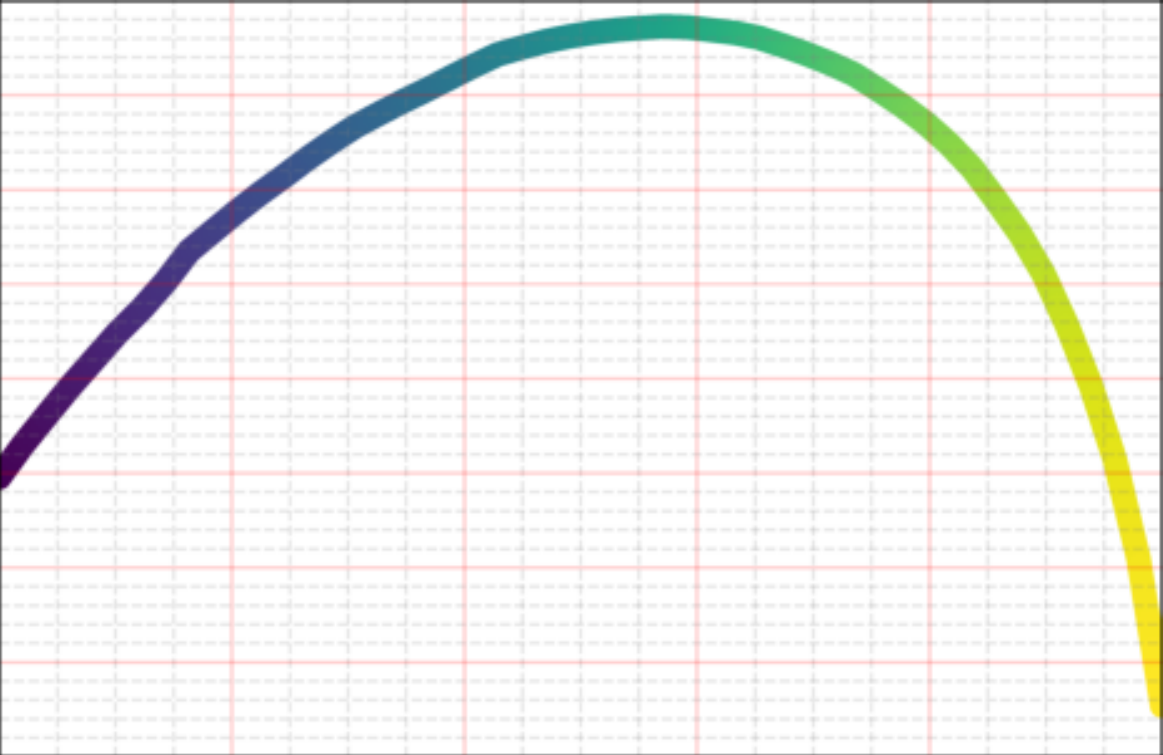};
		\end{axis}
	\end{tikzpicture}%
	\vspace{0.5em}
	\begin{tikzpicture}
% 		\node at (2.5, 3.8) {\small ReLU};
		\begin{axis}
			[
			xmin=0, xmax=1,
			ymin=5.4, ymax=6.7,
			ylabel={\small $\forgetting{10,000}$}
		]
		\addplot graphics [xmin=0, xmax=1,ymin=5.4,ymax=6.7] {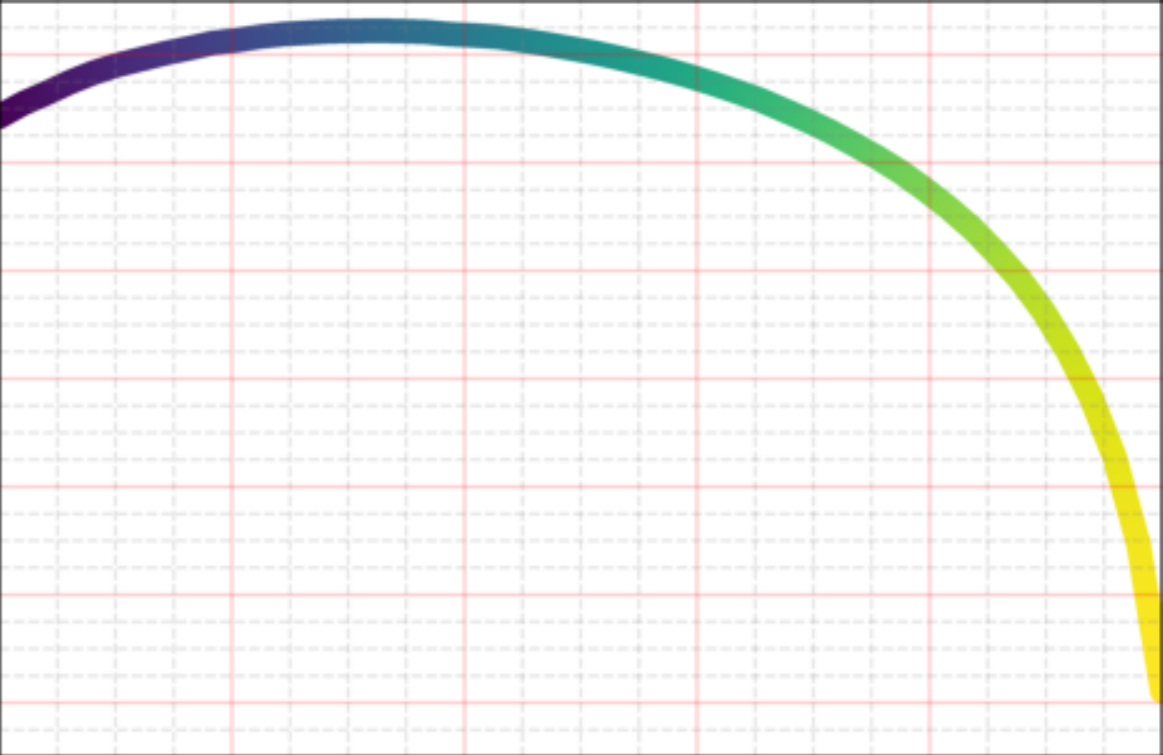};
		\end{axis}
	\end{tikzpicture}%
    \hspace{1em}
    	\begin{tikzpicture}
% 		\node at (2.5, 3.8) {\small ReLU};
		\begin{axis}
			[
			xmin=0, xmax=1,
			ymin=5.3, ymax=7.2,
			ylabel={\small $\forgetting{50,000}$}
		]
		\addplot graphics [xmin=0, xmax=1,ymin=5.3,ymax=7.2] {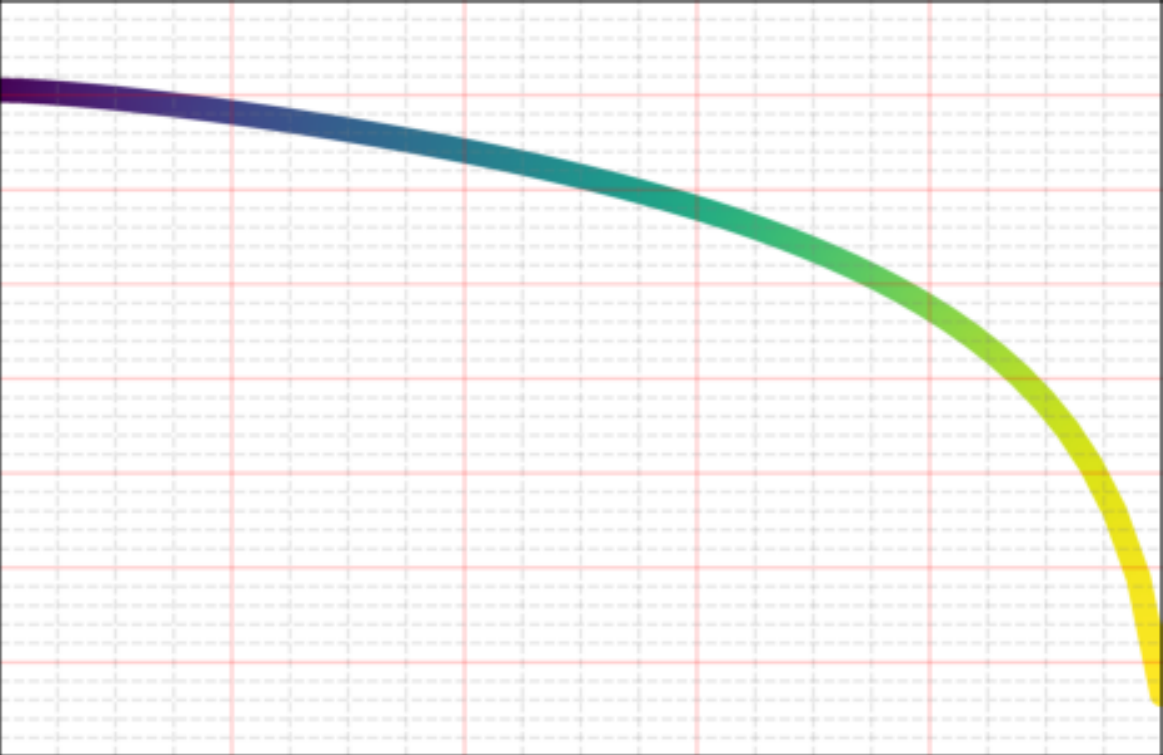};
		\end{axis}
	\end{tikzpicture}%
	\caption[Aggregate forgetting, $\forgetting{t}$, vs. teacher-teacher overlap, $V$, at 
	different time intervals post task-switch for \gls{relu} networks.]{Aggregate forgetting, $\forgetting{t}$, vs. 
    teacher-teacher overlap, $V$, at different time intervals post task-switch for \gls{relu} networks. 
    The distribution of error changes moves faster compared to the sigmoid case in~\autoref{sec: feature_sim_exp}. By the 
    second task switch, the function is monotonic.}\label{fig: relu_cross_sections}
\end{figure}

This sections contains the same experiments as those presented in~\autoref{sec: feature_sim_exp}, 
but for networks with \gls{relu} nonlinearities.~\autoref{fig: relu_forgetting} shows for various values of 
$V$ the generalisation error of the first teacher over time.~\autoref{fig: relu_cross_sections} shows the 
cross sections of forgetting vs. $V$ at various time intervals after the task switch.

\newpage
\section{Effect of Activation \& Distribution Evolution}\label{app: activation_evolution}

The learning dynamics and corresponding forgetting/transfer distributions for 
varying teacher-teacher overlaps presented above are for sigmoidal activation functions. 
In our investigations we found that different activation functions can have a strong impact 
on how the forgetting vs. teacher-teacher overlap distributions change over time. 
In particular, in the \gls{relu} case, the distribution moves relatively quickly from a hump 
curve (seen in the sigmoidal case) to a monotonic function, where the higher overlaps lead 
to less forgetting. Detailed plots for the \gls{relu} case are shown in~\autoref{app: relu_networks}. 
Forgetting and transfer are not stationary attributes, hence the inclusion of a time 
component in our definitions of these quantities. The unsurprising observation that the 
distribution of forgetting over different overlaps changes as time progresses beyond the 
switch point is not discussed in previous research. The nature of this evolution and its 
contributing factors are worthy of further investigation.
\section{First Task Convergence}\label{app: first_task_convergence}

\begin{figure}[!ht]
\centering
    % \hspace{1em}
    \pgfplotsset{
		width=0.4\textwidth,
		height=0.32\textwidth,
		xlabel={$s$},
		}
    \begin{tikzpicture}
        \node at (-1.8, 3.9) {\small $V$};
        \node at (-2.15, 3.7) {\small 1};
        \node at (-2.15, -0.2) {\small 0};
	    \node [opacity=1] at (-1.8, 1.7) {\includegraphics[width=0.23\textwidth, angle=90]{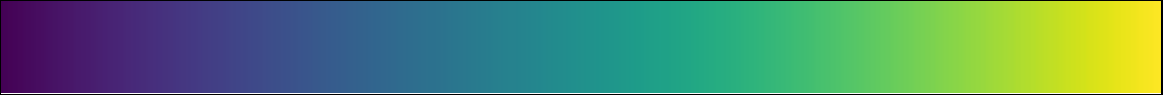}}; 
	    \hspace{1em}
		\begin{axis}
			[
			scaled x ticks = true,
			xmin=0, xmax=70000,
			ymin=-3.5, ymax=-0.5,
			ylabel={$\log{\epsilon^*}$}
		]
		\addplot graphics [xmin=0, xmax=70000,ymin=-3.5,ymax=-0.5] {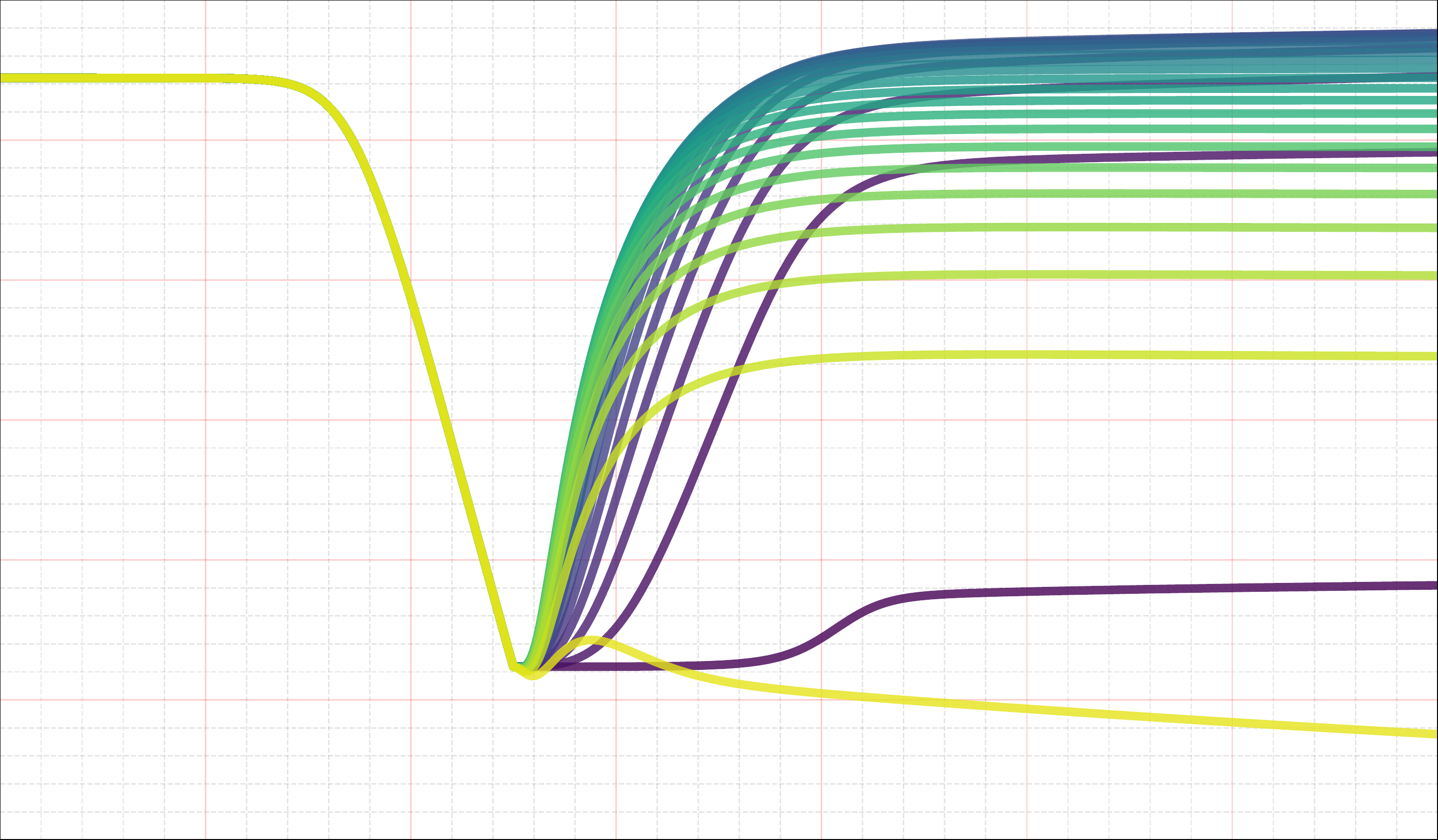};
		\end{axis}
	\end{tikzpicture}%
	\caption{Generalisation 
	error with respect to first teacher, $\log{\epsilon^*}$, vs. timestep, $s$, for a range of teacher-teacher 
	overlaps, $V$. Here the task switch occurs relatively early---before convergence on the first task. Unlike in settings where better convergence has been achieved the initial rate of forgetting is not largest for highest overlap. In fact here there is a period of co-learning just after the task-switch.}\label{fig: less_convergence}
\end{figure}

The setting we work in throughout our experimentation is one in which good convergence has been achieved on the first task before the switch. Some of the observations we make are therefore conditional on this convergence. We show below in~\autoref{fig: less_convergence} one example of a phenomenon (higher rate of forgetting for greater task similarity) we observed in the main results that does not hold in settings where lesser convergence is achieved on the first task.

\section{Forgetting/Transfer Metrics Procedure (Mean-Field Limit)}\label{app: 2d_metric_procedure}

In~\autoref{fig: 2d_aggregate}, we present metrics of forgetting and transfer for various task similarity configurations averaged over 50 random seeds. Specifically we give the initial rates, maxima, and long-time values. Here we provide details on how these are evaluated.

\subsection{Initial Rate}

Let $\tilde{s}$ be the training step at which the teacher switches. We approximate the initial rate of forgetting as:
\begin{equation}
    \frac{1}{N}\sum_{i=1}^N \epsilon^\dagger|_{\tilde{s} + i} - \epsilon^\dagger|_{t=\tilde{s} + i - 1},
\end{equation}
where $N$ is the number of steps over which we take the average change ($N=20$ for experiments shown in~\autoref{fig: 2d_aggregate}). Since we are not using the ODE solutions but pure simulation of the mean-field limit in~\autoref{fig: 2d_aggregate}, such a sampling is necessary to accurately approximate the rates. Likewise the initial rate of transfer is computed via:
\begin{equation}
    \frac{1}{N}\sum_{i=1}^N \epsilon^\ddag|_{\tilde{s} + i - 1} - \epsilon^\ddag|_{t=\tilde{s} + i}.
\end{equation}

\subsection{Maxima}

The maximum forgetting and transfer amounts are computed with
\begin{align}
    \max_{t} (\epsilon^\dagger|_{\tilde{s} + t}) - \epsilon^\dagger|_{\tilde{s}} \quad \text{and} \quad \epsilon^\dagger|_{\tilde{s}} - \min_{t} (\epsilon^\ddag|_{\tilde{s} + t}).
\end{align}

\subsection{Long-Time Limit}

\begin{figure}[!ht]
\centering
    % \hspace{1em}
    \pgfplotsset{
		width=0.5\textwidth,
		height=0.4\textwidth,
		xlabel={$s$},
		}
    \begin{tikzpicture}
		\begin{axis}
			[
			scaled x ticks = true,  
			xmin=0, xmax=3000000,
			ymin=-3.5, ymax=-0.5,
			ylabel={$\log\epsilon$}
		]
		\addplot graphics [xmin=0, xmax=3000000,ymin=-3.5,ymax=-0.5] {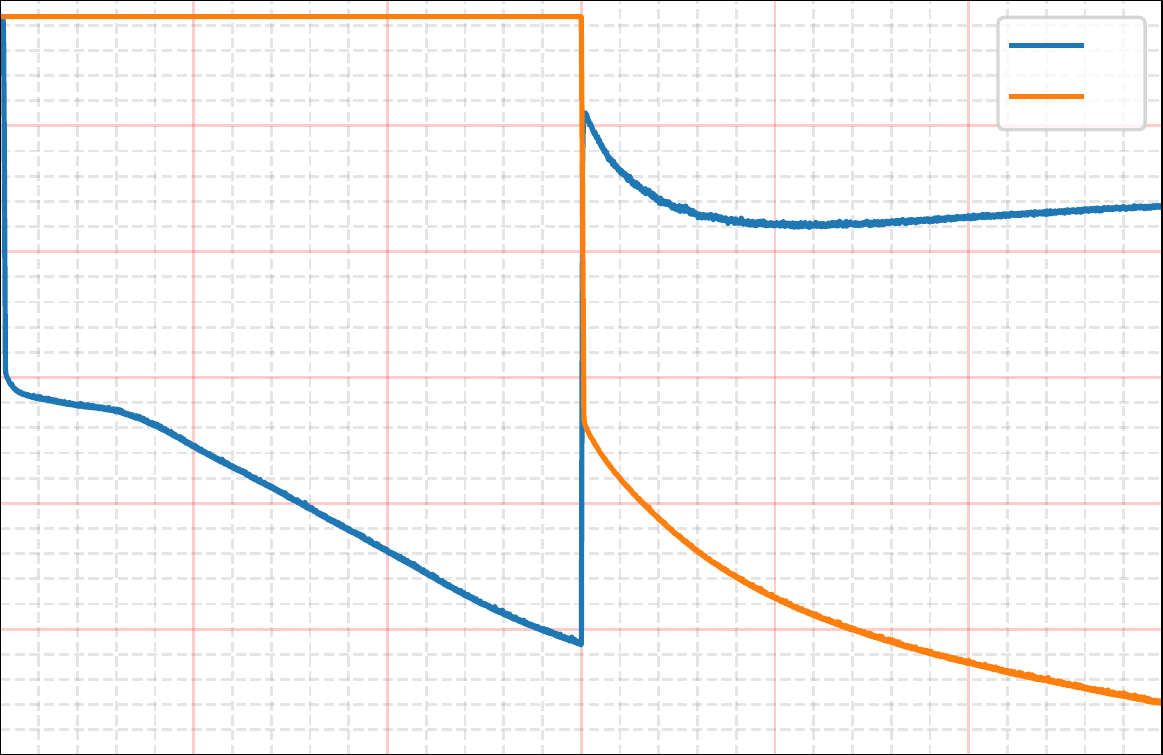};
		\end{axis}
		\node at (6.7, 4.9) {\small $\dagger$};
		\node at (6.7, 4.6) {\small $\ddag$};
	\end{tikzpicture}%
	\caption{Generalisation errors, $\log\epsilon$, vs. training step, $s$ for both teacher 1 ($\dagger$) and teacher 2 ($\ddag$) for the mean-field limit with full feature similarity between teachers. In the second task phase, there is sharp initial forgetting. This is followed by a period of co-learning. Then at around two million steps there is a second turn of forgetting. This corresponds with the point at which the performance on the second task matches the best performance attained by the student with respect to the first teacher in the first phase.}\label{fig: long_time_adjusted_example}
\end{figure}

Initially we computed the long-time limits simply as the differences in generalisation error at the end of training with those at the switch point. However, for forgetting we needed to adjust this procedure slightly. \autoref{fig: long_time_adjusted_example} shows a run associated with a single task configuration in the mean-field limit---in particular, this run is for tasks with full feature overlap.

\newpage
\section{$\tilde{V} = 0$ Row in Mean-Field Limit}\label{app: isolated_row}

\begin{figure}[h!]
\centering
	\pgfplotsset{
		width=0.3\textwidth,
		height=0.2\textwidth,
		every tick label/.append style={font=\tiny},
		xlabel={\small $\alpha$},
		yticklabel style={
        /pgf/number format/fixed,
        /pgf/number format/precision=5
        },
        scaled y ticks=false
		}
	\begin{tikzpicture}
% 		\node at (2.5, 3.8) {\small ReLU};
		\begin{axis}
			[
			xmin=0, xmax=1,
			ymin=-0.04, ymax=0.04,
			ylabel={\small $\forgetting{0}$}
		]
		\addplot graphics [xmin=0, xmax=1,ymin=-0.04,ymax=0.04] {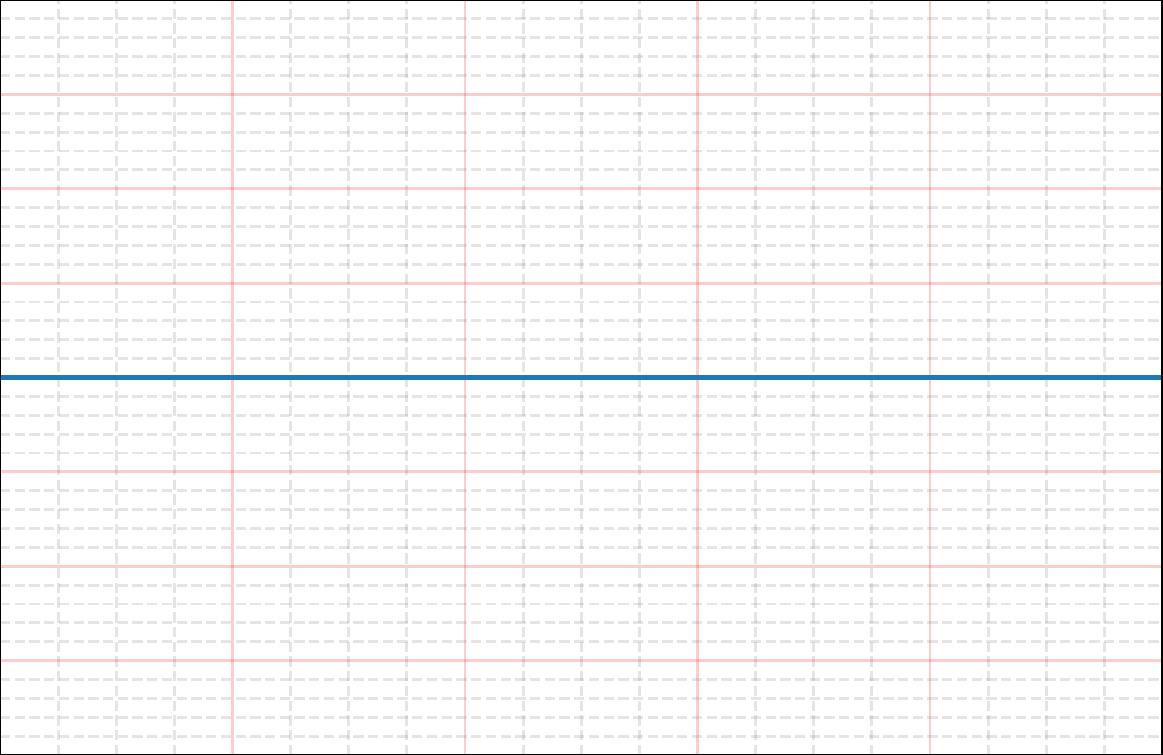};
		\end{axis}
	\end{tikzpicture}%
		\hspace{1em}
		\begin{tikzpicture}
% 		\node at (2.5, 3.8) {\small ReLU};
		\begin{axis}
			[
			xmin=0, xmax=1,
			ymin=0, ymax=0.6,
			ylabel={\small $\forgetting{10}$}
		]
		\addplot graphics [xmin=0, xmax=1,ymin=0,ymax=0.6] {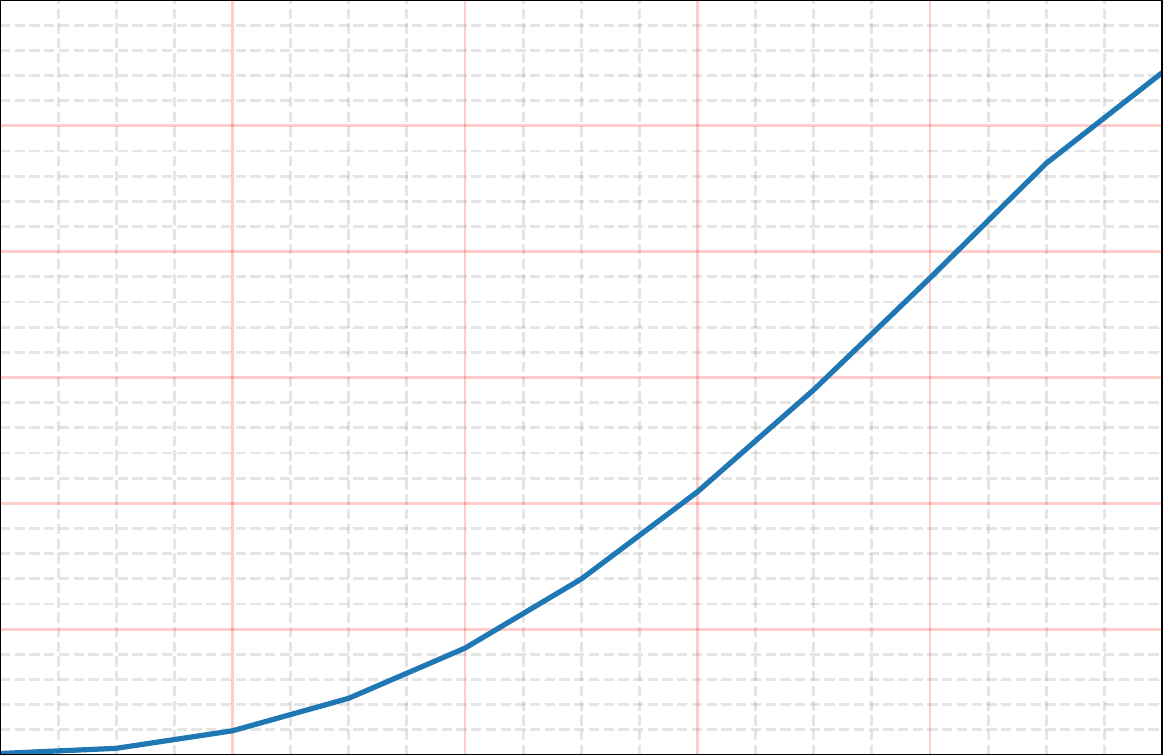};
		\end{axis}
	\end{tikzpicture}%
	\vspace{0.5em}
		\begin{tikzpicture}
% 		\node at (2.5, 3.8) {\small ReLU};
		\begin{axis}
			[
			xmin=0, xmax=1,
			ymin=0.4, ymax=2.1,
			ylabel={\small $\forgetting{100}$}
		]
		\addplot graphics [xmin=0, xmax=1,ymin=0.4,ymax=2.1] {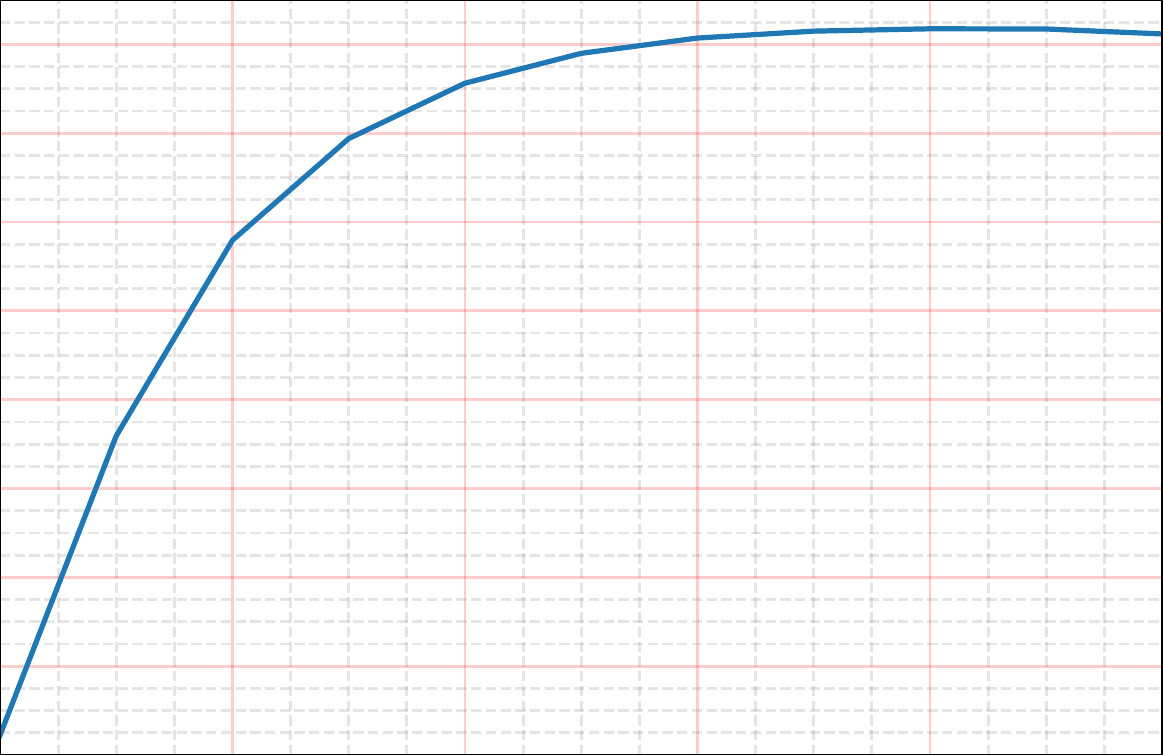};
		\end{axis}
	\end{tikzpicture}%
    \vspace{1em}
    	\begin{tikzpicture}
% 		\node at (2.5, 3.8) {\small ReLU};
		\begin{axis}
			[
			xmin=0, xmax=1,
			ymin=0.6, ymax=2,
			ylabel={\small $\forgetting{1000}$}
		]
		\addplot graphics [xmin=0, xmax=1,ymin=0.6,ymax=2] {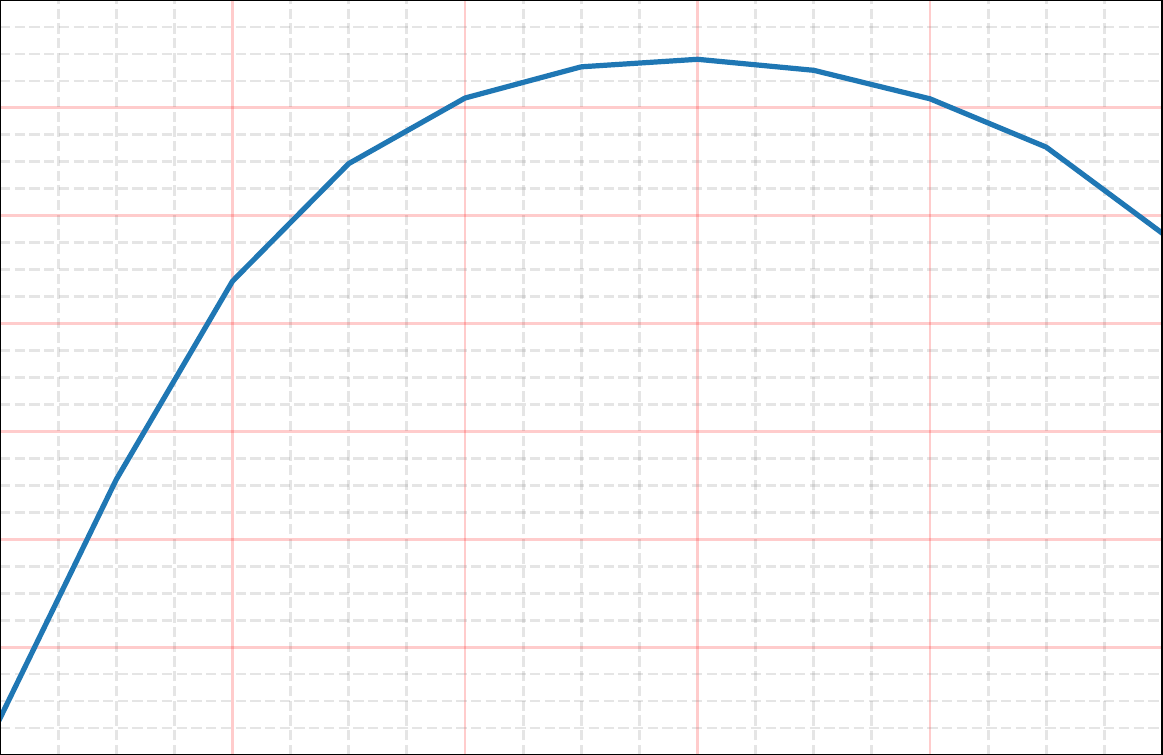};
		\end{axis}
	\end{tikzpicture}%
	\hspace{0.5em}
	\begin{tikzpicture}
% 		\node at (2.5, 3.8) {\small ReLU};
		\begin{axis}
			[
			xmin=0, xmax=1,
			ymin=0.8, ymax=1.7,
			ylabel={\small $\forgetting{10000}$}
		]
		\addplot graphics [xmin=0, xmax=1,ymin=0.8,ymax=1.7] {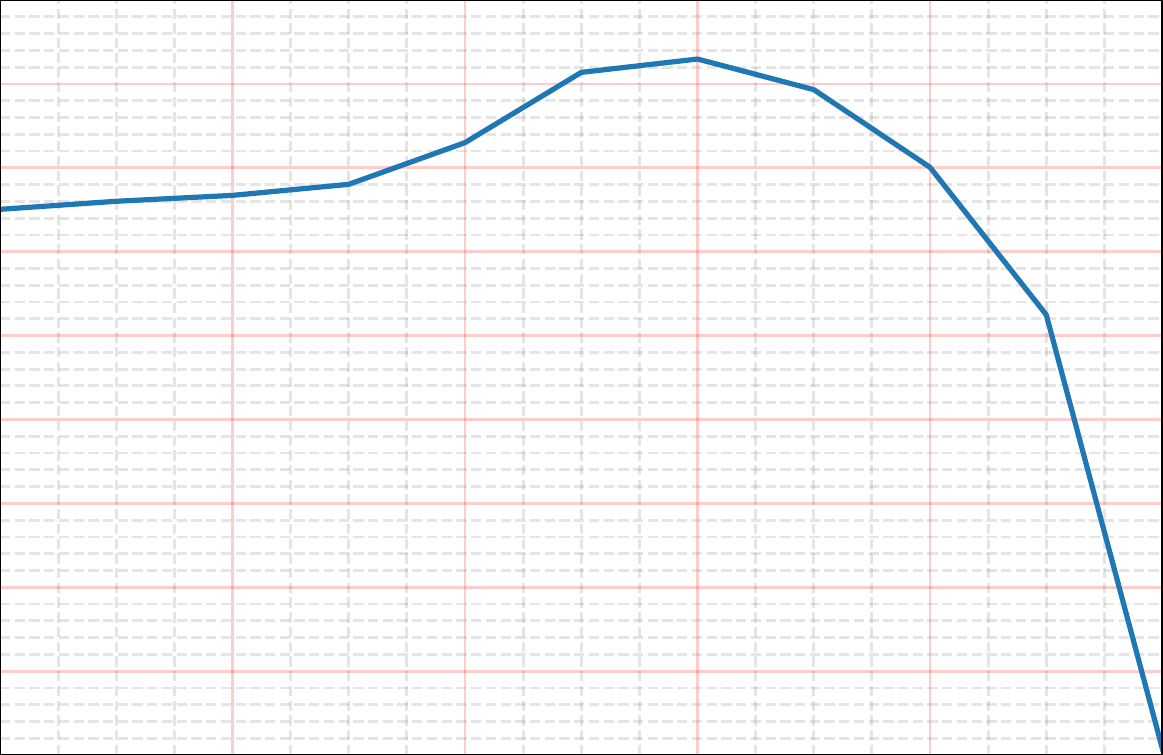};
		\end{axis}
	\end{tikzpicture}%
    \hspace{1em}
    	\begin{tikzpicture}
% 		\node at (2.5, 3.8) {\small ReLU};
		\begin{axis}
			[
			xmin=0, xmax=1,
			ymin=0.3, ymax=1.7,
			ylabel={\small $\forgetting{T}$}
		]
		\addplot graphics [xmin=0, xmax=1,ymin=0.3,ymax=1.7] {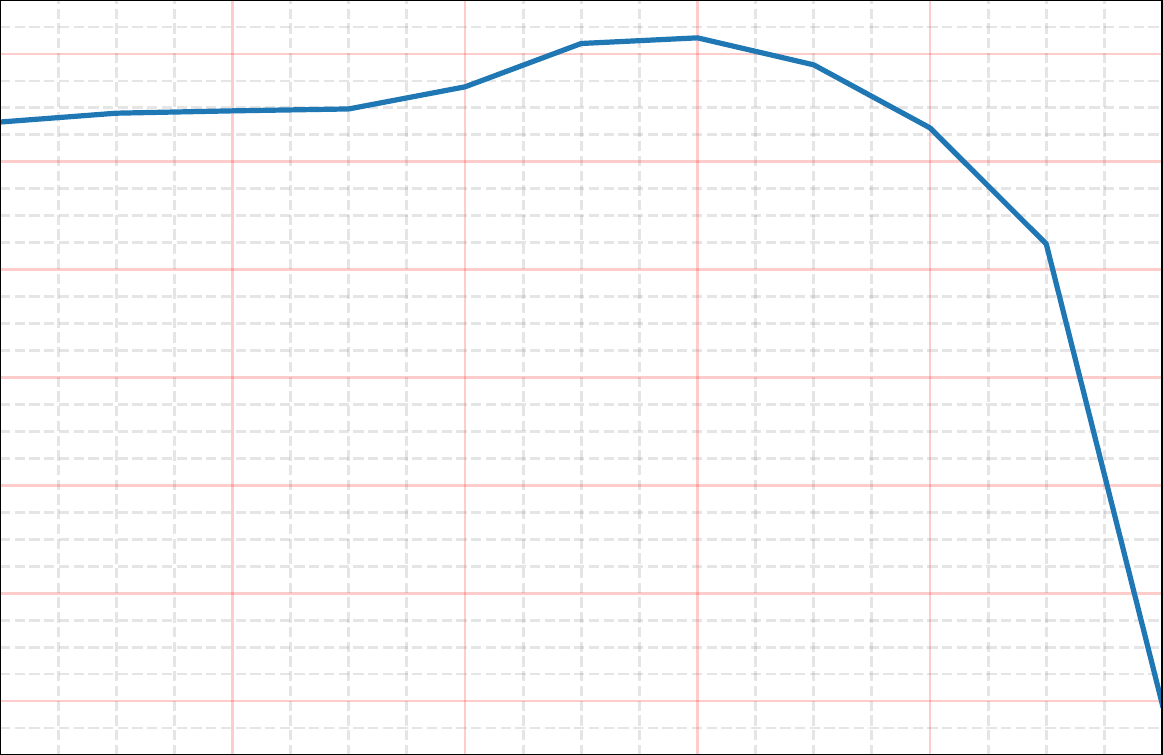};
		\end{axis}
	\end{tikzpicture}%
	\caption{Aggregate forgetting, $\forgetting{t}$, vs. 
	teacher-teacher feature overlap, $\alpha$, for constant zero readout overlap, $\tilde{V}=0$ in the mean-field limit, at different time intervals post task-switch.}\label{fig: individual_row_cross_sections}
\end{figure}

We noted in~\autoref{fig: 2d_aggregate} that the orthogonal readout row, $\tilde{V}$, displays similar trends to the results of varying the feature similarity in the ODE limit. Here we show more details plots from the row beyond the coarse heatmap in~\autoref{fig: 2d_aggregate}.~\autoref{fig: individual_row_cross_sections} shows cross sections of forgetting vs. $\alpha$ at different intervals after the switch. They are the equivalent plots of~\autoref{fig: empirical_task_sim} but for the orthogonal readout row runs of~\autoref{fig: 2d_aggregate}. They show that as for the feature similarity variation in the ODE limit, there is a non-mnotonic relationship between similarity and forgetting such that the intermediate similarity is worst. The development of the shape of the cross-section is also similar. Trivially it begins flat. The non-monoticity is sharpest at intermediate intervals after the switch, and in the long-time limit flattens out again with a wide peak and very little forgetting for large overlap. 

\newpage
\section{Readout Bias on Feature Solution}\label{app: larger_feature_movement}

\begin{figure}[!ht]
\centering
    % \hspace{1em}
    \pgfplotsset{
		width=0.5\textwidth,
		height=0.4\textwidth,
		xlabel={$s$},
		}
    \begin{tikzpicture}
		\begin{axis}
			[
			scaled x ticks = true,  
			xmin=0, xmax=3000000,
			ymin=-0, ymax=1,
			ylabel={MSE($\wb|_s, \wb|_{\tilde{s}})$)}
		]
		\addplot graphics [xmin=0, xmax=3000000,ymin=0,ymax=1] {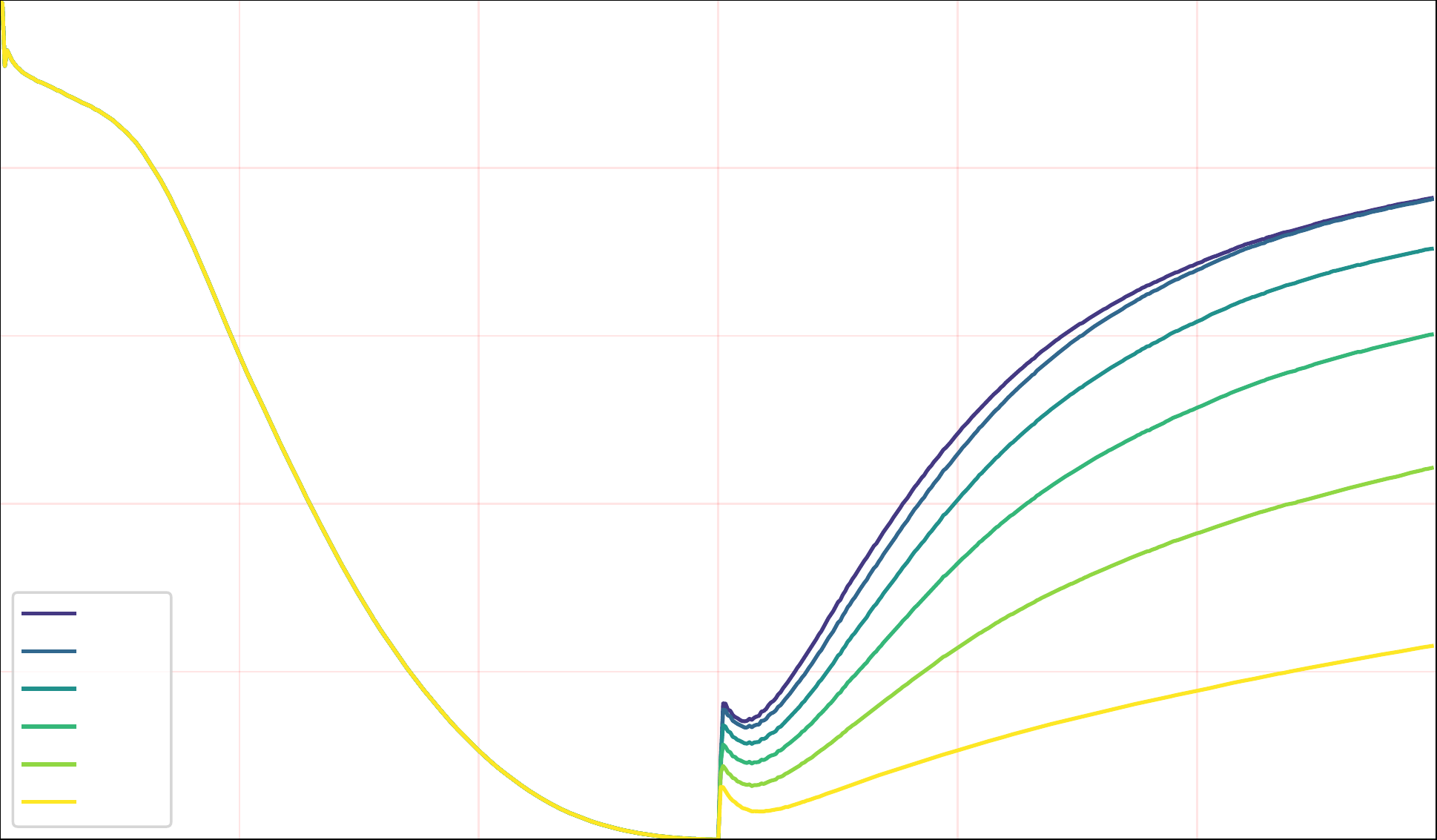};
		\end{axis}
		\node at (0.6, 1.4) {\tiny 0.0};
        \node at (0.6, 1.175) {\tiny 0.2};
        \node at (0.6, 0.95) {\tiny 0.4};
        \node at (0.6, 0.7) {\tiny 0.6};
        \node at (0.6, 0.45) {\tiny 0.8};
        \node at (0.6, 0.225) {\tiny 1.0};
	    \hspace{1em}
	\end{tikzpicture}%
	\caption{(Normalised) mean squared error between the student feature weights at a given step of training and the student feature weights at the switch point, $\wb|_s, \wb|_{\tilde{s}})$, vs. training step, $s$ for full feature similarity and various readout similarity configurations in the mean-field limit. Trivially the MSE is 0 at the switch. After the switch, despite moving onto a new teacher with the same features as the first teacher, the student feature weights move. However they move more for task configurations in which the second readout weights are more dissimilar from the first.}\label{fig: feature_movement}
\end{figure}

One of the interesting results we found from the experiment shown in~\autoref{fig: 2d_aggregate} was that for full feature overlap there was still variation in transfer ability for different readout similarities. After the switch in our multi-head student setup, the student is given a new set of randomly initialised head weights. The previously learned readout weights for the first task are (as far as the transfer ability is concerned) discarded. This newly initialised student head will be (approximately) orthogonal to all of the second teacher head weights, regardless of the relationship between the second teacher head weights and the first teacher head weights. Despite this, there is better transfer for the tasks where there is overlap in the teacher readouts. We hypothesis that this is due to a bias in~\gls{sgd} dynamics: during the first task phase, the local minimum that the solution finds within the feature space is biased by the readout weights it is concurrently trying to optimise. Taking an extreme example, suppose you have two hidden nodes and teacher 1 has readout weight = 1 on node 1, and 0 on node 2. While training on task 1, the network will not learn the input-to-hidden node 2 weights since this node does not impact the output. Therefore there will be a transfer cost if the second task relies on both nodes, which arises from the requirement to learn the input-to-hidden weights that were unimportant for task 1. We verify this idea empirically by tracking the movement of the feature weights after the task switch for different readout similarities. The results are shown in~\autoref{fig: feature_movement} and demonstrate that the feature weights move more (further away from the solution found for task 1, which has identical features to task 2) for task configurations with lower readout similarity.

\end{document}